\documentclass{article}

\usepackage{microtype}
\usepackage{graphicx}
\usepackage{subfigure}
\usepackage{booktabs} 

\usepackage{hyperref}

\usepackage[accepted]{icml2024}

\usepackage{amsmath}
\usepackage{amssymb}
\usepackage{mathtools}
\usepackage{amsthm}
\usepackage{bm}

\usepackage{enumitem}
\setitemize{itemsep=2pt,topsep=0pt,parsep=0pt,partopsep=0pt,leftmargin=*}

\setenumerate{itemsep=2pt,topsep=0pt,parsep=0pt,partopsep=0pt,leftmargin=*}

\makeatletter
\renewcommand{\paragraph}{%
  \@startsection{paragraph}{4}%
  {\z@}{1ex \@plus 1ex \@minus .2ex}{-1em}%
  {\normalfont\normalsize\bfseries}%
}
\makeatother

\def\w{\bm w}
\def\x{\bm x}

\def\I{\bm I}
\def\v{\bm v}

\def\A{\bm A}

\usepackage[capitalize,noabbrev]{cleveref}

\theoremstyle{plain}

\theoremstyle{definition}

\theoremstyle{remark}

\usepackage[textsize=tiny]{todonotes}

\icmltitlerunning{A Dynamical Model of Neural Scaling Laws}

\begin{document}

\twocolumn[
\icmltitle{A Dynamical Model of Neural Scaling Laws}



\icmlsetsymbol{equal}{*}

\begin{icmlauthorlist}
\icmlauthor{Blake Bordelon}{seas,kemp}
\icmlauthor{Alexander Atanasov}{phys,kemp}
\icmlauthor{Cengiz Pehlevan}{seas,kemp}
\end{icmlauthorlist}

\icmlaffiliation{seas}{SEAS, Harvard University}
\icmlaffiliation{kemp}{Kempner Institute, Harvard University}
\icmlaffiliation{phys}{Department of Physics, Harvard University}

\icmlcorrespondingauthor{Cengiz Pehlevan}{cpehlevan@seas.harvard.edu}

\icmlkeywords{Scaling Laws, Mean Field Theory, Infinite Width Limits, Deep Ensembles}

\vskip 0.3in
]



\printAffiliationsAndNotice{}  

\begin{abstract}
On a variety of tasks, the performance of neural networks predictably improves with training time, dataset size and model size across many orders of magnitude. This phenomenon is known as a neural scaling law. Of fundamental importance is the compute-optimal scaling law, which reports the performance as a function of units of compute when choosing model sizes optimally. We analyze a random feature model trained with gradient descent as a solvable model of network training and generalization. This reproduces many observations about neural scaling laws. First, our model makes a prediction about why the scaling of performance with training time and with model size have different power law exponents. Consequently, the theory predicts an asymmetric compute-optimal scaling rule where the number of training steps are increased faster than model parameters, consistent with recent empirical observations. Second, it has been observed that early in training, networks converge to their infinite-width dynamics at a rate $1/\textit{width}$ but at late time exhibit a rate $\textit{width}^{-c}$, where $c$ depends on the structure of the architecture and task. We show that our model exhibits this behavior. Lastly, our theory shows how the gap between training and test loss can gradually build up over time due to repeated reuse of data. 
\end{abstract}

\vspace{-25pt}
\section{Introduction}
\vspace{-5pt}

\begin{figure*}[h]
    \centering   
    \subfigure[One-Pass CIFAR-5M Test Dynamics ]{\includegraphics[width=0.44\linewidth]{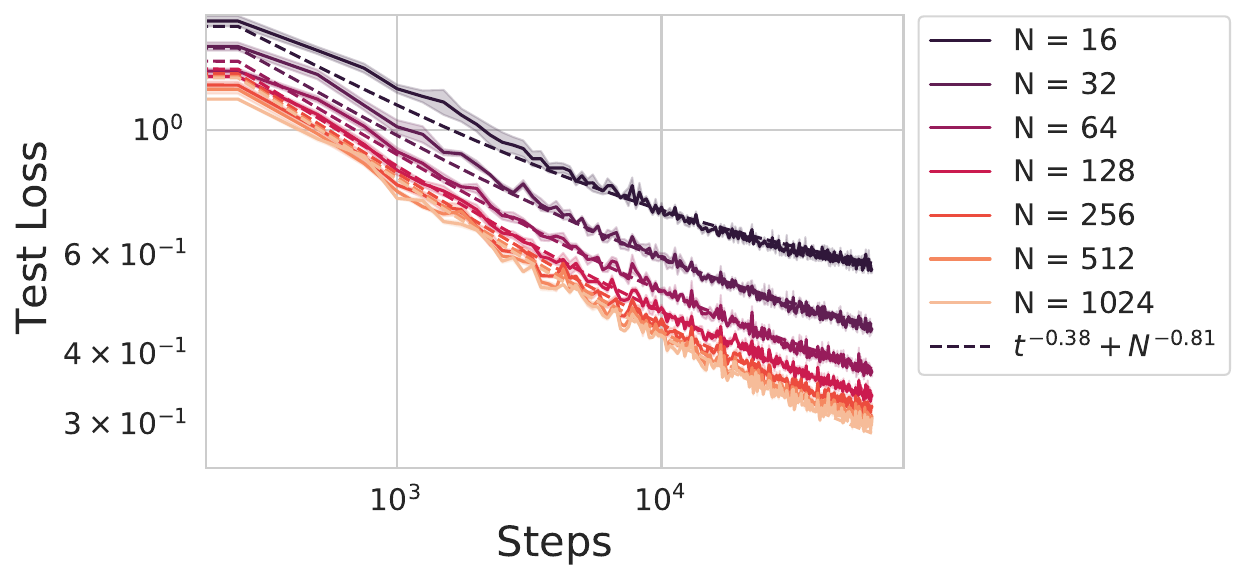}} \hspace{0.5in}
    \subfigure[CIFAR-5M Compute Scaling]{\includegraphics[width=0.3\linewidth]{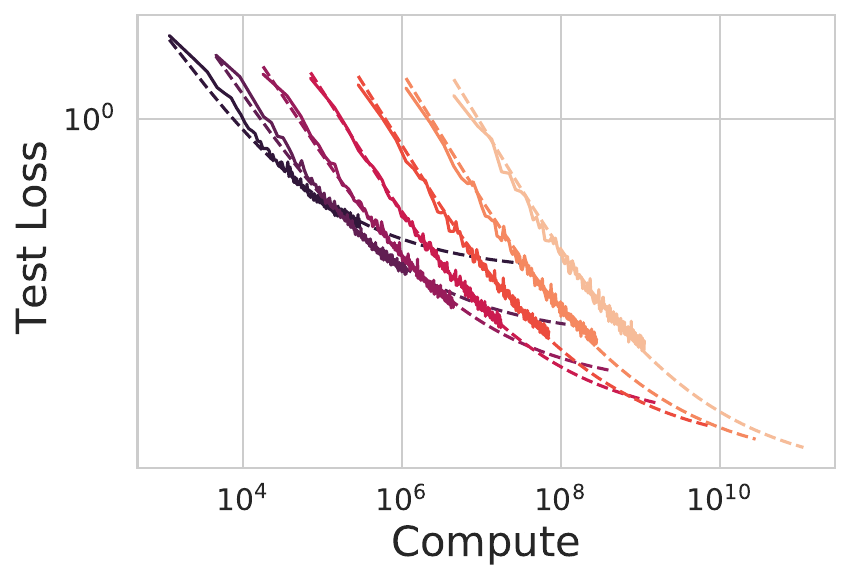}}
    \subfigure[Wikitext-100M Train and Test]{\includegraphics[width=0.42\linewidth]{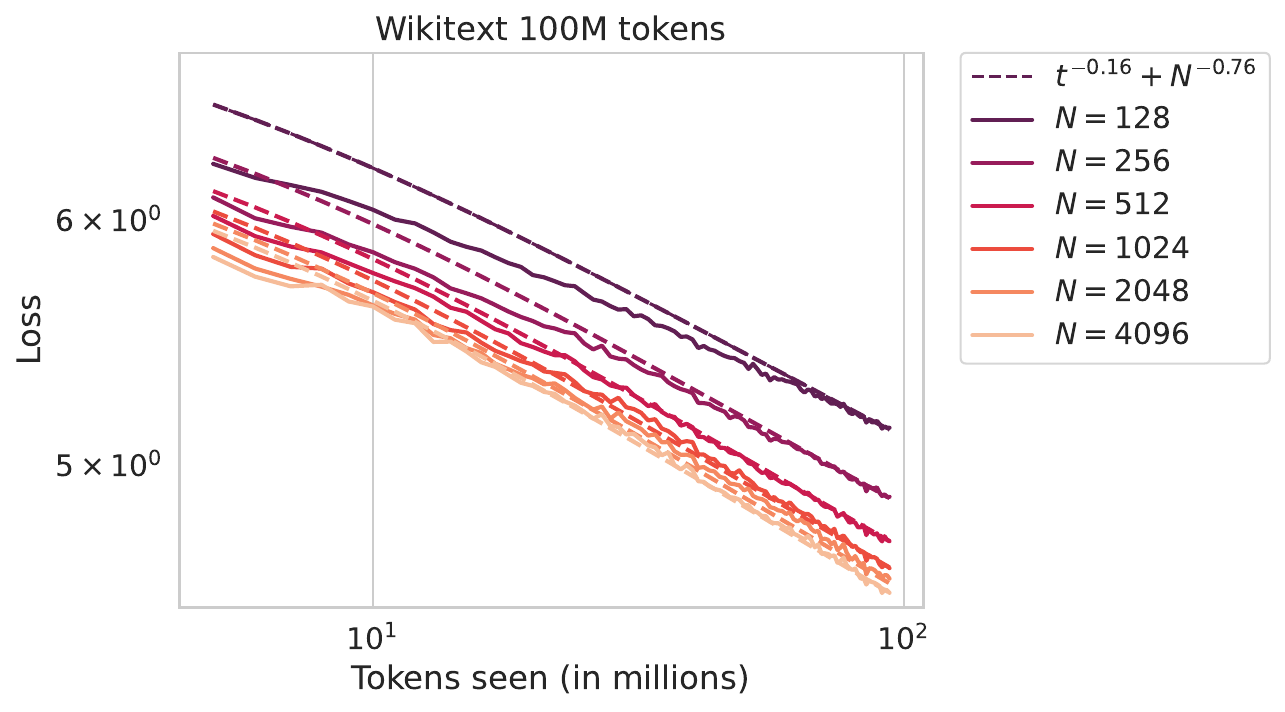}} \hspace{0.3in}
    \subfigure[Wikitext-5M Train and Test]{\includegraphics[width=0.38\linewidth]{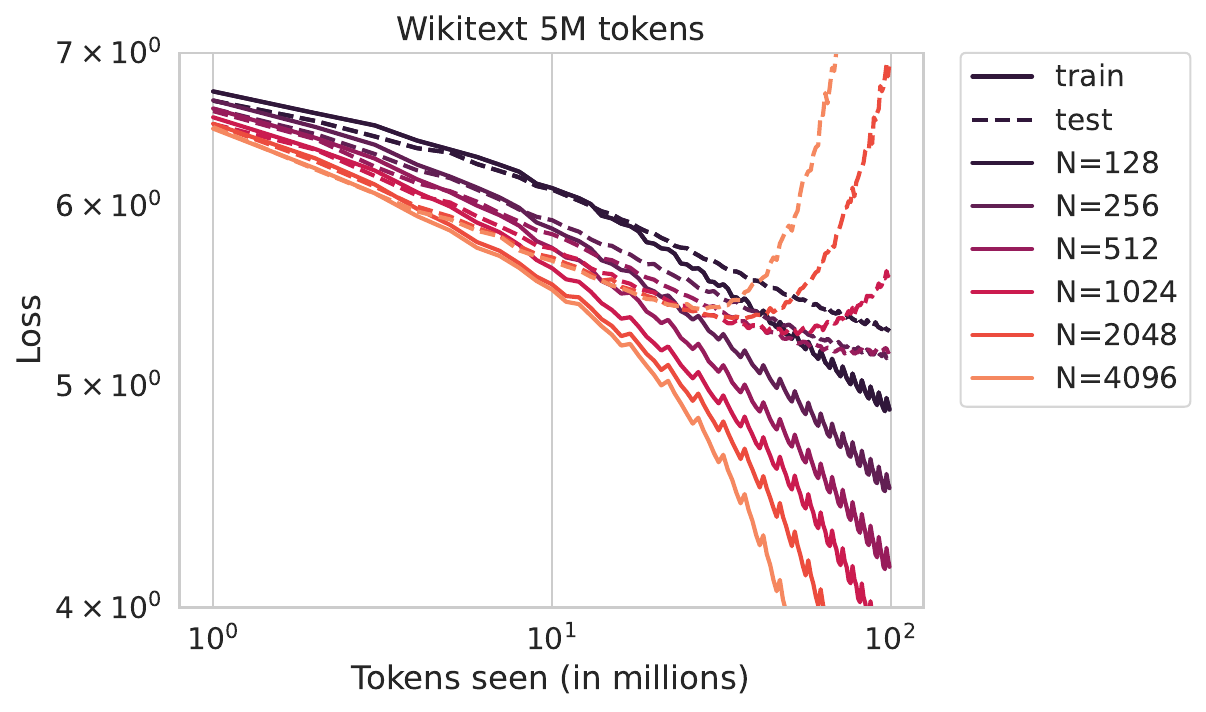}}
    
    \caption{Train and test losses (cross-entropy) as a function of training time $t$ and width $N$. For models trained online, we do not make a distinction between training and test error because each new batch is drawn fresh and would have the same loss in expectation as an independent test set. (a) The test loss of a residual CNN on CIFAR-5M is well described by a fit of the form $\mathcal L \sim t^{-\alpha} + N^{-\beta}$ in the online training regime. (b) The compute optimal strategy requires scaling up both model size and training time simultaneously. (c) Transformer training on wikitext with 100M tokens before data-repetition. Model performance is monotonic in width $N$.  (d)  Wikitext with 5M subsampled tokens. Larger width $N$ is not always better as wider models can overfit.}
    \label{fig:example_width_sweep_data_sweep}
\end{figure*}

Large scale language and vision models have been shown to achieve better performance as the number of parameters and number of training steps are increased. Moreover, the scaling of various loss metrics (such as cross entropy or MSE test loss) has been empirically observed to exhibit remarkably regular, often power law behavior across several orders of magnitude \cite{hestness2017deep,kaplan2020scaling}. These findings are termed ``neural scaling laws". 

Neural scaling laws play a central role in modern deep learning practice, and have substantial implications for the optimal trade-off between model size and training time \cite{hoffmann2022training}, as well as architecture selection \cite{alabdulmohsin2023getting}. Understanding the origin of such scaling laws, as well as their exponents, has the potential to offer insight into better architectures, the design of better datasets \cite{sorscher2022beyond}, and the failure modes and limitations of deep learning systems. Yet, many questions about neural scaling laws remain open.

In this paper, we introduce and analyze a solvable model which captures many important aspects of neural scaling laws. In particular, we are interested in understanding the following empirically observed phenomena:  


\vspace{-10pt}
\paragraph{Test Loss Scales as a Power-law in Training Time and Model Size and Compute.}
In many domains of deep learning, the test loss of a model with $N$ trainable parameters trained for $t$ iterations has been found to scale as $\mathcal L(t,N) \approx \mathcal L_0 + a_t t^{-r_t} + a_N N^{-r_N}$ \cite{kaplan2020scaling, hoffmann2022training}. These scaling law exponents $r_t, r_N$ generally depend on the dataset and architecture. We demonstrate scaling laws on simple vision and language tasks in Figure \ref{fig:example_width_sweep_data_sweep}. The compute is proportional to the number of steps of gradient descent times the model size $C \propto N t$. Setting $N$ and $t$ optimally gives that test loss scales as a power law in $C$. This is the \textit{compute optimal scaling law}. 


\vspace{-3pt}
\paragraph{Compute-Optimal Training Time and Model Size Scaling Exponents Are Different.} A discrepancy in exponents $r_t$ and $r_N$ is usually observed to some degree depdending on the data distribution and architecture \citet{hoffmann2022training, bachmann2024scaling}. The gap between exponents would lead to asymmetric compute-optimal scaling of parameters. For compute budget $C$, model size should scale $N \propto C^{c_1}$ and training time $t \propto C^{c_2}$ with $c_2 > c_1$. This difference in exponents led to a change in the scaling rule for large language models, generating large performance gains.

\vspace{-3pt}
\paragraph{Larger Models Train Faster.}
Provided feature learning is held constant across model scales (i.e. adopting mean-field or $\mu$P scaling), wider networks tend to train faster \cite{yang2021tuning} (Figure \ref{fig:example_width_sweep_data_sweep}). If training proceeds in an online/one-pass setting where datapoints are not repeated, then the wider models will also obtain lower test loss at an equal number of iterations. This observation has been found to hold both in overparameterized and underparameterized regimes \cite{bordelon2023dynamics, vyas2023feature}. 

\vspace{-3pt}

\paragraph{Models Accumulate Finite-Dataset and Finite-Width Corrections.}
Early training can be well described by the learning curves for stochastic gradient descent without reuse of samples (termed the online/ideal limiting dynamics), however over time the effect of reusing data accumulates and leads to worse test performance \cite{nakkiran2021the, mignacco2020dynamical, ghosh2022the, muennighoff2023scaling}. Similarly the gaps in model performance across various model sizes also grow with training time \cite{yang2021tuning, vyas2023feature}. Figure \ref{fig:example_width_sweep_data_sweep} (d) shows overfitting and reversal of ``wider is better" phenomenon due to data reuse.

\vspace{-3pt}
\paragraph{Scaling Exponents are Task-Dependent at Late Training Time, but not at Early Time.}
Prior works \cite{dyerasymptotics, atanasov2023onset, roberts2022principles, bordelon2023dynamics} predict early-time finite-width loss corrections that go as $1/\textit{width}$ near the infinite width limit in either lazy or feature-learning regimes. \citet{bahri2021explaining} et al provide experiments demonstrating the $1/\textit{width}$ convergence.  However, finite-width models trained for a long time exhibit non-trivial exponents with respect to model width \cite{kaplan2020scaling, vyas2023feature}. See Figure \ref{fig:example_width_sweep_data_sweep} for examples of nontrivial scalings at late time on CIFAR-5M and Wikitext. 

\vspace{-5pt}
\paragraph{Ensembling is Not the Same as Going Wider.}
Near the limit of infinite width, finite models can be thought of as noisy approximations of the infinite-width model with noise that can be eliminated through ensembling \cite{dyerasymptotics, geiger2020scaling, atanasov2023onset}. However recent experiments \cite{vyas2023feature} indicate that ensembling is not enough to match performance of larger models.

These phenomena are not unique to deep networks, but can be observed in linear models, or linearized neural networks operating in the lazy/kernel regime. Though this regime does not capture feature learning, it has benefit of analytical tractability. In this paper, we focus on such linearized models to attempt to gain insight into the dynamics of training.


To attempt to explain these phenomena, we develop a mathematically tractable model of neural scaling laws which allows one to simultaneously vary time, model size, and dataset size.
Our contributions are as follows: 
\begin{enumerate}
    \item We analyze the learning dynamics of a structured and randomly projected linear model trained with gradient flow, discrete time SGD, and momentum. In an asymptotic limit of the model, we obtain a dynamical mean field theory (DMFT) description of the learning curve in terms of correlation functions, which measure the cross-time correlation of training and test errors, and response functions which measure sensitivity of the dynamics to small perturbations. 
    \item We solve for the response functions exactly in Fourier domain. This solution reveals faster training for larger models. The low frequency range of these functions allow us to extract the long time limit of the loss. 
    \item We show that the model and data corrections to the dynamics accumulate over time. At early time, each of these corrections has a universal scaling, consistent with prior works \cite{bahri2021explaining}.
    \item For power-law structured features we show that the model exhibits power law scaling of test loss with time, model size and dataset size. While the data and model exponents are the same, the time and model exponents are different in general. We show that this gives rise to an asymmetric compute optimal scaling strategy where training time increases faster than model size.
    \item Our theory explains why ensembling is not compute optimal as it gives less benefit to performance than increase in model size.
    \item We observe in Section \ref{sec:feature_learning} that feature learning networks can obtain better power law scalings, leading to a better compute optimal frontier. We empirically study this phenomenon in Appendix \ref{app:feature_learning}.
\end{enumerate}

\subsection{Related Works}

The learning curves for linear models with structured (non-isotropic) covariates, including infinite-width kernel regression, have been computed using tools from statistical physics and random matrix theory \cite{bordelon2020spectrum, spigler2020asymptotic, canatar2021spectral, simon2021eigenlearning, bahri2021explaining, hastie2022surprises}. \citet{mei2022generalization} analyzed a linear model with random projections of isotropic covariates. There, they study the limiting effects of width and dataset size, and observe model-wise and sample-wise double descent. In \citet{adlam2020neural} a related model is used to study the finite-width neural tangent kernel (NTK) \cite{jacot2018neural} of a given network. Further, \citet{d2020double} and \citet{adlam2020understanding} extend this analysis to understand the different sources of variance in the predictions of random feature models and the effect of ensembling and bagging on the test loss. Other works have extended this to models where an additional untrained projection is applied to the structured covariates \cite{loureiro2021learning, loureiro2022fluctuations,zavatone2022contrasting, atanasov2023onset, maloney2022solvable, zavatoneveth2023learning, ruben2023learning, simon2023more}.  Within this literature, which considered fully trained models, the works of \cite{bordelon2020spectrum,spigler2020asymptotic} derived power-law decay rates for power-law features which were termed resolution limited by \citet{bahri2021explaining} and recovered by \citet{maloney2022solvable}. 

However, we also study the dependence on training time.   The $t \to \infty$ limit of our DMFT equations recovers the final losses computed in these prior works. While these prior works find that the scaling exponents for model-size and dataset-size are the same, we find that the test loss scales with a different exponent with training time, leading to a different (model and task dependent) compute optimal scaling strategy. 

DMFT methods have been used to analyze the test loss dynamics for general linear and spiked tensor models trained with high-dimensional random data \cite{mannelli2019passed, mignacco2020dynamical, mignacco2022effective} and deep networks dynamics with random initialization \cite{bordelon2022self, bordelon2023depthwise}. High dimensional limits of SGD have been analyzed with Volterra integral equations in the offline case \cite{paquette2021sgd} or with recursive matrix equations in the online case \cite{varre2021last, bordelon2022learning}. Random matrix approaches have also been used to study test loss dynamics in linear regression with isotropic covariates by \cite{advani2020high} and for random feature models in \cite{bodin2021model}. In this work, we consider averaging over both the disorder in the sampled dataset and the random projection of the features simultaneously using DMFT. 

Other models and hypotheses for scaling laws instead rely on a discrete collection of subtasks or skills which are learned as compute grows \cite{caballero2022broken, arora2023theory, michaud2023quantization}. Our theory instead focuses on spectral components of a data distribution. 

\section{Setup of the Model}

We consider a ``teacher-student" setting, where data sampled from a generative teacher model is used to train a student random feature model. The teacher and student models mismatch in a particular way that will be described below. This mismatch is the key ingredient that leads to most of the phenomena that we will discuss.

\paragraph{Teacher Model.} Take $\bm x \in \mathbb R^D$ to be drawn from a distribution $p(\bm x)$ with a target function $y(\x)$ expressible in terms of base features $\bm\psi(\x) \in \mathbb R^M$ up to noise: 
\begin{equation}
    y(\x) = \frac{1}{\sqrt M} \w^\star \cdot \bm\psi(\x) + \sigma \epsilon(\x).
\end{equation}
Here $\psi(\x)$ play the role of the infinite-width NTK eigenfunctions, which form a complete basis for square-integrable functions $L^2[p]$. The $\epsilon(\x)$ function describes a component of $y$ with which is uncorrelated with $\psi(\x)$. We work in the eigenbasis of features as in \cite{bordelon2020spectrum}, so the covariance given by:
\begin{align}
    \left< \psi_k(\x) \psi_\ell(\x) \right>_{\bm x \sim p(\bm x)} = \delta_{k \ell} \lambda_k. 
\end{align}
The power law structure in the $\lambda_k$ and $\bm w^*$ entries will lead to power law scalings for the test loss and related quantities.

\paragraph{Student Model.} Our student model is motivated by a scenario where a randomly initialized finite-width network is trained in the linearized or lazy regime \cite{chizat2019lazy,jacot2018neural}. Such training can be described through learning linear combinations of the finite-width NTK features. These features will span a lower-dimensional subspace of the space of square-integrable functions, and relate to infinite-width NTK features in a complicated way. 

To model this key aspect, the student model uses a projection of the $\bm\psi(\x)$ features, $\bm A \bm\psi(\x)$ where $\bm A \in \mathbb{R}^{N \times M}$. These projected features represent the \textit{finite-width} (i.e. empirical) NTK eigenfunctions. This is motivated by the fact that finite width kernel's features can be linearly expanded in the basis of infinite-width features, because infinite-width kernel eigenfunctions are complete.

Our learned function then has the form:
\begin{equation}
\begin{aligned}
    f(\x) = \frac{1}{\sqrt{N}} \w \cdot \bm A \bm\psi(\x).
\end{aligned}
\end{equation}
Here, we will interpret $N$ as the model size with the $N \to \infty$ limit recovering original kernel.  Similar models were studied in \cite{maloney2022solvable, atanasov2023onset}.

We will focus on the setting where the elements of $\A$ are drawn iid from a distribution of mean zero and variance one. See Appendix \ref{app:field_th_deriv} for details on the technical assumptions. The motivations for this choice are (1) tractability and (2) it satisfies the constraint that as $N \to \infty$ the student's kernel approaches the infinite-width kernel $\bm\psi$. 
In more realistic settings, such as when projecting the eigenfunctions of an infinite-width NTK to a finite-width NTK, the form of the $\A$ matrix is generally not known. 



\paragraph{Training.} The model is trained on a random dataset $\mathcal D = \{ \x_\mu, y_\mu \}_{\mu=1}^P$ of size $P$ with gradient flow on MSE loss
\begin{align}
    \frac{\partial}{\partial t} \w(t) = \frac{\sqrt{M}}{P \sqrt{N}} \sum_{\mu=1}^P ( y(\x_\mu) - f(\x_\mu,t) ) \bm A \bm\psi(\x_\mu).
\end{align}
We explore extensions (momentum, discrete time, one-pass SGD in Appendix \ref{app:opt_exts}). We track the test and train loss 
\begin{equation}
\begin{aligned}
    \mathcal L(t) &= \mathbb{E}_{\x} [(f(\x,t) - y(\x))^2],  
    \\ 
    \hat{\mathcal{L}}(t) &= \frac{1}{P} \sum_{\mu=1}^P (f_\mu(t) - y_\mu)^2.
\end{aligned}    
\end{equation}
In small size systems, these losses depend on the precise realization of the data $\mathcal D$ and matrix $\bm A$. These two quantities can be viewed as the \textit{disorder}. For large systems, these losses approach a well-defined limit independent of the specific realization of $\mathcal D, \bm A$. We will use this fact in the next section when analyzing the model.

\section{DMFT for Scaling Laws} 

We next describe a theoretical approach for characterizing the learning curves for this model. The full details of this approach is detailed in Appendices \ref{app:DMFT Equaitons}, \ref{app:field_th_deriv}. 

We derive a mean field theory for $M, N, P$ large. We analyze both the (1) proportional regime where $N/M = \nu, P/M = \alpha$ and $M,N,P \to \infty$, and (2) non-proportional regime where $M \to \infty$ first and $N,P \gg 1$. The theories derived in these limits are structurally identical (App. \ref{app:non_prop_limit}). \footnote{While the proportional limit is exact, the finite size $N,P$ theory will also contain fluctuations across realizations of disorder. When relevant, we show these in experiments by plotting standard deviations over draws of data and projection matrices $\bm A$. This variance decays as $\mathcal{O}(1/P + 1/N)$.}
 
Let $\bm \Psi \in \mathbb R^{P \times M}$ with $\bm \Psi^\mu_k = \psi_k(\bm x^\mu)$. Also define $\bm \Lambda_{ij} = \lambda_i \delta_{ij}$.The discrepancy between the target weights and the model's effective weights is 
\begin{align}
\bm v^0 \equiv \w^\star -  \frac{1}{\sqrt N} \A^\top \w(t).
\end{align}
The test loss is then given by $\mathcal L(t) = \frac{1}{M}\sum_{k} \lambda_k v^0_k(t)^2$. The $\v^0$ vector has the following dynamics:
\begin{align}
    \frac{d}{dt} \v^0(t) = - \left( \frac{1}{N} \A^\top \A \right)\left( \frac{1}{P} \bm\Psi^\top \bm\Psi \right) \v^0(t).
\end{align}
Already, we can see that generalization can be limited if $\bm A^\top \bm A$ or $\bm\Psi^\top \bm \Psi$ are low rank as the dynamics will be frozen in the nullspace of $\left( \frac{1}{N} \A^\top \A \right)\left( \frac{1}{P} \bm\Psi^\top \bm\Psi \right)$.  Using DMFT, we characterize this limit by tracking $\v^0$ together with the following random vectors:
\begin{equation} \label{eq:v_defns}
\begin{aligned}
    \v^1(t) &= \frac{1}{\sqrt M} \bm \Psi \v^0(t), \quad  \v^2(t) = \frac{1}{\alpha \sqrt{M}} \bm \Psi^\top \v^1(t),\\
    \v^3(t) &= \frac{1}{\sqrt M} \bm A \bm v^2(t), \quad \v^4(t) = \frac{1}{\nu \sqrt{M}} \bm A^\top \v^3(t).
\end{aligned}    
\end{equation}
The key summary statistics (also called \textit{order parameters}) are the correlation functions:
\begin{equation*}
\begin{aligned}
    C_0(t,s) &= \frac{1}{M} \v^0(t)^\top \bm \Lambda \v^0(s), \ C_1(t,s) = \frac{1}{P} \v^1(t) \cdot \v^1(s),
    \\ 
    C_2(t,s) &= \frac{1}{M} \v^2(t) \cdot \v^2(s), \ C_3(t,s) = \frac{1}{N} \v^3(t) \cdot \v^3(s), 
\end{aligned}    
\end{equation*}
as well as the response functions:
\begin{equation*}
\begin{aligned}
    &R_1(t,s) = \frac{1}{P} \mathrm{Tr} \left[ \frac{\delta \v^1(t)}{\delta \v^1(s)} \right]\!,  R_{2,4}(t,s) = \frac{1}{M} \mathrm{Tr} \left[ \frac{\delta \v^2(t)}{\delta \v^4(s)}  \right],
    \\
    &R_3(t,s) = \frac{1}{N }\mathrm{Tr} \left[ \frac{\delta \v^3(t)}{\delta \v^3(s)} \right]\!,  R_{0,2}(t,s) = \frac{1}{M} \mathrm{Tr} \left[ \bm\Lambda\frac{\delta \v^0(t)}{\delta \v^2(s)}  \right].
\end{aligned}    
\end{equation*}
Here $\frac{\delta \v^i(t)}{\delta \v^j(s)}$ is the response of $\v^i(t)$ to a kick in the dynamics of $\v^j$ at time $s$. See appendix \ref{app:resp_defn} for details.

\begin{figure*}[ht]
    \centering
\subfigure[$P = 1000$ Test Loss Dynamics]{\includegraphics[width=0.32\linewidth]{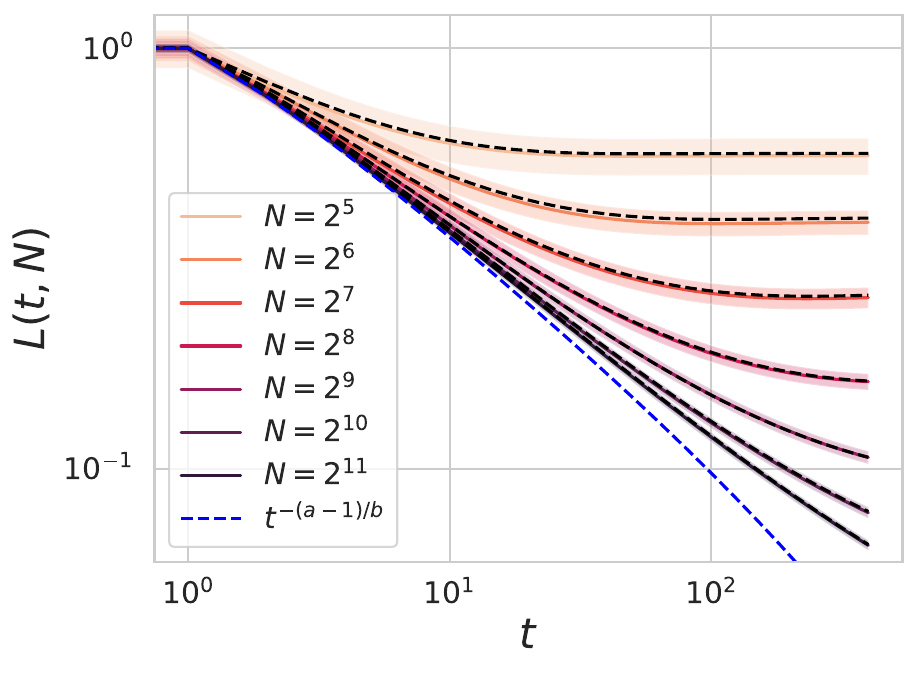}}
\subfigure[Early Time Model Convergence]{\includegraphics[width=0.32\linewidth]{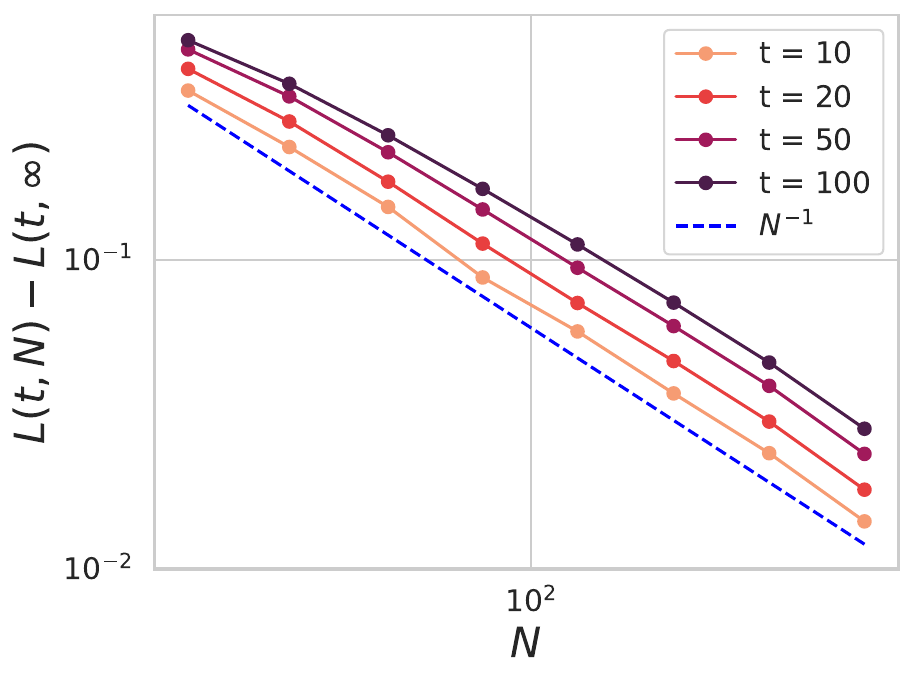}}
\subfigure[Late Time Model Bottleneck]{\includegraphics[width=0.32\linewidth]{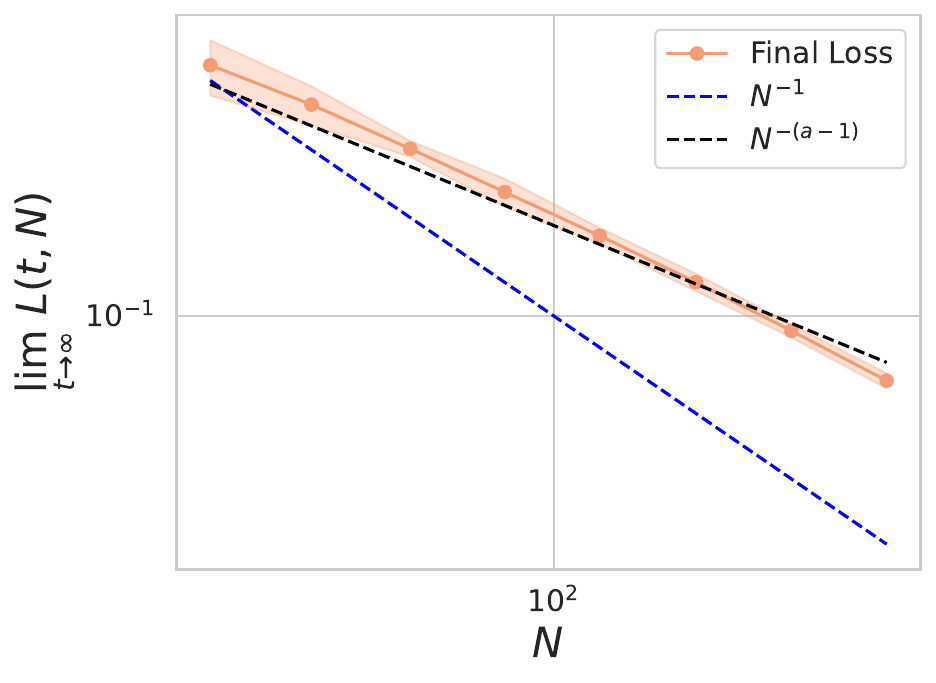}}
\subfigure[$N=1000$ Test Loss Dynamics]{\includegraphics[width=0.32\linewidth]{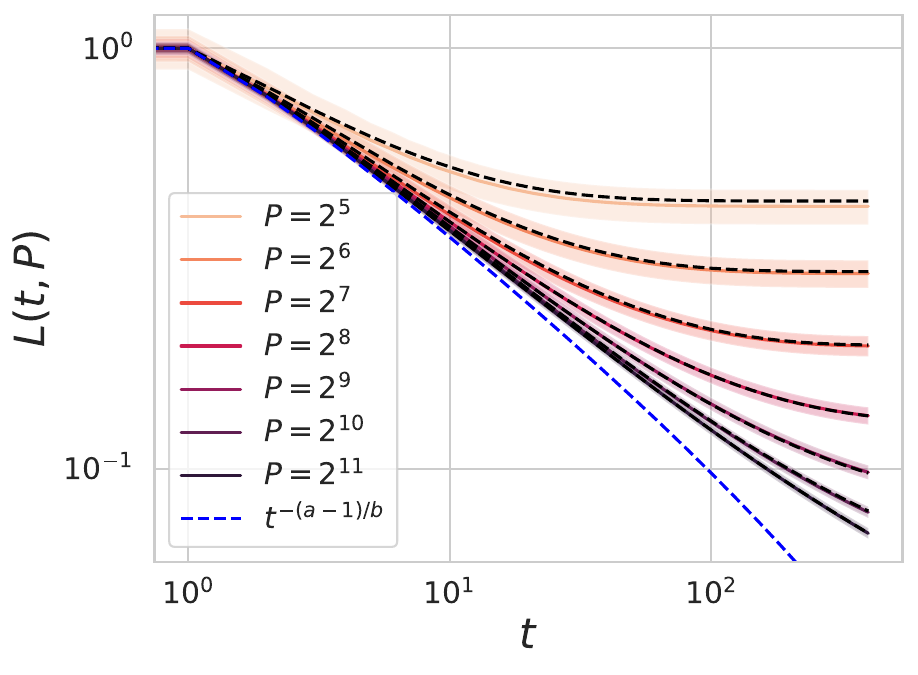}}
\subfigure[Early Time Data Convergence]{\includegraphics[width=0.32\linewidth]{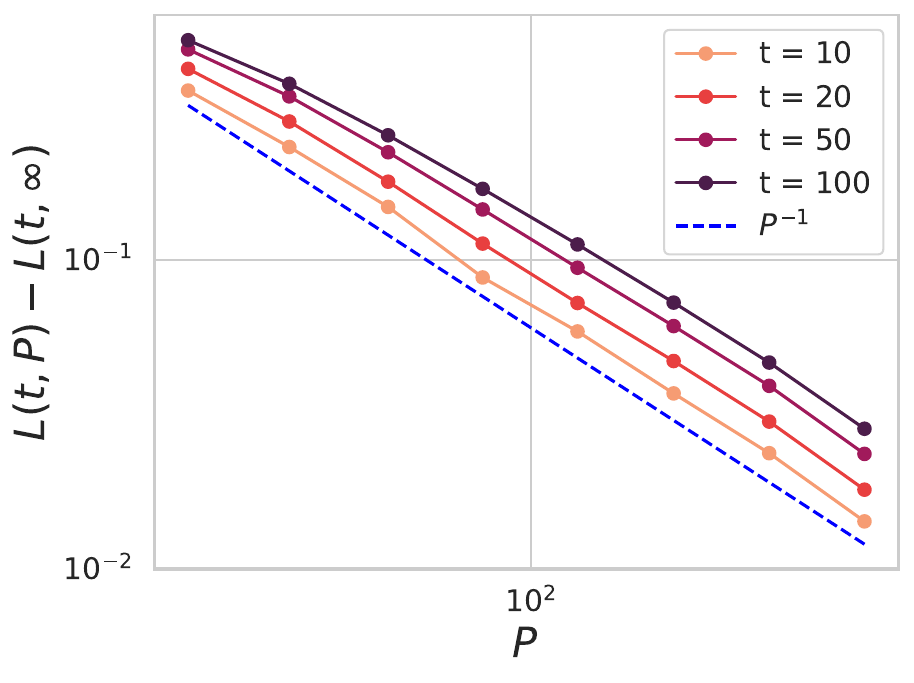}}
\subfigure[Late Time Data Bottleneck]{\includegraphics[width=0.32\linewidth]{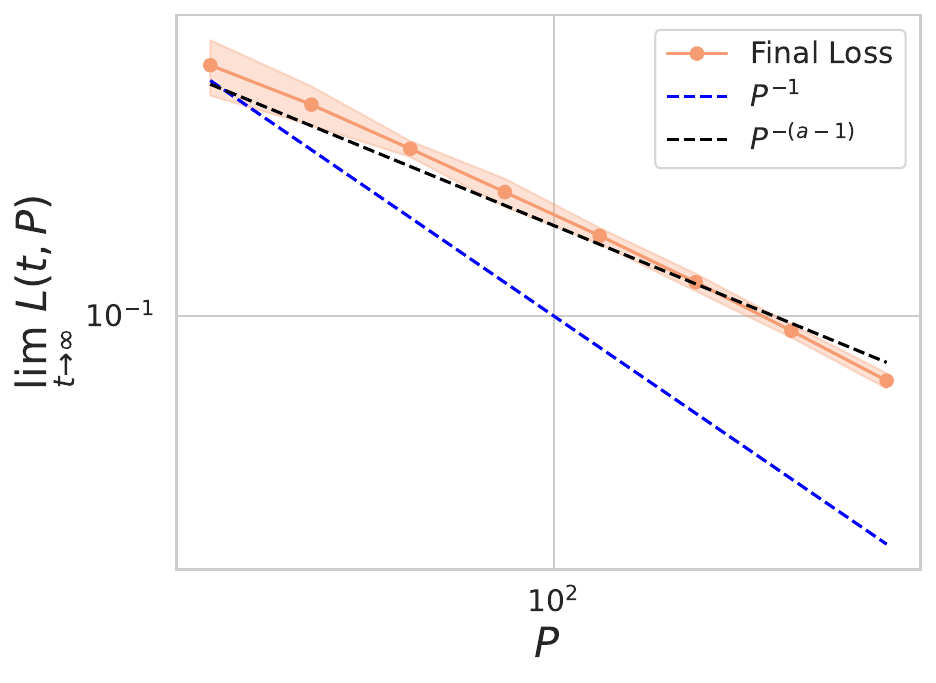}}
\caption{Verification of the various bottleneck scalings for power-law features with $a = 1.5$ and $b=1.25$. Dashed black lines are DMFT solutions while colors are simulations with standard deviation highlighted. (a) The loss dynamics at large $\alpha$ will be bottlenecked by either time or finite $\nu$. (b) Early in training, the loss converges to its limit as $N^{-1}$ (App. \ref{app:early_time}). (c) At long times, the model's asymptotic loss scales as $N^{-(a-1)}$ (App. \ref{app:final_value_dmft}). (d)-(f) The same results but for $N$ and $P$ switched. The model exhibits $1/P$ corrections and early time and power law data bottleneck scalings at late time. }
    \label{fig:power_laws_nu_alpha_sweeps}
\end{figure*}

The test loss $\mathcal L$ and train loss $\hat{\mathcal L}$ are related to the time-time diagonal of $C_0(t,s)$ and $C_1(t,s)$ respectively
\begin{align}
    \mathcal{L}(t) = C_0(t,t) + \sigma^2 \ , \ \mathcal{\hat L}(t) = C_1(t,t).
\end{align}

These collective quantities concentrate over random draws of the disorder \cite{sompolinsky1981dynamic}. We show that these correlation and response functions satisfy a closed set of integro-differential equations which depend on $\alpha, \nu$ which we provide in the Appendices \ref{app:closed_eqns_order_params}. 

Further, we show in Appendix \ref{app:TTI_response_fns} that the response functions possess a \textit{time-translation invariance} property $R(t,s) = R(t-s)$. This enables exact analysis in the Fourier domain $R(\tau) = \int \frac{d\omega}{2\pi} e^{i \omega \tau} \mathcal{R}(\omega)$. These response functions can then be used to solve for the correlation functions $\{ C_0, C_1, C_2, C_3 \}$.

To understand the convergence of the learned function $f$ along each eigenfunction of the kernel, we introduce the transfer function\footnote{There are dynamical analogues of the mode errors in \cite{bordelon2020spectrum,canatar2021spectral} or learnabilities in \cite{simon2021eigenlearning}.} for mode $k$, $H_k(t) \equiv \frac{\partial}{\partial w_k^\star} \left< v_k^0(t) \right>$. Our key result is that the Fourier transform of $H_k$ can be simply expressed in terms of the Fourier transforms of $R_1, R_3$:
\begin{equation}
    \begin{aligned}
     \mathcal{H}_k(\omega) &=  \frac{1}{ i \omega  + \lambda_k \mathcal{R}_1(\omega) \mathcal{R}_3(\omega)  }.
\end{aligned}
\end{equation}

These functions satisfy the self-consistent equations:
\begin{equation}\label{eq:response_fourier}
\begin{aligned}
    \mathcal{R}_1(\omega) = 1 - \frac{1}{P} \sum_k \frac{\lambda_k\mathcal{R}_1(\omega) \mathcal{R}_3(\omega)}{i\omega + \lambda_k\mathcal{R}_1(\omega) \mathcal{R}_3(\omega)  },
    \\
    \mathcal{R}_3(\omega) = 1 - \frac{1}{N} \sum_k \frac{\lambda_k\mathcal{R}_1(\omega) \mathcal{R}_3(\omega)}{ i\omega + \lambda_k\mathcal{R}_1(\omega) \mathcal{R}_3(\omega)  }.
\end{aligned}    
\end{equation}
From these solved response functions $\mathcal R_1,\mathcal{R}_3$, we can compute local solutions to the correlation functions' two-variable Fourier transform $\mathcal C(\omega, \omega')$ which are independent equations for each pair of $\omega,\omega'$. Information about the early dynamics can be extracted from high frequencies $\omega \gg 1$ while information about the late-time limit of the system can be extracted from $\omega, \omega' \to 0$ (App. \ref{app:final_value_dmft}, \ref{app:early_time}). For example, for the final test loss,  
\begin{align}
    \lim_{t \to \infty} \mathcal{L}(t, \alpha, \nu) = \lim_{\omega,\omega' \to 0} (i\omega) (i\omega') \ \mathcal{C}_0(\omega,\omega') .
\end{align}
The full temporal trajectory can be obtained with an inverse Fourier transform of $\mathcal{C}_0(\omega,\omega')$. See Appendix \ref{app:fourier_correlation}.

\begin{figure*}[ht]
    \centering
    \subfigure[SGD with batchsize $B = 32$]{\includegraphics[width=0.375\linewidth]{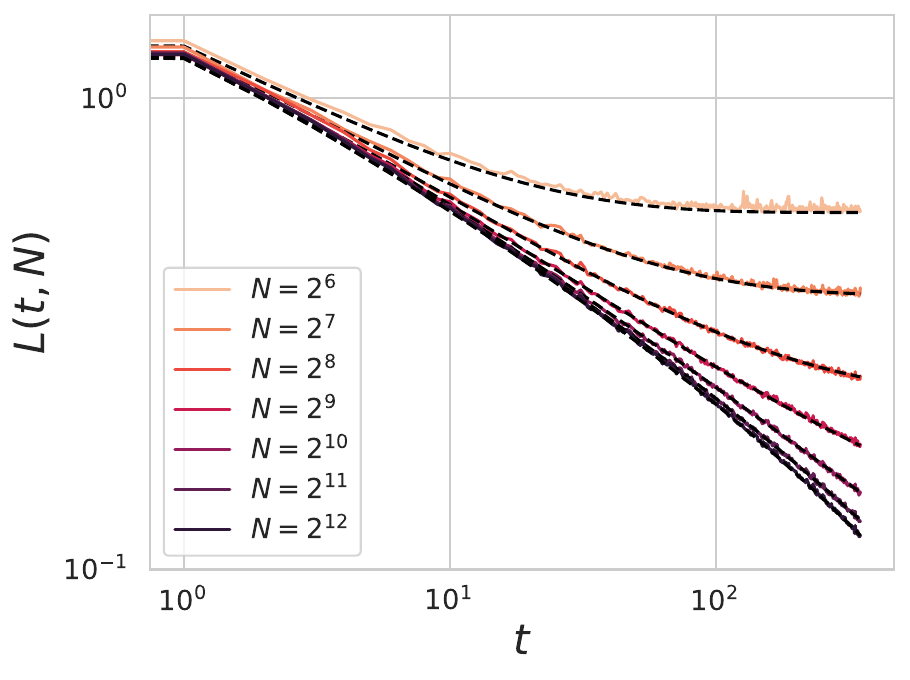}}
    \subfigure[SGD with $N=256$ and varying batchsize $B$ ]{\includegraphics[width=0.375\linewidth]{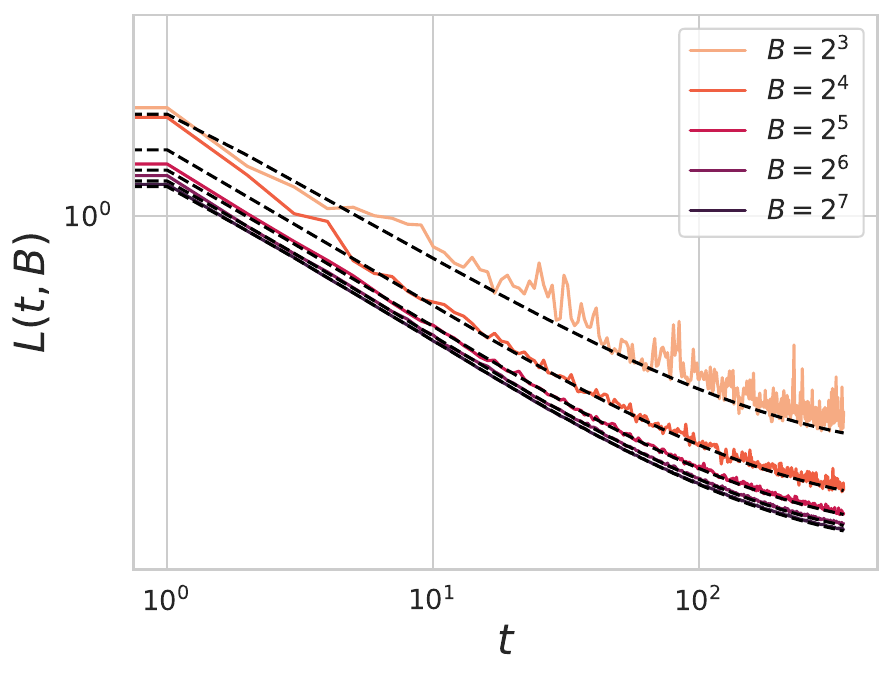}}
    \caption{Our DMFT can also capture online SGD learning including the effect of batch size fluctuations on the loss and the finite $N$ bottleneck. (a) Power law features trained with SGD and a fixed random projection still generates asymptotes which depend on $N$. (b) The batchsize $B$ impacts the loss through additional variance in the dynamics but does not lead to an asymptotic plateau. }
    \label{fig:online_sgd_dmft_version}
\end{figure*}

\begin{figure*}[ht]
    \centering
    \subfigure[$(a,b) = (2,1) \ , \ L(C) \sim C^{-0.5}$]{\includegraphics[width=0.32\linewidth]{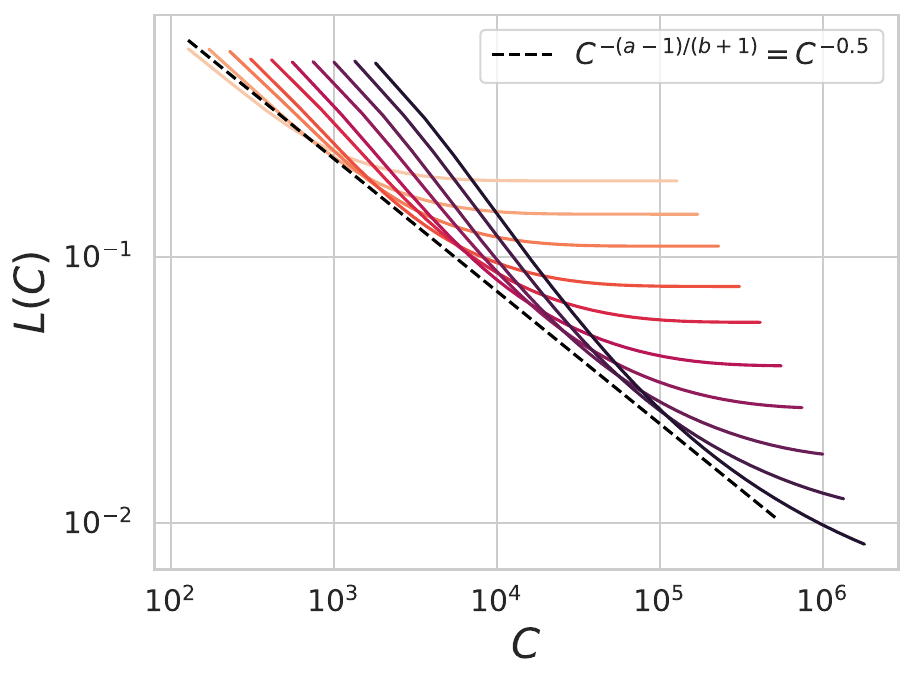}}
    \subfigure[$(a,b)=(2,1.5) \ , \  L(C) \sim C^{-0.4}$]{\includegraphics[width=0.32\linewidth]{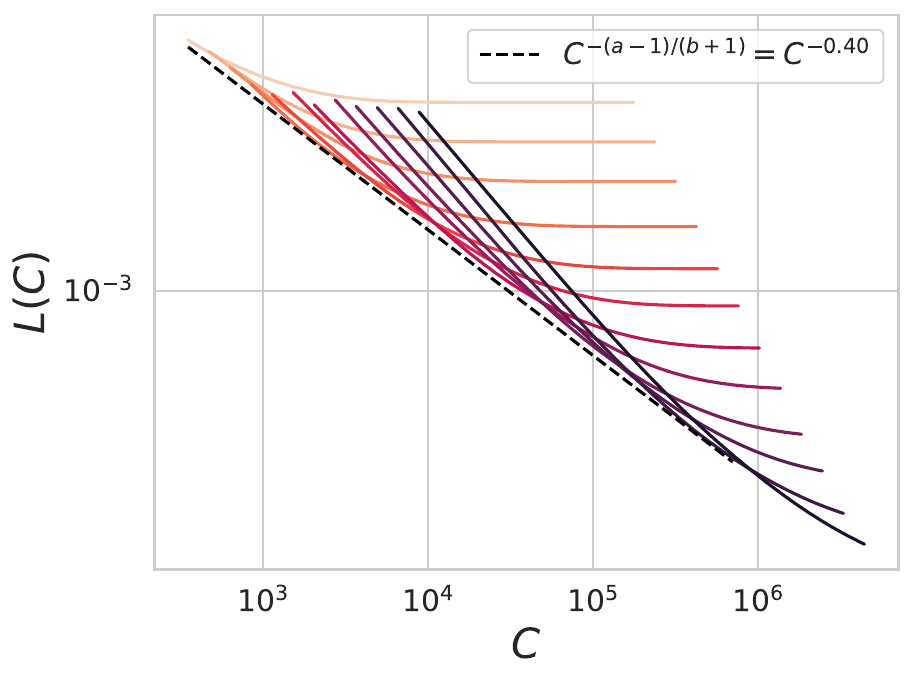}}
    \subfigure[$(a,b) = (2,2) \ , \ L(C) \sim C^{-0.33}$]{\includegraphics[width=0.32\linewidth]{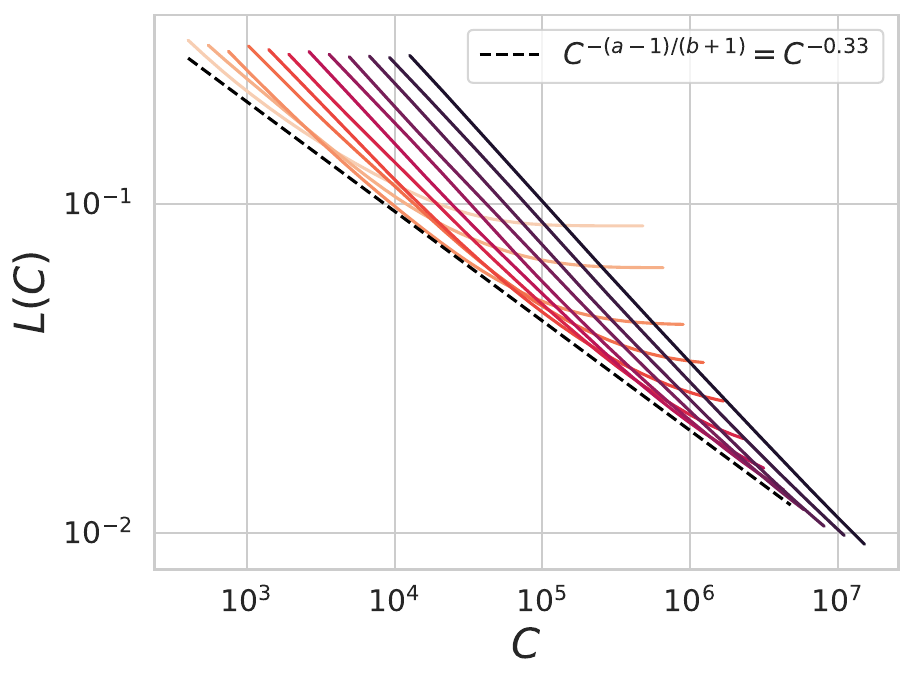}}
    \caption{Compute optimal scaling in our model is determined by tradeoff of time and model-size bottlenecks. Solid colored lines are simulations with power law features and in dashed black is the theoretical prediction of compute optimal scaling. Each color represents varying model sizes with $N \in [2^5, 2^{10}]$. The Pareto frontier is defined as the minimum value of $L$ at each compute $C$ over all possible choices of model size $N$. Although the final losses do not depend on the spectral decay rate $b$ but only on the task-power exponent $a$, the compute optimal scaling depends does depend on $b$.}
    \label{fig:compute_opt_scaling}
\end{figure*}

\vspace{-10pt}
\section{Results}
\vspace{-5pt}

\begin{figure*}[ht]
    \centering
    \subfigure[NTK Spectra for Varying Widths]{\includegraphics[width=0.32\linewidth]{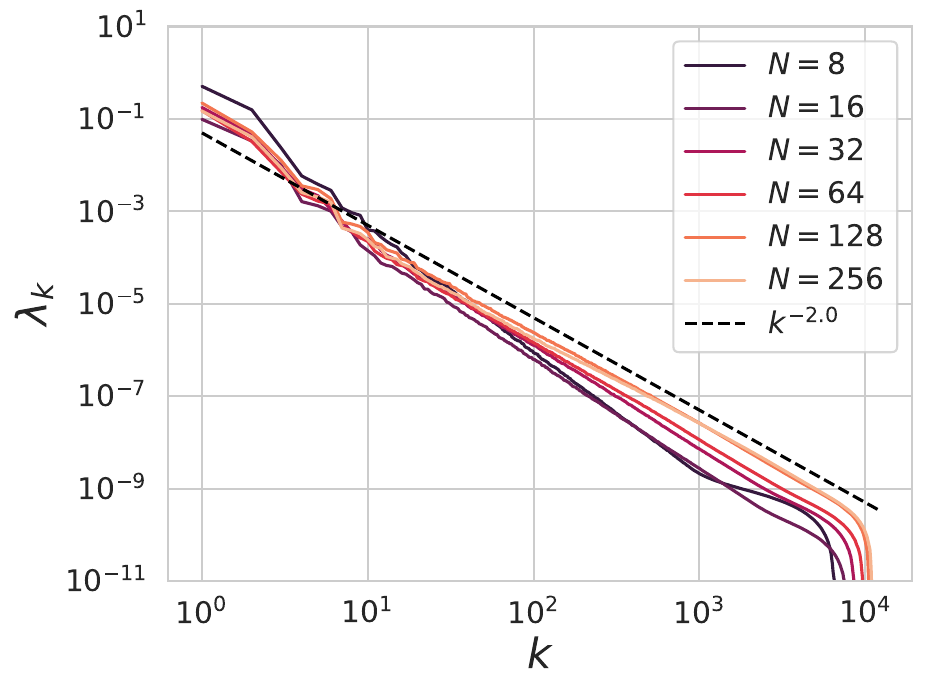}}
    \subfigure[Task-Power Decay]{\includegraphics[width=0.32\linewidth]{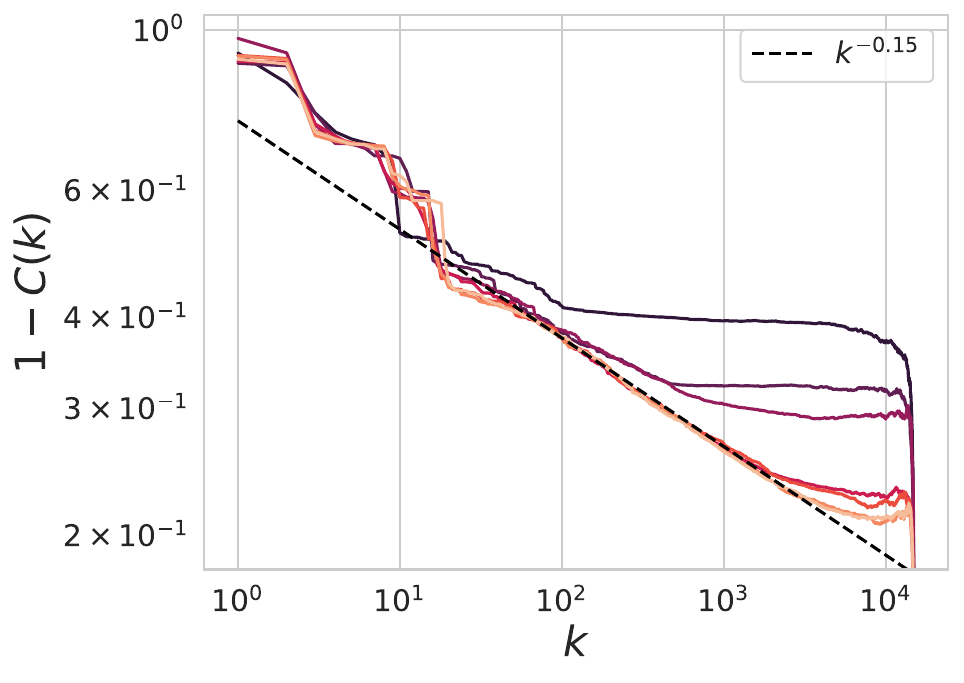}}
    \subfigure[Lazy vs Feature Compute Scaling]{\includegraphics[width=0.32\linewidth]{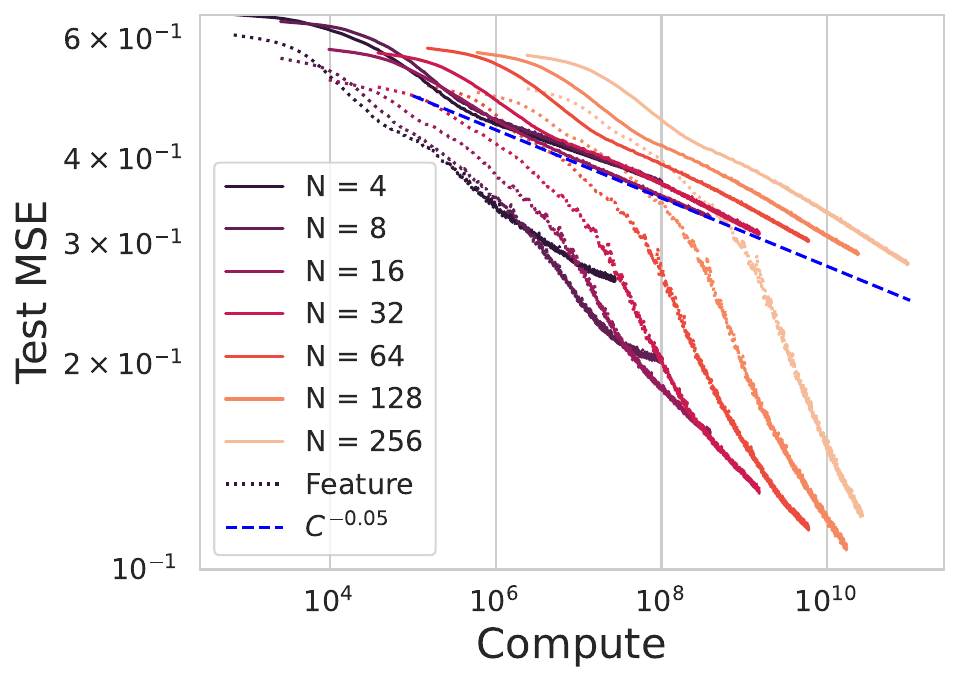}}
    \caption{Our theory predicts time and compute scalings for linearized networks on realistic datasets. (a) The initial NTK spectra and (b) task-power distributions for ResNets of width $N$ on CIFAR-5M are well described by powerlaws $\lambda_k \sim k^{-2.0}$ and $k^{-0.15}$ for large $k$. (c) The predicted compute optimal scaling of for the ResNet obeys $\mathcal{L}_\star(C) \sim C^{-0.05}$.  However, for networks outside of the kernel regime (dashed lines), feature learning can substantially alter the observed scaling laws and improve the loss curves as a function of compute.}
    \label{fig:cifar_5m}
\end{figure*}

Our results hold for any $\lambda_k$ and $w_k^\star$ and we provide some simple analytically solvable examples in Appendix \ref{app:white}. However, based on empirical observations of NTK spectral decompositions on realistic datasets  \cite{bordelon2022learning, spigler2020asymptotic, bordelon2022learning, bahri2021explaining, maloney2022solvable}, here, we focus on the case of power law features. In this setting, eigenvalues and target coefficients decay as a power law in the index $k$
\begin{align}
     (w^\star_k)^2 \lambda_k \sim k^{-a} \ , \  \lambda_k \sim k^{-b}. 
\end{align}
We will refer to $a$ as the \textit{task-power} exponent and $b$ as the \textit{spectral decay} exponent\footnote{These power-law decay rates are also known as source and capacity conditions in the kernel literature \cite{caponnetto2005fast, cui2021generalization}}. See Figure \ref{fig:cifar_5m} (a)-(b) for an example with a Residual CNN on CIFAR-5M.

\paragraph{Test loss power laws.} For power law features, the test loss will generally be bottlenecked by either training time $t$ (steps of gradient descent), the size of the training set $P$, or the size of the model $N$. We can derive bottleneck scalings from our exact expressions for $\mathcal{L}(t,P,N)$ (Appendix \ref{app:power_law_bottleneck})\footnote{The alternative scaling exponents $\mathcal{L} \sim N^{-2b}, P^{-2b}$ occur for very easy tasks which satisfy $a> 2b + 1$, but this condition is rarely satisfied in natural data (Appendix \ref{app:power_law_bottleneck}).}: 
\begin{align}\label{eq:bottlenecks}
    \mathcal{L}(t, P , N) \approx \begin{cases}
        t^{-(a-1)/b}   \  , \ P,N \to \infty  \  , \ \text{(Time)}
        \\
        P^{-\min\{a-1,2b\}}  \ , \ t, N \to \infty \  , \ \text{(Data)}
        \\
        N^{-\min\{a-1,2b\}} \ , \ t, P \to \infty \  , \ \text{(Model)}
    \end{cases}
\end{align}
A consequence of this is an \textit{asymmetry in exponent} between the model and data bottlenecks compared to the time bottleneck. We verify this asymmetry in Figure \ref{fig:power_laws_nu_alpha_sweeps}.

\vspace{-5pt}
\paragraph{Bottlenecks as Rank-Constraints} All three of the bottleneck scalings arise due to \textit{rank constraints} in the effective dynamics. Heuristically, finite training time or the subsampling of data/features leads to an approximate projection of the target function onto the top $k_\star(t, P, N)$ eigenspace of the infinite-width kernel. The components of the target function in the null-space of this projection are not learned. This leads to an approximate test loss of the form
\begin{align}
    \mathcal{L} \approx \sum_{k > k_\star} (w^\star_k)^2 \lambda_k \approx k_\star^{-(a-1)}.
\end{align}
For model and data bottlenecks we have that $k_\star \propto N$ and $k_\star \propto P$ respectively (App. \ref{app:power_law_bottleneck}). On the other hand, $k_\star$ for the time bottleneck also depends on the structure of the features through the exponent $b$. This is because the $k$-th eigenfeature is learned at a timescale $\tau_k \sim k^b$. At time $t$, we have learned the first $k_\star \approx t^{1/b}$ modes and the variance in the remaining modes gives $\sim t^{-(a-1)/b}$. In the limit of $t \to \infty$ our data and model bottleneck scalings agree with the resolution and variance-limited scalings studied in \cite{bahri2021explaining} as well as prior works on kernels and random feature models \cite{bordelon2020spectrum, maloney2022solvable}.

\paragraph{Connection to Online Learning with SGD}

Many modern deep learning models are trained in an online learning setting where each step of training uses a fresh batch of data to estimate the gradient of the population loss and batches are not reused over multiple steps. Our theoretical methods can also handle this regime. At each step $t$ a fresh minibatch of $B$ examples is used to estimate the gradient. In discrete time with learning rate $\eta$ this leads to the following DMFT description of $v^0_k(t)$
\begin{align}
    v^0_k(t+1) &= v^0_k(t) - \eta u^4_k(t)  \nonumber
    \\
    &- \eta \sum_{s \leq t} R_{3}(t,s) [ u^2_k(s) + \lambda_k v^0_k(s)]
\end{align}
where $u^4_k(t), u^2_k(t)$ are zero-mean Gaussian variables with known covariance (see Appendix \ref{app:one_pass_sgd}). The response function $R_3(t,s)$ satisfies a discrete time analog of Equation \eqref{eq:response_fourier}. The most important observation about this regime is that there is no longer a data bottleneck regime. Rather, the bias component of the test error can only be limited by either training time or model size. The finite batch $B$ introduces SGD noise which introduces an additional variance component to the test loss. We illustrate these results in Figure \ref{fig:online_sgd_dmft_version}. The $N \to \infty$ limit recovers the results of \citet{bordelon2022learning} which study online SGD without averaging over a random projection. The continuous time limit of the above expressions obtained from evaluating the theory for small $\eta$ exactly matches the $P \to \infty$ limit of our gradient flow theory presented in the previous section. We will therefore use this limiting behavior to analyze compute optimal tradeoffs of model size and training time. 

\vspace{-5pt}
\paragraph{Asymmetric Compute Optimal Scaling Strategy} We now consider the regime where the total amount of data does not limit performance, but rather training is bottlenecked by time or model size. This could arise in the offline model with very large $P$ or in one-pass SGD with sufficiently small learning rate or sufficiently large batch size (App. \ref{app:one_pass_sgd}). In either case, time and model size should scale differently with compute budget $C = N t$ and $m = \min\{a-1,2b \}$
\begin{align}
    t \sim C^{\frac{b m}{a-1+b m}} \ , \ N \sim C^{\frac{a-1}{a-1+b m}}
    \ , \  \implies \mathcal{L}^\star(C) \sim C^{-\frac{(a-1) m }{a-1+b m}}. 
\end{align}
For the regime of interest where $m = \min\{ a-1, 2b \} = a-1$ this gives $ \mathcal{L}^\star(C) \sim C^{-\frac{a-1}{1+b}}$. We obtain the above scaling by approximating the loss as a sum of the three terms in equation \eqref{eq:bottlenecks} and a constant as in \cite{hoffmann2022training}, see Appendix \ref{app:C_scale}. This analysis suggests that for features that have rapid decay in their eigenspectrum, it is preferable to allocate greater resources toward training time rather than model size as the compute budget increases. This is consistent with the small asymmetries observed in \cite{hoffmann2022training} for language models and the larger asymmetries in MLPs on vision from \cite{bachmann2024scaling}. In the limit as $b \to 1$, the time and parameter count should be scaled linearly together. We verify this scaling rule and its $b$-dependence in Figure \ref{fig:compute_opt_scaling}. 

\vspace{-10pt}

\begin{figure*}[ht]
    \centering
    \subfigure[Test Loss ($P = 128$)]{\includegraphics[width=0.32\linewidth]{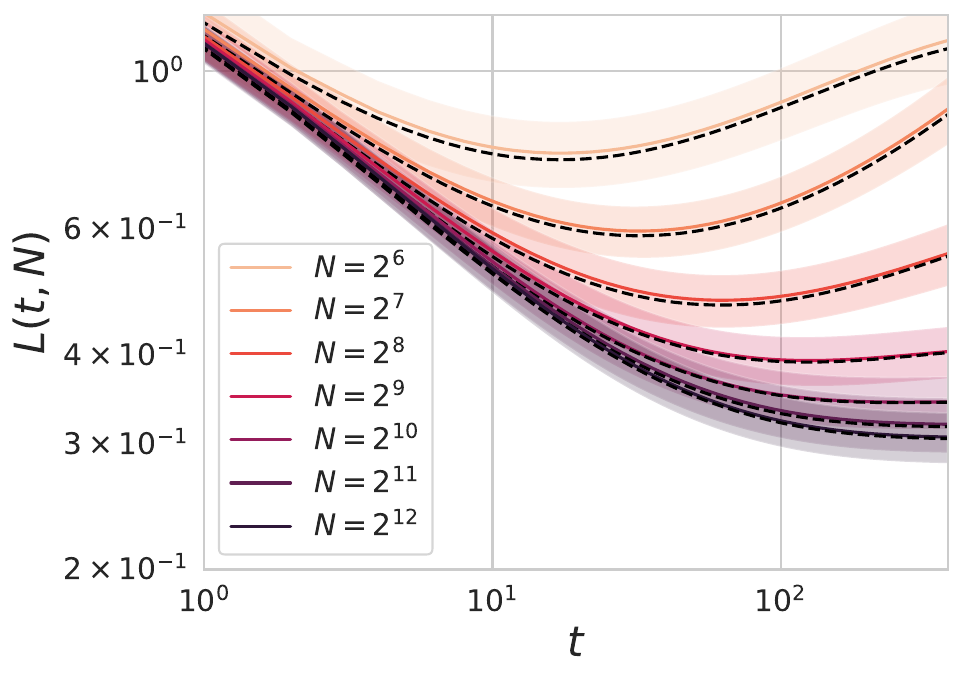}}
    \subfigure[Train Loss ($P=128$)]{\includegraphics[width=0.32\linewidth]{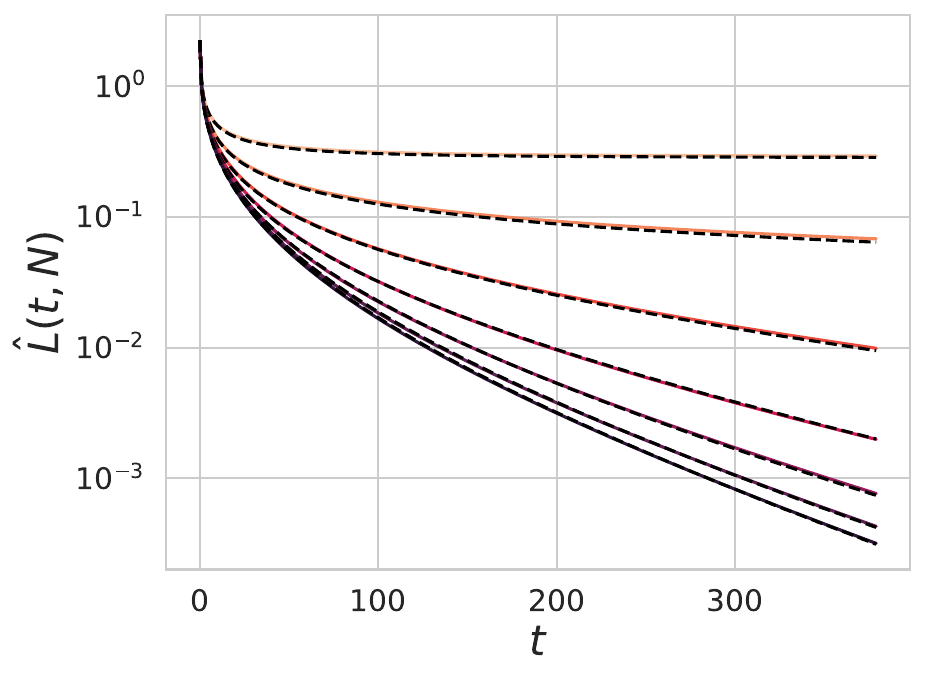}}
    \subfigure[Train-Test Gaps $N = 512$]{\includegraphics[width=0.32\linewidth]{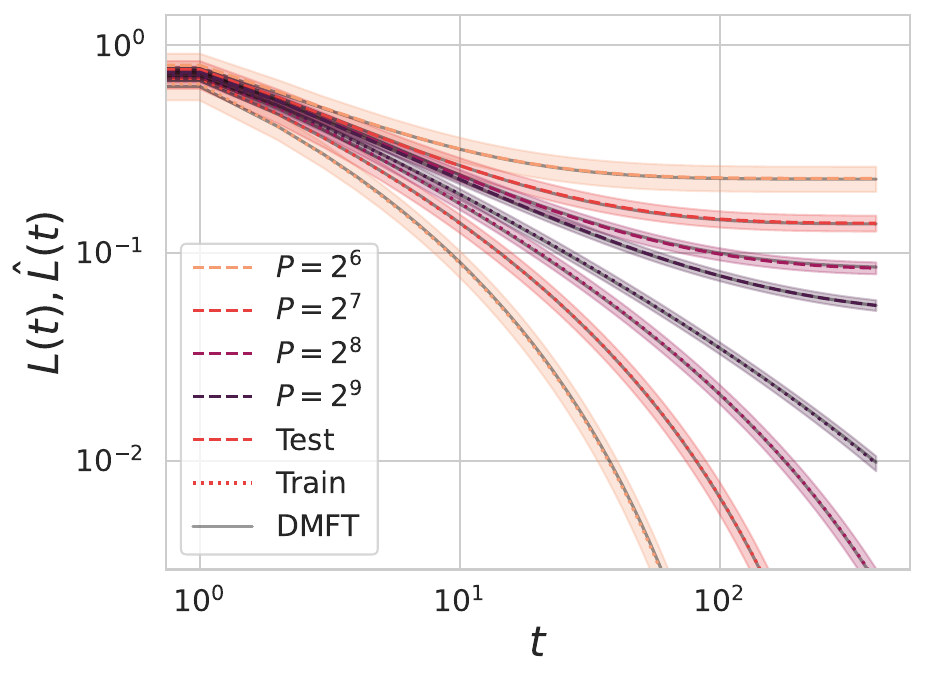}}
    \caption{In a data limited regime, wider networks train faster but cannot indefinitely improve generalization by making $N$ larger. (a) Test loss for power-law features with $a=1.5$ and $b = 1.25$ with $P=128$ and varying $N$. In this regime, there are diminishing returns to making the model size larger. (b) For $N < P$, the model is underparameterized and cannot achieve zero train loss. For $N > P$, the train loss will eventually decay at exponential rate which depends on $N$, despite the test loss saturating. (c) The train and test losses gradually separate at a rate which depends on $P$.  }
    \label{fig:data_bottleneck_effects}
\end{figure*}

\paragraph{Wider is Better Requires Sufficient Data}

Larger models are not always better in terms of test loss for all time $t$, as we showed in Figure \ref{fig:example_width_sweep_data_sweep} (c), especially if the dataset is limited. In Figure \ref{fig:data_bottleneck_effects}, we illustrate that larger $N$ can improve convergence to a data-bottlenecked loss for power law features. However, the loss may still be non-monotonic in training time, motivating regularization or early stopping. 

\vspace{-5pt}

\paragraph{Gradual Buildup of Overfitting Effects}
The exact gap between train and test losses can exactly be expressed in terms of the DMFT order parameters:
\begin{equation}
    \begin{aligned}
    &\mathcal L(t) - \mathcal{\hat L}(t) = - \frac{2}{P} \int_0^t dt' \ R_{0,2}(t,t') C_1(t,t') 
    \\
    &+ \frac{1}{P^2} \int_0^t \int_0^t dt' ds' R_{0,2}(t,t') R_{0,2}(t,s') C_1(t',s').
\end{aligned}
\end{equation}

We derive this relation in the Appendix \ref{app:buildup_overfitting}. At early time this gap goes as $\mathcal{O}(1/P)$ (App. \ref{app:early_time}, \ref{app:buildup_overfitting}). At late time, however, this picks up a nontrivial task-dependent scaling with $P$ as we show in Figure \ref{fig:power_laws_nu_alpha_sweeps} (e)-(f) and App. \ref{app:final_value_dmft}. In Figure \ref{fig:data_bottleneck_effects} (c) we show this gradual accumulation of finite data on the test-train loss gap. For larger datasets $P$ it takes longer training time to begin overfitting (App. \ref{app:buildup_overfitting}). 

\vspace{-5pt}
\paragraph{Ensembling is Not Always Compute Optimal}

\begin{figure*}[ht]
    \centering
    \subfigure[$E$ vs $N$ allocation of compute]{\includegraphics[width=0.33\linewidth]{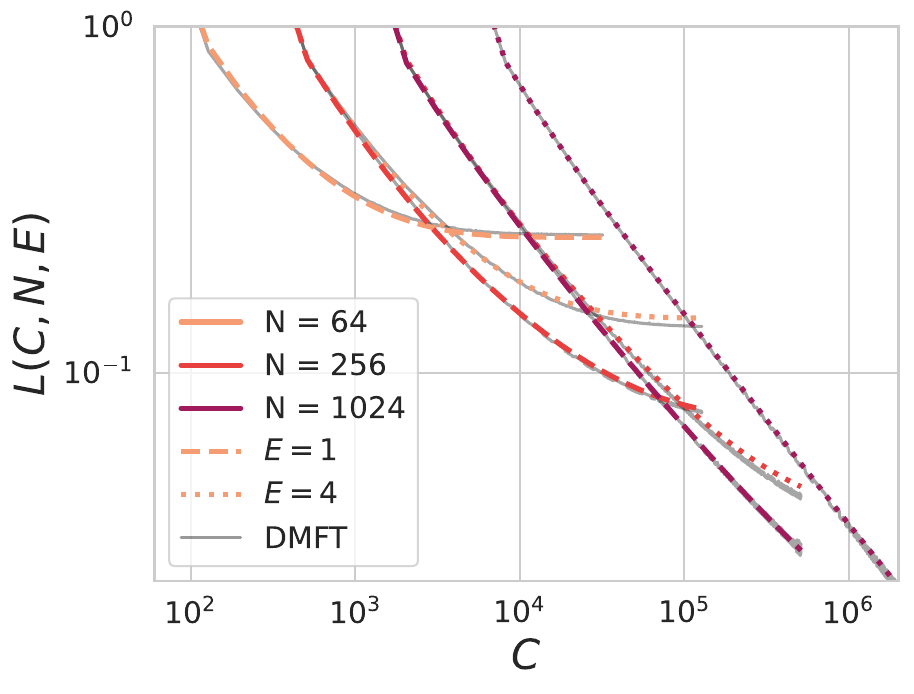}}
    \hspace{0.4in}
    \subfigure[Transfer Functions $H_k(t)$ for $N = 128$]{\includegraphics[width=0.425\linewidth]{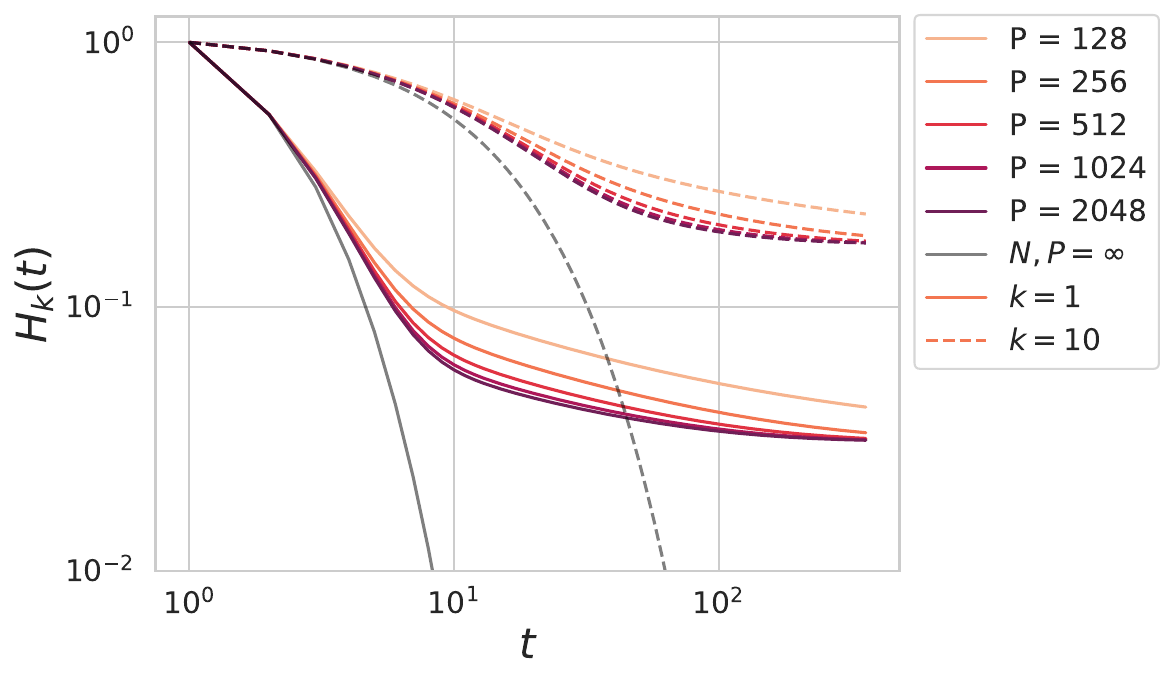}}    
    \caption{Ensembling $E$ models of size $N$ improves performance by reducing initialization variance by a factor of $E$ (see Appendix \ref{app:bias_var_defn}) (a) However, at fixed compute $C = N E t$, increasing the model size $N$ is preferable, since the bias is also reduced. (b) The transfer functions ${H}_k(t)$ computed from the DMFT determine the error as $E \to \infty$ depend on $N, P$ and saturate in performance at long times, while the $N,P \to \infty$ curves decay exponentially.}
    \label{fig:ensemble_figure}
\end{figure*}

Ensembling a set of models means averaging their predictions over the same datasets but with different intitialization seeds. This reduces test loss by reducing the variance of the model output $f$ due to initialization. This improvement can be predicted from an extension of our DMFT (App. \ref{app:ens_bag}). Analogously, bagging over $B$ datasets reduces variance due to sampling of data.

One might imagine that ensembling many finite sized models would allow one to approach the performance of an infinite sized model ($N \to \infty$). If this were possible, the compute optimal strategy could involve a tradeoff between ensemble count and model size. However, recent experiments show that there is a limited benefit from ensembling on large datasets when compared to increasing model size \cite{vyas2023feature}. We illustrate this in Figure \ref{fig:ensemble_figure} (a). Our theory can explain these observations as it predicts the effect of ensembling $E$ times on the learning dynamics as we show in App. \ref{app:ens_bag}. The main reason to prefer increasing $N$ rather than increasing $E$ is that larger $N$ has lower \textit{bias} in the dynamics, whereas ensembling only reduces variance. The bias of the model $\mathcal B$ has the form 
\begin{align}
   \mathcal{B}(t,N,P) = \sum_k \lambda_k (w_k^\star)^2 H_{k}(t,N,P)^2,\vspace{-2pt}
\end{align}
which depend on transfer function $H_k$ that we illustrate for power-law features in Figure \ref{fig:ensemble_figure} (b). Since $H_k(t)$ depend on $N,P$, we see that ensembling/bagging cannot recover the learning curve of the $N,P \to \infty$ system since the bias is limited by finite $N,P$.

\vspace{-10pt}
\section{Tests on Realistic Networks}
\vspace{-5pt}
We now move beyond synthetic power-law datasets and consider realistic image datasets and architectures. We take the CIFAR-5M dataset introduced in \cite{nakkiran2021deep} and consider the task of classfiying animate vs inanimate objects. We plot the spectra of the finite-width NTK at initialization across different widths for a Wide ResNet \cite{zagoruyko2016wide} in Figure \ref{fig:cifar_5m} a). Here the width parameter corresponds to the number of channels in the hidden layers. Following \cite{canatar2021spectral}, we define $C(k)$ as the fraction of the task captured by the top $k$ kernel eigenmodes:
\vspace{-3pt}
\begin{equation}\label{eq:Ck_defn}
    C(k) \equiv \frac{\sum_{i \leq k} \lambda_i (w_i^*)^2 }{\sum_{i} \lambda_i (w_i^*)^2 }.
\end{equation}
Then $1-C(k)$ is the portion of the task left unexplained. We plot this for the initial NTKs across widths in Figure \ref{fig:cifar_5m} b). We extract the spectral decay exponent $b$ and the the task power exponent $a$ from these two curves. Together, these give the learning scaling laws of the linearized neural network model on this dataset. We plot the compute optimal scaling laws of these linearized models in Figure \ref{fig:cifar_5m} c). We also plot the predicted scaling law $C^{-(a-1)/(1+b)}$ in blue and find excellent agreement. 

\vspace{-3pt}
\subsection{The Role of Feature Learning}\label{sec:feature_learning}

\vspace{-3pt}
We also compare these scalings to those of the compute optimal learning curves for feature-learning networks. We train several networks with different widths and initialization seeds for 64 epochs through the dataset. We observe substantially different compute-optimal scaling exponents in the dotted curves of Figure \ref{fig:cifar_5m} c). This means that although our random feature model does capture the correct linearized scaling trends, which have all of the qualities observed in realistic scaling laws, more is needed to capture the acceleration of scaling induced by feature learning. Further analyses of the after-kernels of feature learning networks are performed in Appendix \ref{app:feature_learning}. We see that the kernels continue to evolve substantially throughout training. This indicates that a full explanation of the compute optimal scaling exponents will require something resembling a mechanistic theory of kernel evolution \cite{long2021properties, FortDPK0G20, atanasov2022neural, bordelon2022self}.


\vspace{-3pt}
\section{Conclusion}
\vspace{-3pt}
We have presented a model that recovers a wide variety of phenomena observed in more realistic deep learning settings. Our theory includes not just model size and dataset size as parameters but also explicitly treats the temporal dynamics of training. We observe different scaling exponents for performance in terms of model size and number of time steps. Future work to incorporate kernel evolution into this model could further shed insight into the improved scaling laws in the feature-learning regime. 
Overall, our results provide a theoretical interpretation of compute-optimal scaling as a competition between the training dynamics of the infinite width/infinite data limit and finite model-size bottleneck. 

\vspace{-5pt}
\section*{Impact Statement}
\vspace{-5pt}

This paper presents work whose goal is to advance the field of Machine Learning. There are many potential societal consequences of our work, none which we feel must be specifically highlighted here.

\section*{Acknowledgements}
We are grateful to Yasaman Bahri, Stefano Mannelli, Francesca Mignacco, Jascha Sohl-Dickstein, and Nikhil Vyas for useful conversations. We thank Clarissa Lauditi and Jacob Zavatone-Veth for comments on the manuscript. 

B.B. is supported by a Google PhD Fellowship. A.A. is supported by a Fannie and John Hertz Fellowship. C.P. is supported by NSF grant DMS-2134157, NSF CAREER Award IIS-2239780, and a Sloan Research Fellowship. This work has been made possible in part by a gift from the Chan Zuckerberg Initiative Foundation to establish the Kempner Institute for the Study of Natural and Artificial Intelligence.

\bibliography{ref}
\bibliographystyle{icml2024}

\newpage
\appendix
\onecolumn

\section{Derivation of Dynamical Model of Scaling Laws}\label{app:DMFT Equaitons}

We investigate the simplest possible model which can exhibit task-dependent time, model size and finite data bottlenecks. We therefore choose to study a linear model with projected features
\begin{align}
    f(\x) = \frac{1}{\sqrt N} \w^\top \left( \frac{1}{\sqrt M} \bm A \bm\psi(\x) \right) \ , \ y(\x) = \frac{1}{\sqrt M} \w_\star \cdot \bm\psi(\x).
\end{align}
The weights $\w$ are updated with gradient descent on a random training dataset which has (possibly) noise corrupted target values $y_\mu = y(\x_\mu) + \sigma \epsilon_\mu$. This leads to the following gradient flow dynamics
\begin{align}
    \frac{\partial}{\partial t}\w(t) = \frac{\sqrt M}{P \sqrt{N}} \sum_{\mu=1}^P ( y_\mu - f_\mu) \bm A   \bm\psi_\mu = \frac{1}{\sqrt N} \bm A  \left( \frac{1}{P} 
 \sum_{\mu=1}^P \bm\psi_\mu \left[ \bm\psi_\mu^\top  \left(\w_\star - \frac{1}{\sqrt N} \bm A^\top \w \right) + \sqrt{M} \sigma \epsilon_\mu \right] \right).
\end{align}
We introduce the variable $\bm v^0 = \w_\star - \frac{1}{\sqrt N} \bm A^\top \w$ to represent the residual error of the learned weight vector. This residual error has the following dynamics:
\begin{align}
    \partial_t  \v^0(t) = - \left( \frac{1}{N} \bm A^\top \bm A \right) \left[ \left( \frac{1}{P} \bm\Psi^\top \bm\Psi \right) \v^0(t) + \frac{\sigma }{\alpha \sqrt M} \bm\Psi^\top \bm\epsilon \right] .
\end{align}
The entries of each matrix are treated as random with $\Psi^\mu_k \sim \mathcal{N}(0, \lambda_k)$ and $A_{jk} \sim \mathcal{N}(0,1)$. 
To study the dynamical evolution of the test error $\mathcal{L}(t) = \frac{1}{M} \v^0(t)^\top \bm\Lambda \v^0(t) + \sigma^2$, we introduce the sequence of vectors
\begin{align}
    \v^1(t) &= \frac{1}{\sqrt M} \bm\Psi \v^0(t) + \sigma \bm\epsilon \ , \ \v^2(t) = \frac{1}{\alpha \sqrt M} \bm\Psi^\top \bm v^1(t) \nonumber
    \\
    \bm v^3(t) &= \frac{1}{\sqrt M} \bm A \bm v^2(t) \ , \ \v^4(t) = \frac{1}{ \nu  \sqrt{M}} \bm A^\top \v^3(t).
\end{align}
The train and test losses can be computed from the $\bm v^0$ and $\bm v^1$ fields
\begin{align}
    \hat{\mathcal{L}}(t) = \frac{1}{P} \sum_{\mu=1}^P v^1_\mu(t)^2 \ , \ \mathcal{L}(t) = \frac{1}{M} \sum_{k=1}^M \lambda_k v^0_k(t)^2 + \sigma^2.
\end{align}
In the next section, we derive a statistical description of the dynamics in an appropriate asymptotic limit using dynamical mean field theory methods. 

\subsection{DMFT Equations for the Asymptotic Limit}

Standard field theoretic arguments such as the cavity or path integral methods can be used to compute the effective statistical description of the dynamics in the limit of large $M,N,P$ with fixed ratios $\alpha = P/M$ and $\nu = \frac{N}{M}$ (see Appendix \ref{app:field_th_deriv}). This computation gives us the following statistical description of the dynamics.

\begin{equation}
    \begin{aligned}
    &v^1(t) = u^1(t) + \frac{1}{\alpha} \int ds R_{0,2}(t,s) v_1(s) + \sigma \epsilon \ , \ u^1(t) \sim \mathcal{GP}(0,C_0) \ , \ \epsilon \sim \mathcal{N}(0,1),
    \\
    &v^2_k(t) = u^2_k(t) + \lambda_k \int ds R_1(t,s) v^0_k(s) \ , \ u^2_k(t) \sim \mathcal{GP}\left(0, \frac{1}{\alpha} \lambda_k C_1 \right)  ,
    \\
    &v^3(t) = u^3(t) + \frac{1}{\nu} \int ds R_{2,4}(t,s) v^3(s) \ , \ u^3(t) \sim \mathcal{GP}\left(0, C_2 \right) , 
    \\
    &v^4_k(t) = u^4_k(t) + \int ds R_3(t,s) v^2_k(s) \ , \ u^4_k(t) \sim \mathcal{GP}\left(0, \frac{1}{\nu} C_3 \right) ,
    \\
    &\partial_t v^0_k(t) = - v^4_k(t).
\end{aligned}
\end{equation}
The correlation and response functions obey
\begin{equation}
    \begin{aligned}
    C_0(t,s) &= \frac{1}{M} \sum_k \lambda_k \left< v^0_k(t) v^0_k(s) \right> \ , \ C_1(t,s) = \left< v^1(t) v^1(s) \right> \ , \ C_{2}(t,s) = \frac{1}{M} \sum_{k=1}^M \left< v_k^2(t) v_k^2(s) \right> \nonumber
    \\
    R_{0,2}(t,s) &= \frac{1}{M} \sum_k \lambda_k \left< \frac{\delta v^0_k(t)}{\delta u^2_k(s)} \right> \ , \ R_{2,4}(t,s) = \frac{1}{M} \sum_k \left< \frac{\delta v^2_k(t)}{\delta u^4_k(s)} \right> \nonumber
    \\
    R_1(t,s) &= \left< \frac{\delta v^1(t)}{\delta u^1(s)} \right> \ , \ R_3(t,s) = \left< \frac{\delta v^3(t)}{\delta u^3(s)} \right>
\end{aligned}
\end{equation}
These equations are exact in the joint proportional limit for any value of $\alpha, \nu$.

\subsection{Closing the Equations for the Order Parameters}\label{app:closed_eqns_order_params}
Though we expressed the dynamics in terms of random fields, we stress in this section that all of the dynamics for the correlation and response functions close in terms of integro-differential equations. To shorten the expression, we will provide the expression for $\beta = 0$, but momentum can easily be added back by making the substitution $\partial_t \to \partial_t + \beta \partial^2_t$.

First, our closed integral equations for the response functions are
\begin{align}
    &R_{0,2,k}(t,s) = -\int dt' \Theta(t-t') R_3(t',s) - \lambda_k \int dt' dt'' dt''' \Theta(t-t') R_3(t',t'') R_1(t'',t''') R_{0,2,k}(t''',s) \nonumber
    \\
    &R_1(t,s) = \delta(t-s) + \frac{1}{\alpha} \int dt' R_{0,2}(t,t') R_1(t',s) \nonumber
    \\
    &R_{2,4,k}(t,s) = - \lambda_k \int dt' dt'' R_{1}(t,t') \Theta(t'-t'') - \lambda_k \int dt' dt'' dt''' R_1(t,t') \Theta(t'-t'') R_3(t'',t''') R_{2,4,k}(t''', s)   \nonumber
    \\
    &R_3(t,s) = \delta(t-s) + \frac{1}{\nu} \int dt' R_{2,4}(t,t') R_3(t',s) \nonumber
    \\
    &R_{0,2}(t,s) = \frac{1}{M} \sum_k \lambda_k R_{0,2,k}(t,s) \ , \ R_{2,4}(t,s) = \frac{1}{M} \sum_k R_{2,4,k}(t,s)
\end{align}
We note that these equations imply causality in all of the response functions since $R(t,s) = 0$ for $t < s$. Once these equations are solved for the response functions, we can determine the correlation functions, which satisfy
\begin{align}
    \partial_{ts}^2 C_{0,k}(t,s) = &- \lambda_k \int dt' dt'' R_3(t,t') R_1(t',t'') \partial_s C_{0,k}(t'',s) \nonumber
    \\
    &- \lambda_k \int ds' R_3(s,s') R_1(s',s'') \partial_t C_{0,k}(t,s'')  \nonumber
    \\
    &+ \lambda_k^2 \int dt' dt'' ds' ds'' R_3(t,t') R_1(t',t'') R_3(s,s') R_1(s',s'') C_{0,k}(t'',s'') \nonumber
    \\
    &- (w_k^\star)^2\delta(t)\delta(s) - \frac{1}{\nu} C_3(t,s) - \frac{1}{\alpha} \int dt' ds' R_3(t,t') R_3(s,s') C_1(t',s') \nonumber
    \\
    C_1(t,s) = &\int dt' R_1(t,t') R_1(s,s') C_0(t',s')  \nonumber
    \\
    C_{2,k}(t,s) = &-\lambda_k \int dt' dt'' R_1(t,t') R_3(t',t'') C_{2,k}(t'',s)  - \lambda_k \int ds' ds'' R_1(s,s') R_3(s',s'') C_2(t,s'') \nonumber
    \\
    &+ \lambda_k^2 \int dt' dt'' ds' ds''  R_1(t,t') R_3(t',t'') R_1(s,s') R_3(s',s'') C_{2,k}(t'',s'')  \nonumber
    \\
    C_3(t,s) = &\int dt' ds' R_3(t,t') R_3(s,s') C_2(t',s')
\end{align}
Solving these closed equations provide the complete statistical characterization of the limit. The test and train losses are given by the time-time diagonal of $C_0(t,t), C_1(t,t)$. 

\subsection{Time-translation Invariant (TTI) Solution to Response Functions}\label{app:TTI_response_fns}

From the structure of the above equations, the response functions are time-translation invariant (TTI) since they are only functionals of TTI $\delta(t-s)$ Dirac-Delta function and $\Theta(t-s)$ Heaviside step-function. As a consequence, we write each of our response functions in terms of their Fourier transforms
\begin{align}
    R(t,s) = R(t-s) = \int_{-\infty}^\infty \frac{d\omega}{2\pi} \ e^{i \omega (t-s)}  \mathcal{R}(\omega) .
\end{align}
Using the fact that
\begin{align}
    \delta(\tau) = \int \frac{d\omega}{2\pi} e^{i\omega \tau} \ , \ \Theta(\tau) = \lim_{\epsilon \to 0^+} e^{-\epsilon \tau}\Theta(\tau) = \lim_{\epsilon \to 0^+} \int \frac{d\omega}{2\pi} \frac{e^{i\omega \tau}}{\epsilon + i\omega}
\end{align}
We will keep track of the regulator $\epsilon$ and consider $\epsilon \to 0^+$ at the end of the computation. The resulting DMFT equations for the response functions have the following form in Fourier space
\begin{align}
    &\mathcal{R}_{1}(\omega) = 1 + \frac{1}{\alpha} \mathcal{R}_{2,4}(\omega) \mathcal{R}_{1}(\omega) \nonumber
    \\
    &\mathcal{R}_{3}(\omega) = 1 + \frac{1}{\nu}\mathcal{R}_{2,4}(\omega) \mathcal{R}_{3}(\omega) \nonumber
    \\
    &\mathcal{R}_{0,2}(\omega) = - \frac{1}{M} \sum_k \frac{ \lambda_k }{\epsilon + i\omega + \lambda_k \mathcal{R}_{1}(\omega) \mathcal{R}_{3}(\omega)} \mathcal{R}_{3}(\omega) \nonumber
    \\
    &\mathcal{R}_{2,4}(\omega) = - \frac{1}{M}\sum_k \frac{\lambda_k }{\epsilon + i\omega  + \lambda_k\mathcal{R}_1(\omega)\mathcal{R}_3(\omega) }\mathcal{R}_1(\omega)
\end{align}
where $\epsilon \to 0$ will be taken after. Combining these equations, we arrive at the simple set of coupled equations
\begin{align}
   \mathcal{R}_1(\omega) &= 1 - \frac{1}{P}\sum_k \frac{\lambda_k \mathcal{R}_3(\omega)\mathcal{R}_1(\omega) }{\epsilon + i\omega   + \lambda_k\mathcal{R}_1(\omega)\mathcal{R}_3(\omega)} \nonumber
    \\
   \mathcal{R}_3(\omega) &= 1 -  \frac{1}{N}\sum_k \frac{\lambda_k \mathcal{R}_1(\omega) \mathcal{R}_{3}(\omega)}{\epsilon + i\omega +   \lambda_k\mathcal{R}_1(\omega)\mathcal{R}_3(\omega)}
\end{align}
After solving these equations for all $\omega$, we can invert the dynamics of $v^0_k(t)$ to obtain its Fourier transform
\begin{align}
    v^0_k(\omega) &= \mathcal{H}_k(\omega) \left[ w_k^\star - {u}^4_k(\omega) - \lambda_k \mathcal{R}_3(\omega) {u}^2_k(\omega) \right] \nonumber
    \\
    \mathcal{H}_k(\omega) &\equiv \frac{1}{\epsilon + i\omega    + \lambda_k \mathcal{R}_1(\omega) \mathcal{R}_3(\omega)}
\end{align}
where we defined the transfer functions $\mathcal{H}_k(\omega)$. From this equation, we can compute $v^0_k(t)$ through inverse Fourier-transformation and then compute the correlation function to calculate the test error. An interesting observation is that the response functions $\mathcal{R}_1(\omega), \mathcal{R}_3(\omega)$ alter the pole structure in the transfer function, generating $\nu,\alpha$ dependent timescales of convergence.

\subsection{Fourier Representations for Correlation Functions} \label{app:fourier_correlation}

While the response functions are TTI, the correlation functions transparently are not (if the time-time diagonal $C_0(t,t)$ did not evolve, then the loss $\mathcal L(t)$ wouldn't change!).  We therefore define the need to define the double Fourier transform $\mathcal{C}(\omega,\omega')$ for each correlation function $C(t,s)$
\begin{align}
     \mathcal{C}(\omega,\omega') = \int dt ds \  e^{-i\omega t - i \omega' s} C(t,s) \  , \ C(t,s) = \int \frac{d\omega }{2\pi} \frac{d\omega'}{2\pi} e^{i\omega t + i\omega' s} \mathcal{C}(\omega, \omega')
\end{align}
Assuming that all response functions and transfer functions $\mathcal H_k$ have been solved for, the correlation functions satisfy the closed set of linear equations.
\begin{align}
    \mathcal C_0(\omega, \omega') &= \frac{1}{M} \sum_k \lambda_k  \mathcal{H}_k(\omega) \mathcal{H}_k(\omega') \left [ (w^\star_k)^2 + \frac{1}{\nu}  \mathcal{C}_3(\omega,\omega') + \frac{1}{\alpha} \lambda_k \mathcal{R}_3(\omega) \mathcal{R}_3(\omega') \mathcal{C}_1(\omega,\omega') \right] \nonumber
    \\
    \mathcal{C}_1(\omega,\omega') &=  \mathcal{R}_1(\omega) \mathcal{R}_{1}(\omega')  \mathcal{C}_0(\omega,\omega')  \nonumber
    \\
    \mathcal{C}_2(\omega,\omega') &= \frac{1}{M} \sum_k \lambda_k \mathcal{H}_k(\omega)\mathcal{H}_k(\omega')\left[  \frac{1}{\alpha} (i\omega) (i\omega') \mathcal C_1(\omega,\omega') + \lambda_k \mathcal{R}_1(\omega) \mathcal{R}_{1}(\omega') \left( (w^\star_k)^2 + \frac{1}{\nu} \mathcal C_3(\omega,\omega') \right) \right]  \nonumber
    \\
    \mathcal{C}_3(\omega,\omega') &= \mathcal{R}_3(\omega) \mathcal{R}_{3}(\omega') \mathcal{C}_2(\omega,\omega') 
\end{align}

These equations can be efficiently solved for all pairs of $\omega, \omega'$ after the response functions have been identified. Then one can take an inverse Fourier transform in both indices.

\section{Field Theoretic Derivation of DMFT Equations}\label{app:field_th_deriv}

In this section, we derive the field theoretic description of our model. We will derive this using both the Martin-Siggia-Rose (MSR) path integral method \cite{martin1973statistical} and the dynamical cavity method. For a recent review of these topics in the context of neural networks, see \cite{helias2020statistical}. 

\subsection{Statistical Assumptions for DMFT}

The DMFT that we derive in the next few sections requires some assumptions on the structure of $\bm A$ and $\bm \Psi$. To carry out the classic MSR path integral computation, we assume that the entries of both matrices are Gaussian with mean zero and covariance
\begin{align}
    \left< A_{ij} A_{kl} \right> = \delta_{ik} \delta_{jl}  \ , \ \left< \Psi_{\mu k} \Psi_{\nu l} \right> = \delta_{\mu\nu} \delta_{kl} \lambda_k .
\end{align}
These are sufficient conditions for the DMFT description to hold and we will take them as our primary assumptions. However, we note that these restrictions are not strictly necessary and can be relaxed. In general, a more flexible cavity derivation in Appendix \ref{app:cavity} shows that independent entries from any well behaved distribution which admits a central limit theorem for sums of independent draws would also have the same DMFT description of the proportional limit. Prior works on DMFT of M-estimators with random data have demonstrated universality for any data matrix $\bm\Psi$ with a covariance that has bounded spectral norm \cite{gerbelot2022rigorous}.

\subsection{Path Integral Derivation}

With the MSR formalism, we evaluate the moment generating functional for the field $\bm v^0(t)$:
\begin{equation}
    \begin{aligned}
    Z[\{ \bm j(t) \}] &= \left< \int \mathcal D \v^0(t) \, \delta\left(\dot \v^0(t) + \frac{1}{N P} \A^\top \A \bm \Psi^\top \bm \Psi \v^0(t)\right)  \exp\left( \int dt \ \bm j(t) \cdot \bm v^0(t) \right) \right>_{\bm A, \bm \Psi}.
    \end{aligned}
\end{equation}

Note that at zero source, we have the important identity that
\begin{equation}
    Z[0]  = 1.
\end{equation}


We insert a Dirac delta functions to enforce the definitions of each of the fields $\{ \bm v^1 , \bm v^2, \bm v^3 ,\bm v^4 \}$ as in equation \ref{eq:v_defns}.
\begin{equation}\label{eq:grand}
    \begin{aligned}
        Z[\{ \bm j(t) \}] = \int &\mathcal D[\v^0, \dots, \v^4, \hat \v^1 \dots \hat \v^4] \, \delta(\dot \v^0 + \v^4) \, \exp\left( \int dt \ \bm j(t) \cdot \bm v^0(t) \right) \\
        &\times \left< \exp\left[  i \int dt \left[ \hat \v_1(t)  \cdot \left( \v^1(t) - \frac{1}{\sqrt M} \bm \Psi \v^0(t) \right) + \hat \v_2(t) \cdot \left( \v^2(t) - \frac{1}{\alpha \sqrt M} \bm \Psi^\top \v^1(t) \right) \right] \right] \right>_{\bm \Psi} \\
         &\times \left<\exp\left[ i \int dt \left[ \hat \v_3(t) \cdot \left( \v^3(t) - \frac{1}{\sqrt M} \bm \A \v^2(t) \right) + \hat \v_4(t) \cdot \left( \v^4(t) - \frac{1}{\nu \sqrt M} \bm \A^\top \v^3(t) \right) \right]  \right]\right>_{\A}.
    \end{aligned}
\end{equation}
At this stage we can add sources $\tilde{\bm j}$ for each $\hat {\bm v_i}$ variable, yielding a $Z[\bm j(t), \tilde{\bm j}(t)]$. Interpreting each source as modification of the respective evolution equation, we see that this modified moment-generating function remains equal to unity at any value of $\tilde {\bm j}$, $Z[0, \tilde{\bm j}(t)] = 1$. As a consequence, all correlation functions consisting only of $\hat \v^i$ variables vanish. See \cite{crisanti2018path} for further details and a worked example.

We now average over the sources of disorder. We assume that the entries of $\bm A$ are i.i.d. with mean zero and variance 1. In the proportional limit, we can replace the entries of $\bm A$ as a draw from a Gaussian $\mathcal N(0, 1)$ by appealing to Gaussian equivalence. We furhther justify this in the cavity derivation in the next section. This allows us to evaluate the averages over the matrix $\bm A$. 
\begin{equation}\label{eq:A_avg}
\begin{aligned}
    &\left< \exp\left( - \frac{i}{\sqrt M} \text{Tr} \bm A^\top \int dt \left[ \bm\hat{\v}^3(t) \v^2(t)^\top + \nu^{-1} \bm v^3(t) \bm{\hat \v}^4(t)^\top \right] \right) \right>_{\A} 
    \\
    &= \exp\left( - \frac{1}{2} \int dt ds [ \hat{\bm v}^3(t) \cdot \hat{\bm v}^3(s) \underbrace{\frac{1}{M} \v^2(t) \cdot \v^2(s)}_{C_2(t, s)}  + \nu^{-1} \hat{\bm v}^4(t) \cdot \hat{\bm v}^4(s) \underbrace{\frac{1}{N} \v^3(t) \cdot \v^3(s)}_{C_3(t, s)}  ]   \right) 
    \\
    & \quad \times \exp\left( - \int dt ds \ \underbrace{ \frac{1}{N} \hat{\bm v}^3(t) \cdot \v^3(s)}_{i R_3(s, t)} \, \v^2(t) \cdot \hat{\bm v}^4(s)\right).
\end{aligned}
\end{equation}
Similarly, we can calculate the averages over the data, which enters via the design matrices $\bm \Psi$. Again in this proportional limit we can invoke Gaussian equivalence on $\bm \Psi$ to have it take the form $\bm \Psi \sim \bm \Phi \bm \Lambda^{1/2}$ where $\bm \Phi$ has entries drawn from a unit normal. Taking the average then gives us 
\begin{equation}\label{eq:Psi_avg}
\begin{aligned}
&\left< \exp\left( - \frac{i}{\sqrt M} \text{Tr} \ \bm \Psi^\top \int dt \left[ \bm\hat{\v}^1(t) \v^0(t)^\top + \alpha^{-1} \bm v^1(t) \bm{\hat \v}^2(t)^\top \right] \right) \right>_{\bm \Psi}
    \\
    &= \exp\left( - \frac{1}{2} \int dt ds [ \hat{\bm v}^1(t) \cdot \hat{\bm v}^1(s) \underbrace{\frac{1}{M} \v^0(t) \cdot \bm \Lambda \v^0(s)}_{C_0(t, s)}  + \alpha^{-1} \hat{\bm v}^2(t) \cdot \bm \Lambda  \hat{\bm v}^2(s) \underbrace{\frac{1}{P} \v^1(t) \cdot \v^1(s)}_{C_1(t, s)}  ]   \right), 
    \\
    & \quad \times \exp\left( - \int dt ds \ \underbrace{\frac{1}{P} \hat{\bm v}^1(t) \cdot \v^1(s)}_{i R_1(s, t)} \,  \v^0(t) \cdot \bm \Lambda \cdot \hat{\bm v}^2(s) \right).
\end{aligned}
\end{equation}
We now insert delta functions for following bracketed terms: $C_0, C_1, C_2, C_3$ and $R_1, R_3$ using the following identity (e.g. for $C_0$ at times $s, t$):
\begin{equation}
    1 = \int \frac{d C_0(s, t) d \hat C_0(s, t)}{4 \pi i M^{-1}} \exp\left[\frac12 M  \int dt ds\, \hat C_0(t,s) \left(C_0(t,s) - \frac{1}{M} \v^0(t) \cdot \bm \Lambda \v^0(s) \right) \right].
\end{equation}
Here the $\hat C_i, \hat R_i$ integrals are taken over the imaginary axis. This yields a moment generating function (here we'll take $\bm j = 0$):
\begin{equation}
\begin{aligned}
    Z = \int \mathcal D [C_1, \hat C_1, \dots] \exp\left[M S[C_0, C_1, C_2, C_3, R_1, R_3, \hat C_0, \hat C_1, \hat C_2, \hat C_3, \hat R_1, \hat R_3] \right].
\end{aligned}
\end{equation}
The constraint that $Z = 1$ means that $S=0$ at the saddle point. $S$ here is given by:
\begin{equation}
\begin{aligned}
    S[\dots] &= \frac12 \int dt ds \Big[ \hat C_0(t,s) C_0(t,s)  + \alpha \hat C_1(t,s) C_1(t,s) + \hat C_2(t,s) C_2(t,s) + \nu \hat C_3(t,s) C_3(t,s) \Big] \\
    & \quad + \int dt ds \Big[ - R_1(t,s) \hat R_1(s, t) - R_3(t,s) \hat R_3(s,t)  \Big]   \\
    & \quad + \alpha \log \mathcal        Z_1 + \nu \log \mathcal Z_3 + \frac{1}{M} \sum_k \log \mathcal Z_{0, 2, 4; k}.
\end{aligned}
\end{equation}
We have chosen to take $\hat R_i(s, t)$ to have a different sign and $s,t$ ordering convention than the $\hat C_i$ to simplify our notation later on.  We have also used that Equations \eqref{eq:A_avg}, \eqref{eq:Psi_avg} factorize over their respective indices, so each $\mathcal Z$ is a partition function over a single index. The individual $\mathcal Z_i$ are given by:
\begin{equation}
\begin{aligned}
    \mathcal Z_1 = \int \mathcal  D[v^1, \hat v^1] \exp \left[ i \int dt ds \left(\delta(t -s ) - \alpha^{-1} \hat R_1(s, t) \right) v^1(t) \hat v^1(s)   \right] \\
   \times \exp\left[ - \frac12 \int dt ds (\hat v^1(t) \hat v^1(s) C_0(t, s) + v^1(t) v^1(s) \hat C_1(t, s) ) \right],
\end{aligned}
\end{equation}
\begin{equation}
\begin{aligned}
    \mathcal Z_3 = \int \mathcal D[v^3, \hat v^3] \exp\left[i \int dt ds \left(\delta(t -s ) - \nu^{-1} \hat R_3(s, t) \right) v^3(t) \hat v^3(s) \right] \\
     \times \exp \left[- \frac12 \int dt ds (\hat v^3(t) \hat v^3(s) C_2(t, s) + v^3(t) v^3(s) \hat C_3(t, s) )  \right],    
\end{aligned}
\end{equation}
\begin{equation}
    \begin{aligned}
    \mathcal Z_{0, 2, 4; k} &= \int \mathcal D[v^{0, 2, 4}, \hat v^{0,2,4}] \exp\left[- \frac12 \int dt ds \left( \alpha^{-1} \lambda_k \hat v^2_k(t) \hat v^2_k(s) C_1(t, s) +  \nu^{-1} \hat v^4_k(t) \hat v^4_k(s) C_3(t, s) \right) \right]\\
    & \qquad \times \exp\left[-\frac12 \int dt ds \left(\lambda_k v^0_k(t) v^0_k(s) \hat C_0(t, s) + v^2_k(t) v^2_k(s) \hat C_2(t, s) + v^4_k(t) v^4_k(s) \hat C_4(t, s) \right) \right]\\
    & \qquad \times \exp\left[-i \int dt ds \left(R_3(t, s) v^2_k(s) \hat v^4_k(t) + \lambda_k R_1(t, s) v^0_k(s) \hat v^2_k(t)  \right) \right].
    \end{aligned}
\end{equation}

In the large $M$ limit we evaluate this integral via saddle point. The saddle point equations give:
\begin{equation}
\begin{aligned}
    C_0(t,s) &= \frac{1}{M} \sum_k \lambda_k \left< v^0_k(t) v^0_k(s) \right>
    \\
    C_i(t, s) &= \langle v^\ell(t) v^\ell(s) \rangle, \quad \ell = \{1, 2, 3, 4\}\\
    \\
    R_1(t, s) &= -i \langle v^1(t) \hat v^1(s) \rangle\\
    R_3(t, s) &= -i \langle v^3(t) \hat v^3(s) \rangle\\
    \hat R^1(t, s) &= - i \frac{1}{M} \sum_k \lambda_k \hat v_k^2(t) v_k^0(s) \equiv R_{0, 2}(t, s)\\
    \hat R^3(t, s) &= - i \frac{1}{M} \sum_k \lambda_k \hat v_k^4(t) v_k^2(s) \equiv R_{2, 4}(t, s).
\end{aligned}
\end{equation}
Here $\langle \cdot \rangle$ denotes an average taken with respect to the statistical ensemble given by the corresponding partition function $\mathcal Z_{i}$. Lastly, the saddle point equations for the $\hat{C}_i(t,s)$ variables are all quadratic functions of the variables $\{ \hat{v}^0, \hat{v}^1, \hat{v}^2 , \hat{v}^3 \}$ which vanish under the average defined by $\mathcal Z$ \cite{helias2020statistical}. Following the discussion below Equation \ref{eq:grand}, we take $\hat C_i(t,s) = 0$, which will enforce $\langle \hat v_i(t) \hat v_i(s) \rangle =0$ and lead to the correct dynamical equations.

To evaluate the remaining, we can integrate out the $\hat v^i$ variables. First let us look at $\mathcal Z_1$. Using the Hubbard-Stratonovich trick we can write the action in terms linear in $\hat v^1$. This gives
\begin{equation}\label{eq:single_site1}
\begin{aligned}
    \mathcal Z_1 &= \int \mathcal D[v^1, \hat v^1, u^1] \exp\left[i \int dt ds\, \hat v^1(t) \left[ \delta(t- s) (v^1(s) - u(s)) - \alpha^{-1} \hat R_1(t, s) v^1(s) \right] \right]\\
    & \quad \times \exp\left[- \frac12 \int dt ds \, u(t) u(s) C^{-1}_0(t, s)  + v^1(t) v^1(s) \hat C_1(t, s) ) \right]
\end{aligned}
\end{equation}
We now replace $\hat C_1$ by its saddle point value of $0$ and $\hat R_1$ by $R_{0, 2}$. Integrating over $\hat v$ gives a delta function:
\begin{equation}
    v^1(t) = u^1(t) + \frac{1}{\alpha} \int ds R_{0, 2}(t, s) v_1(s), \quad u^1(t) \sim \mathcal{GP}(0, C_0).
\end{equation}

Analogously for $v^3$ we get
\begin{equation}
    v^3(t) = u^3(t) + \frac{1}{\nu} \int ds R_{2, 4}(t, s) v^3(s), \quad  u^3(t) \sim \mathcal{GP}(0, C_2)
\end{equation}
For $\mathcal Z_{0, 2, 4; k}$ after replacing $\hat C_0, \hat C_2, \hat C_4$ with their saddle point values we get:
\begin{equation}
\begin{aligned}
\mathcal Z_{0, 2, 4; k}&= \exp\left[- \frac12 \int dt ds \left( \alpha^{-1} \lambda_k \hat v^2_k(t) \hat v^2_k(s) C_1(t, s) +  \nu^{-1} \hat v^4_k(t) \hat v^4_k(s) C_3(t, s) \right) \right]\\
    & \qquad \times \exp\left[- \int dt ds \left(i R_3(t, s) v^2_k(s) \hat v^4_k(t) + i \lambda_k R_1(t, s) v^0_k(s) \hat v^2_k(t)  \right) \right]
\end{aligned}
\end{equation}
Using the same Hubbard-Stratonovich trick on $\hat v^2_k$ gives:
\begin{equation}
    v^2_k(t) = u^2_k(t) + \lambda_k \int ds R_1(t, s) v_k^0(s), \quad  u^2_k(t) \sim \mathcal{GP}\left(0, \frac{1}{\alpha} \lambda_k C_1 \right).
\end{equation}
On $\hat v^4_k$ we similarly get:
\begin{equation}
    v^4_k(t) = u^4_k(t) + \int ds R_3(t, s) \cdot v^2_k(s), \quad   u^4_k(t) \sim \mathcal{GP}\left(0, \frac{1}{\nu} C_3 \right)
\end{equation}
Lastly, the equations of motion for $v^0_k$ in terms of $v^4_k$ are known:
\begin{equation}
    \partial_t v^0_k(t) = - v^4_k(t).
\end{equation}

One can easily add momentum by replacing $\partial_t v^0_k(t)$ with $(\beta \partial^2_t + \partial_t) v^0_k(t)$ without changing anything else about the derivation.

\subsubsection{Interpretation of the Response Functions}\label{app:resp_defn}

Following \cite{crisanti2018path, helias2020statistical}, we can understand the $\langle \hat v^a(t) v^b(s) \rangle$ correlators by adding in the single-site moment generating function (e.g. Equation \eqref{eq:single_site1}) a source  $\tilde j^b(s)$ that couples to $\hat v^b$ at time $s$. As in the discussion below equation \ref{eq:grand}, differentiating $\langle v^a(t) \rangle$ with respect to that source corresponds to its response to a kick in the dynamics of $v^b$ at time $s$. We denote this by:
\begin{equation}
    R_{i,j}(t,s) = \left< \frac{\delta v^i(t)}{\delta v^j(s)} \right>.
\end{equation}

\subsection{Cavity Derivation}\label{app:cavity}

The cavity derivation relies on Taylor expanding the dynamics upon the addition of a new sample or feature. We will work through each cavity step one at a time by considering the influence of a single new base feature, new sample, and new projected feature. In each step, the goal is to compute the marginal statistics of the added variables. This requires tracking the linear response to all other variables in the system. 

\paragraph{Adding a Base Feature}
Upon addition of a base feature with eigenvalue $\lambda_0$ so that there are $M+1$ instead of $M$ features $\{ v^0_k, v^2_k, v^4_k \}$ for $k \in \{ 0, 1,  ..., M \}$, we have a perturbation to both $v^1_\mu(t)$ and $v^3_n(t)$. Denote the perturbed versions of the dynamics upon addition of the $M+1$st feature as $\tilde{v}^1_\mu(t)$ and $\tilde v^3_n(t)$. At large $M$ we can use linear-response theory to relate the dynamics at $M+1$ features to the dynamics of the original $M$ feature system 
\begin{align}
    &\tilde{v}^1_\mu(t) \sim {v}^1_\mu(t) + \frac{1}{\sqrt M} \sum_{\nu=1}^P \int ds \frac{\partial v^1_\mu(t)}{\partial v^1_\nu(s)} \psi^\nu_{0} v^0_0(s)  \nonumber
    \\
    &\tilde{v}^3_n(t) \sim {v}^1_\mu(t) + \frac{1}{\sqrt M} \sum_{m=1}^N \int ds \frac{\partial v^3_n(t)}{\partial v^3_{m}(s)} A_{m 0} \ v^2_0(s)
\end{align}
The next order corrections have a subleading influence on the dynamics. Now, inserting these perturbed dynamics into the dynamics for the new $(M+1)$st set of variables $\{v^2_0(t), v^4_0(t) \}$. For $v^2_0(t)$, we have
\begin{align}
    v^2_0(t) \sim \frac{1}{\alpha \sqrt{M}} \sum_{\mu=1}^P \psi^\mu_0 v^1_\mu(t) + \frac{1}{\alpha M} \sum_{\mu,\nu=1}^P \int  ds \  \psi^\mu_0 \frac{\partial v^1_\mu(t)}{\partial v^1_\nu(s)} \psi^\nu_0 v^0_0(s)
\end{align}
There are now two key steps in simplifying the above expression in the proportional limit:
\begin{enumerate}
    \item  By the fact that the $v^1_\mu(t)$ dynamics are statistically independent of the new feature $\psi^\mu_0$, we can invoke a central limit theorem for the first term which is mean zero and variance $\mathcal{O}(1)$. 
    \item Similarly, we can invoke a law of large numbers for the second term, which has $\mathcal{O}(1)$ mean and variance on the order of $\mathcal{O}(M^{-1})$. Therefore in the asymptotic limit it can be safely approximated by its mean.
\end{enumerate} 
We note in passing that neither of these steps require the $\psi^\mu_0$ variables to be Gaussian. Thus we obtain the following asymptotic statistical description of the $v^2_0(t)$ random variable
\begin{align}
    v^2_0(t) &\sim u^2_0(t) + \int ds R_1(t,s) v^0_0(s) \nonumber \\
    u^2_0(t) &\sim \mathcal{GP}(0, \alpha^{-1} \lambda_0 C_1) \ , \ R_1(t,s) \equiv \frac{1}{P} \sum_{\mu=1}^P \left<  \frac{\partial v^1_\mu(t) }{\partial v^1_\mu(s) }\right>.
\end{align}
Following an identical argument for $v^4_0(t)$ we have
\begin{align}
    v^4_0(t) &\sim \frac{1}{\nu \sqrt M} \sum_{n=1}^M A_{n0} v^3_n(t) + \frac{1}{\nu M} \sum_{nm} \int ds  \ A_{n 0} \frac{\partial v^3_n(t)}{\partial v^3_m(s)} A_{m0} v^2_0(s) \nonumber 
    \\
    &\sim u^3_0(t) + \int ds R_3(t,s) v^2_0(s) \nonumber
    \\
    &u_0^3(t) \sim \mathcal{GP}(0, \nu^{-1} C_3) \ , \ R_3(t,s) = \frac{1}{N} \sum_{n=1}^N \left< \frac{\partial v^3_n(t)}{\partial v^3_n(s)} \right>
\end{align}

\paragraph{Adding a Sample}

Next, we can consider the influence of adding a new data point. We will aim to characterize a $P+1$ data point system in terms of the dynamics when $P$ points are present. Upon the addition of a new data point $\bm\psi^0$ the field $v_k^0(t)$ will be perturbed to $\tilde{v}_k^0(t)$. Again invoking linear response theory, we can expand the perturbed value around the $P$-sample dynamics
\begin{align}
    \tilde{v}^0_k(t) \sim v^0_k(t) + \frac{1}{\alpha \sqrt{M}} \sum_{\ell=1}^M \int ds 
    \ \frac{\partial v^0_k(t)}{\partial v^2_\ell(s)} \psi^0_\ell v^1_0(s)
\end{align}
Now, computing the dynamics of the new random variable $v^1_0(t)$
\begin{align}
    v^1_0(t) &\sim \frac{1}{\sqrt M} \sum_{k=1}^M \psi^0_k v^0_k(t) + \frac{1}{\alpha M} \int ds \sum_{k \ell} \psi^0_k \  \frac{\partial v^0_k(t)}{\partial v^2_\ell(s)} \ \psi_\ell^0 \ v^1_0(s) \nonumber
    \\
    &\sim u^1_0(t) + \frac{1}{\alpha} \int ds R_{0,2}(t,s) v^1_0(s) \nonumber
    \\
    u^1_0(t) &\sim \mathcal{GP}\left(0, \frac{1}{M} \sum_k \lambda_k C^0_k \right) \ , \ R_{0,2}(t,s) = \frac{1}{M} \sum_{k=1}^M \lambda_k \left< \frac{\partial v^0_k(t)}{\partial v^2_k(s)}  \right>
\end{align}

\paragraph{Adding a Projected Feature}

Now, we finally consider the effect of introducing a single new projected feature so that instead of $N$ we now have $N+1$ projected features. This causes a perturbation to $\{v^2_k(t)\}$ which we 
\begin{align}
    \tilde{v}^2_k(t) \sim v^2_k(t) + \frac{1}{\nu \sqrt M}  \sum_{\ell=1}^M \int ds \  \frac{\partial v^2_k(t)}{\partial v^4_\ell(s)} A_{0 \ell} \ v^3_0(s)   
\end{align}
Now, we compute the dynamics for the added variable $v^3_0(t)$
\begin{align}
    v^3_0(t) &\sim \frac{1}{\sqrt M} \sum_{k=1}^M A_{0 k} v^2_k(t) + \frac{1}{\nu M} \sum_{k \ell} \int ds  A_{0k} \frac{\partial v^2_k(t)}{\partial v^4_\ell(s)} A_{0 \ell} \ v^3_0(s)  \nonumber  
    \\
    &\sim u^3_0(t) + \frac{1}{\nu} \int ds R_{2,4}(t,s) v^3_0(s) \nonumber  
    \\
    u^3_0(t) &\sim \mathcal{GP}(0, C_2) \ , \ R_{2,4}(t,s) = \frac{1}{M} \sum_{k=1}^M \left< \frac{\partial v^2_k(t)}{\partial v^4_k(s)} \right>
\end{align}

\paragraph{Putting it all together} Now, using the information gained in the previous sections, we can combine all of the dynamics for each field into a closed set of stochastic processes. This recovers the DMFT equations of Appendix \ref{app:closed_eqns_order_params}.

\section{Final Losses (the $t \to \infty$ Limit of DMFT)}\label{app:final_value_dmft}

In this section we work out exact expressions for the large time limit of DMFT. By comparing with prior computations of the mean-field statics of this problem computed in \cite{atanasov2023onset, zavatoneveth2023learning, ruben2023learning, maloney2022solvable, simon2021eigenlearning}, we show that the large time and large $M$ limits commute, specifically that $\lim_{M,N,P \to \infty} \lim_{t \to \infty} \mathcal{L}(M,N,P, t) = \lim_{t \to \infty} \lim_{M,N,P\to\infty}  \mathcal{L}(M,N,P, t)$. We invoke the final value theorem and use the response functions as before.

\paragraph{Final Value Theorem} 

We note that for functions which vanish at $t = - \infty$, that
\begin{align}
    \lim_{\omega \to 0} i\omega \ \mathcal{H}(\omega) = -\lim_{\omega \to 0} \int_{-\infty}^\infty d\tau \left[ \frac{\partial}{\partial \tau} e^{-i\omega \tau} \right]H(\tau) = \lim_{\omega \to 0}  \int_{-\infty}^\infty d\tau \left[\frac{\partial}{\partial \tau} H(\tau) \right] e^{-i\omega \tau} = \lim_{\tau \to \infty} H(\tau)
\end{align}
where we invoked integration by parts and used the assumption that $\lim_{\tau \to - \infty} H(\tau) = 0$, a condition that is satisfied for the correlation and response functions in our theory. We can therefore use the identity $\lim_{\tau \to \infty} H(\tau) = \lim_{\omega \to 0} i \omega \mathcal H(\omega)$ to extract the final values of our order parameters. 

\begin{align}
\lim_{\tau \to \infty} H_k(\tau) =   \lim_{\omega \to 0}
   i\omega \mathcal{H}_k(\omega) = \lim_{\omega \to 0} \frac{1}{1 + \lambda_k (i\omega)^{-1} \mathcal R_1(\omega) \mathcal R_3(\omega)}.
\end{align}

We also need to invoke a similar relationship for the final values of the correlation functions
\begin{align}
    \lim_{t,s \to \infty} C(t,s) = \lim_{\omega,\omega' \to 0} (i\omega) (i\omega') \  \mathcal{C}(\omega, \omega') \nonumber
    \\
    \mathcal{C}(\omega,\omega') = \int dt \int ds \  e^{-i(\omega t+\omega' s)} C(t,s)
\end{align}
where $\mathcal C$ is the two-variable Fourier transform. The final value of the test loss is $\lim_{t \to \infty} \mathcal{L}(t) = \lim_{t,s \to \infty} C_0(t,s)$.

\subsection{General Case (Finite $\nu,\alpha$)}
Before working out the solution to the response functions, we note that the following condition is always satisfied
\begin{align}
    \nu(1 - \mathcal{R}_{3}(\omega))  = \alpha(1-\mathcal{R}_{1}(\omega)) .
\end{align}
For $\nu = \alpha$, this equation implies that $\mathcal R_1 = \mathcal R_3$. For $\nu \neq \alpha$, we can have either $\mathcal{R}_1 \to 0$ or $\mathcal{R}_3 \to 0$ but not both. We consider each of these cases below. 

\paragraph{Over-parameterized Case $\nu > \alpha$:}

In this case, the response function $\mathcal{R}_1 \sim \mathcal{O}( i \omega)$ as $\omega \to 0$ and $\mathcal R_3 \sim 1 - \frac{\alpha}{\nu}$ as $\omega \to 0$. We thus define
\begin{align}
    r \equiv \lim_{\omega \to 0} (i\omega)^{-1} \mathcal{R}_1(\omega) \mathcal{R}_3(\omega)  
\end{align}
Using the equation which defines $\mathcal{R}_1$, we find that the variable $r$ satisfies the following relationship at $\omega \to 0$
\begin{align}
    \alpha  = \frac{1}{M} \sum_{k} \frac{\lambda_k r }{1 + \lambda_k r }
\end{align}
After solving this implicit equation, we can find the limiting value of $i\omega \mathcal{H}_k(\omega)$ as
\begin{align}
    H_k^\infty = \lim_{\tau \to \infty} {H}_k(\tau) = \lim_{\omega \to 0 } i\omega \mathcal{H}_k(\omega) = \frac{1}{1 + \lambda_k r}
\end{align}
Next, we can work out the scaling of the correlation functions in the limit of low frequency. We define the following limiting quantities based on a scaling analysis performed on our correlation functions for small $\omega$
\begin{equation}
\begin{aligned}
    C_{0}^\infty &\equiv \lim_{t,s \to \infty} C_0(t,s) = \lim_{\omega,\omega'\to 0} (i\omega)(i\omega') \mathcal{C}_0(\omega,\omega') 
    \\
    C_1^\infty &\equiv \int_{0}^\infty \int_{0}^\infty dt'  ds' C_1(t',s') =  \lim_{\omega,\omega' \to 0} \mathcal{C}_1(\omega,\omega') 
    \\
    C_2^\infty &\equiv \int_{0}^\infty \int_{0}^\infty dt'  ds' C_2(t',s') = \lim_{\omega,\omega' \to 0} \mathcal{C}_2(\omega,\omega') 
    \\
    C_3^\infty &\equiv \int_{0}^\infty \int_{0}^\infty dt'  ds' C_3(t',s') = \lim_{\omega,\omega' \to 0} \mathcal{C}_3(\omega,\omega') 
\end{aligned}    
\end{equation}

These limiting quantities satisfy the closed set of linear equations
\begin{equation}
    \begin{aligned}
    &C_0^\infty = \frac{1}{M} \sum_k \lambda_k (H_k^\infty)^2 \left[ (w^\star_k)^2 + \frac{1}{\nu} C_3^\infty + \frac{1}{\alpha} \lambda_k \left(1-\frac{\alpha}{\nu} \right)^2  C_1^\infty \right] 
    \\
    &C_1^\infty = \frac{r^2}{(1-\frac{\alpha}{\nu})^2} C_0^\infty  
    \\
    &C_2^\infty = \frac{1}{\alpha M} \sum_{k} \lambda_k (H_k^\infty)^2 C_1^\infty + \frac{r^2}{(1-\frac{\alpha}{\nu})^2} \frac{1}{M} \sum_k \lambda_k^2 (H_k^\infty)^2 [(w^\star_k)^2 + \nu^{-1} C_3^\infty ] 
    \\
    &C_3^\infty = \left(1-\frac{\alpha}{\nu} \right)^2 C_2^\infty  
\end{aligned}
\end{equation}
These equations can be solved for $\{ C_0^\infty, C_1^\infty, C_2^\infty, C_3^\infty \}$. Simplifying the expressions to a two-variable system, we find
\begin{equation}
   \begin{aligned}
    C_0^\infty &= \frac{1}{M} \sum_k \lambda_k (H_k^\infty)^2 \left[ (w^\star_k)^2 + \frac{1}{\nu} \left(1-\frac{\alpha}{\nu}\right)^2 C_2^\infty + \frac{1}{\alpha} \lambda_k r^2 C_0^\infty \right] \nonumber
    \\
    C_2^\infty &= \frac{r^2}{\alpha (1-\alpha/\nu)^2 M} \sum_{k} \lambda_k (H_k^\infty)^2 C_0^\infty + \frac{r^2}{(1-\frac{\alpha}{\nu})^2} \frac{1}{M} \sum_k \lambda_k^2 (H_k^\infty)^2 \left[(w^\star_k)^2 + \frac{1}{\nu} \left( 1- \frac{\alpha}{\nu} \right)^{2} C_2^\infty \right]
\end{aligned} 
\end{equation}
This expression recovers the ridgeless limit of the replica results of \cite{atanasov2023onset, zavatoneveth2023learning} and the random matrix analysis of \cite{simon2023more}.

\paragraph{Under-parameterized Case $\nu < \alpha$:} Following the same procedure, we note that for $\nu < \alpha$ that $\mathcal{R}_3 \sim \mathcal{O}(i\omega)$ and $\mathcal{R}_1 \sim 1- \frac{\nu}{\alpha}$. We thus find the following equation for $r = \lim_{\omega \to \infty} (i\omega)^{-1} \mathcal{R}_1(\omega) \mathcal{R}_3(\omega)$. 
\begin{align}
    \nu = \frac{1}{M} \sum_k \frac{\lambda_k r}{\lambda_k r + 1}
\end{align}
where as before $H_k^\infty = \frac{1}{1+\lambda_k r}$. The analogous scaling argument for small $\omega$ gives us the following set of well-defined limiting quantities
\begin{equation}
    \begin{aligned}
    C_{0}^\infty &\equiv \lim_{t,s \to \infty} C_0(t,s) = \lim_{\omega,\omega'\to 0} (i\omega)(i\omega') \mathcal{C}(\omega,\omega') 
    \\
    C_1^\infty &\equiv \lim_{t,s \to \infty} C_1(t,s) =  \lim_{\omega,\omega' \to 0} (i\omega) (i\omega') \mathcal{C}_1(\omega,\omega') 
    \\
    C_2^\infty &\equiv \lim_{t,s \to \infty} C_2(t,s) = \lim_{\omega,\omega' \to 0} (i\omega) (i\omega') \mathcal{C}_2(\omega,\omega') 
    \\
    C_3^\infty &\equiv \int_0^\infty \int_0^\infty dt ds \  C_3(t,s) = \lim_{\omega,\omega' \to 0} \mathcal{C}_3(\omega,\omega') . 
\end{aligned}
\end{equation}
where these limiting correlation values satisfy
\begin{equation}
\begin{aligned}
    C_0^\infty &= \frac{1}{M} \sum_k \lambda_k (H_k^\infty)^2 \left[ (w^\star_k)^2 + \frac{1}{\nu} C_3^\infty + \frac{1}{\alpha} \lambda_k \frac{r^2}{(1-\frac{\nu}{\alpha})^2} C_1^\infty \right] 
    \\
    C_1^\infty &= \left( 1 - \frac{\nu}{\alpha} \right)^2 C_0^\infty 
    \\
    C_2^\infty &= \frac{1}{\alpha M} \sum_k \lambda_k (H_k^\infty)^2 C_1^\infty + \frac{1}{M} \left( 1 - \frac{\nu}{\alpha} \right)^2 \sum_k \lambda_k^2 (H_k^\infty)^2 \left[ (w^\star_k)^2 + \frac{1}{\nu } C_3^\infty \right] 
    \\
    C_3^\infty &= \frac{r^2}{(1-\frac{\nu}{\alpha})^2} C_2^\infty .
\end{aligned}    
\end{equation}
This is again a closed linear system of equations for the variables $\{ C_0^\infty,C_1^\infty, C_2^\infty, C_3^\infty  \}$. In the next section, we recover the result for kernel regression where $\nu \to \infty$ and the learning curve for infinite data $\alpha \to \infty$ with respect to model size $\nu$.

\subsection{Learning Curves for Kernel Regression $\nu,t \to \infty$}

In the $t \to \infty$ and $\nu \to \infty$ limit we recover the learning curve for kernel regression with eigenvalues $\lambda_k$. To match the notation of \cite{canatar2021spectral}, we define
\begin{align}
    \lim_{\omega \to 0} (i\omega)^{-1} \mathcal{R}_1(\omega) \equiv \alpha \kappa^{-1}
\end{align}
which generates the following self-consistent equation for $\kappa$
\begin{align}
    1 = \frac{1}{M} \sum_{k} \frac{\lambda_k}{\lambda_k \alpha + \kappa} . 
\end{align}
Plugging this into the expression for the loss, we find
\begin{align}
    &(i\omega) (i\omega') \mathcal{C}_0(\omega,\omega') \sim \frac{1}{M} \sum_k \lambda_k \frac{\kappa^2}{(\kappa + \lambda_k \alpha)^2} [ (w^\star_k)^2 + \alpha^{-1} \lambda_k  \mathcal{C}_1(\omega,\omega')  ] \nonumber
    \\
    &\mathcal{C}_1(\omega,\omega') = (i\omega) (i\omega') \alpha^2 \kappa^{-2} \mathcal{C}_0(\omega,\omega')
\end{align}
Letting $C_{\infty} \equiv \lim_{\omega,\omega'\to 0} (i\omega)(i\omega') \mathcal{C}(s,s')$, we have
\begin{align}
    C_{\infty} &= \frac{1}{M} \sum_k \lambda_k \frac{\kappa^2}{(\kappa + \lambda_k \alpha)^2}  (w^\star_k)^2 + C_{\infty} \frac{\alpha}{M} \sum_k \frac{\lambda_k^2}{(\lambda_k \alpha + \kappa)^2} \nonumber
    \\
    &= \frac{1}{1- \gamma } \sum_k \lambda_k (w^\star_k)^2  \frac{\kappa^2}{(\kappa + \lambda_k \alpha)^2}  \ , \quad \gamma = \frac{\alpha}{M} \sum_k \frac{\lambda_k^2}{(\lambda_k \alpha + \kappa)^2} 
\end{align}
The variable $\kappa$ decreases from $\left[ \frac{1}{M}\sum_k \lambda_k ,0 \right]$ as $\alpha \in [0,1]$. For $\alpha > 1$ we have $\kappa=0$. The quantity $\frac{1}{1-\gamma}$ comes from overfitting due to variance from the randomly sampled dataset.

\section{Early Time Dynamics (High-Frequency Range)}\label{app:early_time}

In this section, we explore the early time dynamical effects of this model. Similar to how the late time dynamical effects could be measured by examining the low frequency $\omega \ll 1$ part of the response and correlation functions, in this section, we analyze the high frequency components $\omega \gg 1$. We start by noting the following expansions valid near $\omega \to \infty$ 
\begin{align}
    \mathcal{R}_1(\omega) &\sim 1 - \frac{1}{\alpha (i\omega)}  \left[ \frac{1}{M} \sum_k \lambda_k \right] + \mathcal{O}(\omega^{-2}) \nonumber
    \\
    \mathcal{R}_3(\omega) &\sim 1 - \frac{1}{\nu (i\omega) } \left[ \frac{1}{M} \sum_k \lambda_k \right] + \mathcal{O}(\omega^{-2})
\end{align}
We let $c = \frac{1}{M} \sum_k \lambda_k$. These can be plugged into the transfer function for mode $k$
\begin{align}
    \mathcal{H}_k(\omega) \sim \frac{1}{i\omega + \lambda_k - c (\alpha^{-1} + \nu^{-1}) (i\omega)^{-1}   } \sim \frac{1}{i\omega + \lambda_k} \left[  1  + \frac{c \lambda_k (\alpha^{-1} + \nu^{-1})}{i\omega (i\omega + \lambda_k) }  \right] + \mathcal{O}\left( \omega^{-2} \right)
\end{align}
Performing an inverse Fourier transform, we find the following early time asymptotics
\begin{align}
    H_k(t) &\sim e^{-\lambda_k t} + c \lambda_k (\alpha^{-1} + \nu^{-1}) \int \frac{d\omega}{2\pi} \frac{e^{i\omega t}}{i\omega(i\omega + \lambda_k)^2} \nonumber
    \\
    &= e^{-\lambda_k t} - c \lambda_k (\alpha^{-1} + \nu^{-1}) \frac{\partial}{\partial \lambda_k } \int \frac{d\omega}{2\pi} \frac{e^{i\omega t}}{i\omega(i\omega + \lambda_k)} \nonumber
    \\
    &= e^{-\lambda_k t} - c \lambda_k (\alpha^{-1} + \nu^{-1}) \frac{\partial}{\partial \lambda_k } \left[ \frac{1}{\lambda_k} - \frac{1}{\lambda_k} e^{-\lambda_k t} \right] \nonumber
    \\
    &= e^{-\lambda_k t} + \frac{c (\alpha^{-1} + \nu^{-1})}{\lambda_k} \left[ 1-e^{-\lambda_k t} - \lambda_k t \ e^{-\lambda_k t} \right]
\end{align}
We see from this expression that the early time corrections always scale as $1/\alpha$ or $1/\nu$ and that these corrections build up over time. We also note that in this picture, $H_k(t)$ is minimized in the limit of large model and large data $\alpha,\nu \to \infty$ (limited data and limited model size strictly harm performance). A similar expansion can be performed for all of the correlation functions $\mathcal{C}(\omega,\omega')$ with $\omega,\omega' \gg 1$ which also give leading corrections which scale as $1/\alpha$ and $1/\nu$.

\section{Buildup of Overfitting Effects}\label{app:buildup_overfitting}

In this section, we derive a formula for the gap between test loss $\mathcal L(t)$ and train loss $\hat{\mathcal L}(t)$. We start from the following formula 
\begin{align}
    v_1(t) = u_1(t) + \frac{1}{\alpha} \int_0^t ds R_{0,2}(t,s) v_1(s)  
\end{align}
Moving the $v_1(t)$ term to the other side, and using the fact that $\left< u_1(t) u_1(s) \right> = C_1(t,s)$, we find the following relationship between train and test loss
\begin{align}
    \mathcal{L}(t) &= \left< u_1(t) u_1(t) \right> = \left< v_1(t) v_1(t) \right> - \frac{2}{\alpha} \int_0^t dt' R_{0,2}(t,t') \left< v_1(t) v_1(t') \right> \nonumber
    \\
    &+ \frac{1}{\alpha^2} \int_0^t dt' \int_0^t ds' R_{0,2}(t,t') R_{0,2}(t,s') \left< v_1(t') v_1(s') \right> \nonumber
    \\
    &= \hat{\mathcal L}(t) - \frac{2}{\alpha} \int_0^t dt' R_{0,2}(t,t') C_1(t,t') + \frac{1}{\alpha^2} \int_0^t dt' \int_0^t ds' R_{0,2}(t,t') R_{0,2}(t,s') C_1(t',s') .
\end{align}
To get a sense of these expressions at early and late timescales, we investigate the Fourier transforms at high $\omega \gg 1$ and low $\omega \ll 1$ frequencies respectively. 

\subsection{High Frequency Range / Early Time}
The relationship between Fourier transforms at high frequencies $\omega \gg 1$ is 
\begin{align}
    \mathcal C_0(\omega,\omega') &= \frac{1}{\mathcal{R}_1(\omega) \mathcal{R}_1(\omega')} \  \mathcal C_1(\omega,\omega') \sim \mathcal C_1(\omega,\omega') + \frac{c}{\alpha(i\omega') } \mathcal C_1(\omega,\omega') +  \frac{c}{\alpha(i\omega') } \mathcal C_1(\omega,\omega') + \mathcal{O}( (i\omega)^{-2} + (i\omega')^{-2} )
\end{align}
where $c = \frac{1}{M} \sum_k \lambda_k$. Taking a Fourier transform back to real time gives us the following early time differential equation for the test-loss train loss gap
\begin{align}
    \partial^2_{ts} [C_0(t,s) - C_1(t,s) ] = \partial^2_{ts} C_1(t,s) + \frac{c}{\alpha} (\partial_t + \partial_s) C_1(t,s) .
\end{align}
The above equation should hold for early times. We note that $C_0(t,t) - C_1(t,t) = \mathcal L(t) - \mathcal{\hat L}(t)$ exactly recovers the test-train gap.

\subsection{Low Frequency Range/Late Time}

At late time/low frequency, as we showed in Appendix \ref{app:final_value_dmft}, the behavior of the $C_1$ correlation function depends on whether the model is over-parameterized or under-parameterized. In the overparameterized case, the asymptotic train loss is zero while the asymptotic test loss is nonzero. In the underparameterized case, we have a limiting value for both the test and train loss which can be computed from the expressions in Appendix \ref{app:final_value_dmft}.

\section{Timescale/Eigenvalue Density Interpretation}\label{app:timescale_density}
\begin{figure}[h]
    \centering
    \subfigure[$N=50, P=100$]{\includegraphics[width=0.42\linewidth]{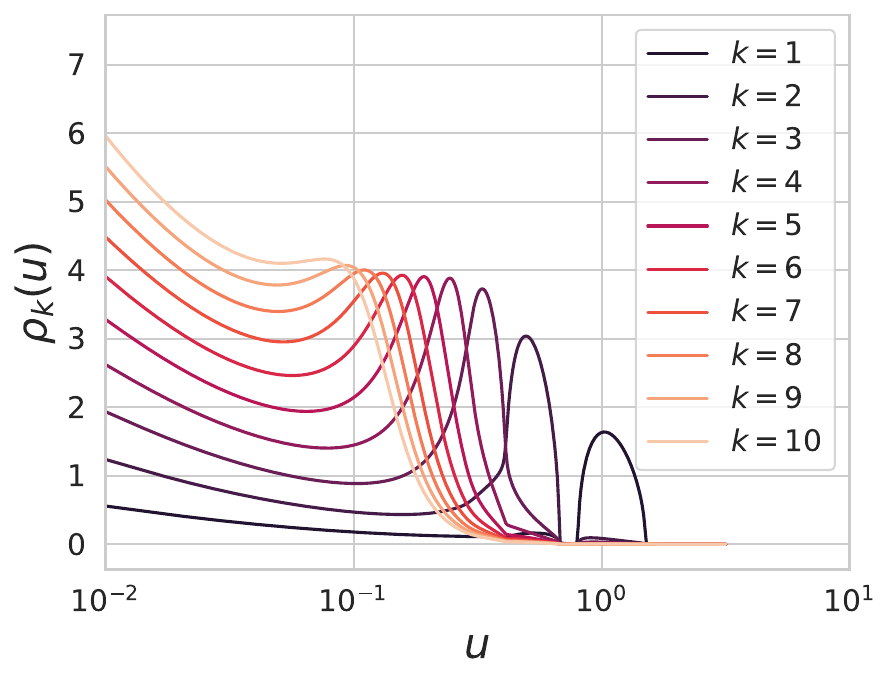}}
     \subfigure[$N=10000, P=10000$]{\includegraphics[width=0.42\linewidth]{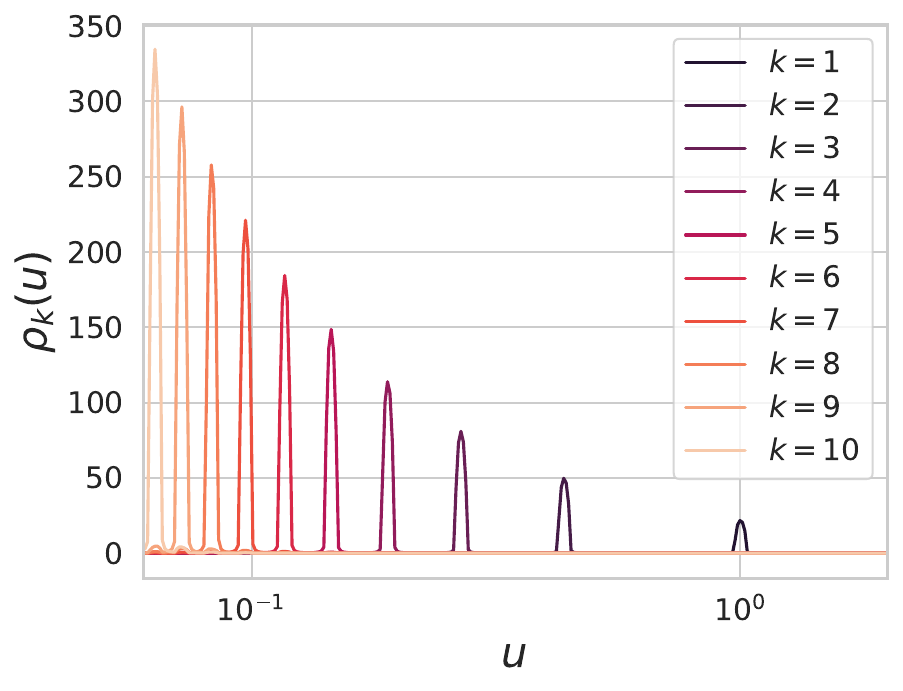}}
    \caption{Timescale (eigenvalue) densities for each transfer function $H_k(\tau)$ with power law features with $b=1.2$. For limited $N,P$ there is a significant spread of timescales for each mode. For $N,P \to \infty$ the density converges to a Dirac mass at $u = \lambda_k$. }
    \label{fig:timescale_densities}
\end{figure}

We can use an alternative interpretation of the Fourier transforms derived in previous sections to obtain the timescale density for the dynamics. Since this is a linear model defined by an effective matrix $\frac{d}{dt} \v^0 = - \left( \frac{1}{N} \bm A^\top \bm A \right) \left( \frac{1}{P} \bm \Psi^\top \bm\Psi \right) \v^0$, this is equivalent to computing the eigenvalue density. We start by expanding the transfer function for mode $k$ in the basis of exponentials
\begin{align}
    H_k(t) = \int_0^\infty du \ \rho_k(u) \ e^{- u t}  . 
\end{align}
We allow for Dirac-delta masses at $u = 0$ which correspond to the constant (unlearnable) components. Next, we note that the Fourier transform has the form
\begin{align}
    \mathcal H_k(\omega) = \int_{-\infty}^\infty dt \ e^{-i \omega t} H_k(t) = \int_0^\infty du \ \rho_k(u) \ \int_{-\infty}^\infty dt \ e^{- (u+i\omega) t} = \int_0^\infty du \ \frac{\rho_k(u)}{i\omega + u} .
\end{align}
We can recover the density $\rho_k(s)$ by using the Sokhotski–Plemelj theorem $\frac{1}{\pi} \text{Im} \frac{1}{- i\epsilon + u-s} = \delta(u-s)$ which gives us
\begin{align}
   \rho_k(u) = \lim_{\epsilon \to 0} \frac{1}{\pi} \text{Im} \ \mathcal H_k(i u - \epsilon) . 
\end{align}
This allows us to interpret the spread of timescales from the random sampling of data and the random projection $\bm A$. In the limit of $\alpha , \nu \to \infty$ we have $\rho_k(s) = \delta(s-\lambda_k)$ but for finite $\alpha,\nu$ the density spreads out. We visualize these densities for power law features in Figure \ref{fig:timescale_densities}.

\subsection{Recovering the Marchenko-Pastur Law from DMFT Response Functions}

To further illustrate the validity of this perspective, we show that it is possible to recover known random matrix theory results using this technique. To illustrate this, we study the case where $\lambda_k = 1$ and take $\nu \to \infty$. In this case, we have the coupled equations
\begin{align}
    \mathcal H(\omega) = \frac{1}{i\omega + \mathcal R_1(\omega)} \ , \ \mathcal R_1(\omega) = 1 - \frac{1}{\alpha} \mathcal R_1(\omega) \mathcal H(\omega)
\end{align}
Combining these equations gives the single equation
\begin{align}
    \mathcal H(\omega) &= \frac{1}{i\omega + \frac{\alpha}{\alpha + \mathcal H(\omega)}} 
    \ , \ \implies   i \omega \mathcal H(\omega)^2 + (\alpha i\omega + \alpha - 1)\mathcal H(\omega) - \alpha = 0 \nonumber
    \\
    \mathcal H(\omega) &= -  \frac{1}{2 i\omega} \left[ (\alpha i\omega + \alpha - 1) + \sqrt{ (\alpha i\omega + \alpha - 1)^2 + 4 i \omega \alpha }  \right]
\end{align}
Now, evaluating this expression at $i\omega = -s - i\epsilon$ gives
\begin{align}
     \mathcal H(is -\epsilon) &=  \frac{1}{2 (s+i\epsilon)} \left[ ( - \alpha s - i\alpha \epsilon + \alpha - 1) + \sqrt{ (- \alpha s - \alpha i\epsilon + \alpha - 1)^2 - 4 (s+i\epsilon) \alpha }  \right]
\end{align}
The radical has an imaginary solution in the $\epsilon \to 0$ limit provided that 
\begin{align}
    s \in [s_{-}, s_{+}] \ , \ s_{\pm} = \left( 1 \pm \frac{1}{\sqrt \alpha} \right)^2
\end{align}
In this interval $[s_{-},s_+]$, the density $\rho(s) = \lim_{\epsilon \to 0} \frac{1}{\pi} \text{Im} \mathcal H(is - \epsilon)$ has the form
\begin{align}
    \rho(s) = \frac{\alpha \sqrt{(s-s_{-})(s-s_+)}}{2\pi s} \ , \ s \in [s_{-}, s_+]
\end{align}
which is precisely the bulk of the Marchenko-Pastur law. 

\section{Non-Proportional (Dimension-Free) Limit}\label{app:non_prop_limit}

We can imagine a situation where the original features are already infinite dimensional ($M \to \infty$ is taken first). This would correspond more naturally to the connection between infinite dimensional RKHS's induced by neural networks at infinite width \cite{bordelon2020spectrum, canatar2021spectral, cheng2022dimension}. Further, we will assume a trace class kernel $K(\x,\x') = \bm\psi(\x) \cdot \bm\psi(\x')$ for the base features $\bm\psi$ which diagonalizes over the data distribution $p(\x)$ as
\begin{align}
    \int K(\x,\x')  \phi_k(\x') p(\x') d\x' = \lambda_k \phi_k(\x) \quad , \quad   \sum_{k=1}^\infty \lambda_k < \infty  .
\end{align}
As before, we are concerned with the test and train losses
\begin{align}
    \mathcal{L}(t) = \sum_{k=1}^\infty \lambda_k v^0_k(t)^2 \ , \ \mathcal{\hat L}(t) = \frac{1}{P} \sum_{\mu=1}^P v^1_\mu(t)^2 .
\end{align}
The appropriate scaling of our four fields of interest in this setting are 
\begin{align}
    \v^1(t) = \bm\Psi \v^0(t) \ , \ \v^2(t) = \frac{1}{P} \bm\Psi^\top \v^1(t) \nonumber
    \\
    \v^3(t) = \A \v^2(t) \ , \ \v^4(t) = \frac{1}{N} \A^\top \v^3(t) .
\end{align}
Following the cavity argument given in the previous section, we can approximate the the correlation and response functions as concentrating to arrive at the following field description of the training dynamics
\begin{equation}
\begin{aligned}
    &\partial_t v^0_k(t) = - v^4_k(t) 
    \\
    &v^1(t) = u^1(t) + \frac{1}{P} \int ds \ R_{0,2}(t,s) v^1(s) \ , \ u^1(t) \sim \mathcal{GP}(0,C_0) 
    \\
    &v^2_k(t) = u^2_k(t) + \lambda_k \int ds \ R_1(t,s) v^0_k(s) \ , \ u^2_k(t) \sim \mathcal{GP}\left(0, \frac{1}{P} \lambda_k C^1 \right)  
    \\
    &v^3(t) = u^3(t) + \frac{1}{N} \int ds \ R_{2,4}(t,s) v^3(s) \ , \ u^3(t) \sim \mathcal{GP}(0, C_2) 
    \\
    &v^4_k(t) = u^4_k(t) + \int ds \ R_3(t,s) v^2_k(s)  \ , \ u^4_k(t) \sim \mathcal{N}\left(0, \frac{1}{N} C_3 \right) .
\end{aligned}
\end{equation}
which are exactly the same equations as in the proportional limit except with the substitution $\nu \to N$ and $\alpha \to P$. The correlation and response functions have the form
\begin{equation}
\begin{aligned}
    &C_0(t,s) = \sum_{k=1}^\infty  \lambda_k  \left< v^0_k(t) v^0_k(s) \right> \ , \  C_1(t,s) = \left< v^1(t) v^1(s) \right> 
    \\
    &C_2(t,s) = \sum_{k=1}^\infty v^2_k(t) v^2_k(s) \ , \ C_3(t,s) = \left< v^3(t) v^3(s) \right> 
    \\
    &R_{0,2}(t,s) = \sum_{k=1}^\infty \lambda_k \left< \frac{\delta v^0_k(t)}{\delta u^2_k(s)} \right> \ , \ R_1(t,s) = \left< \frac{\delta v^1(t)}{\delta u^1(s)} \right> 
    \\
    &R_{2,4}(t,s) = \sum_{k=1}^\infty  \left< \frac{\delta v^2_k(t)}{\delta u^4_k(s)} \right> \ , \ R_3(t,s) = \left< \frac{\delta v^3(t)}{\delta u^3(s)} \right> 
\end{aligned} 
\end{equation}
which will all be $\mathcal{O}(1)$ under this scaling.

\section{Effect of Ensembling and Bagging on Dynamics}\label{app:ens_bag}

\subsection{What Does/Doesn't Concentrate in the DMFT Limit?}

To help gain insight into bias and variance decompositions, we first provide a short primer on which entities concentrate over random draws of matrices $\bm A$ and $\bm \Psi$. For any distinct randomly sampled system, the following objects will always be the same in the asymptotic limit
\begin{enumerate}
    \item The response functions $\{ R_{0,2}(t,s), R_1(t,s), R_{2,4}(t,s), R_{3}(t,s)  \}$
    \item The correlation functions $\{ C_{i}(t,s) \}_{i \in \{1,2,3,4\} }$.
    \item The train and test loss dynamics
\end{enumerate}

While the above quantities behave as concentrating or "self-averaging" random variables, many important quantities are not the same across different realizations of $\{ \bm A, \bm \Psi \}$. For example, 
\begin{enumerate}
    \item The (random) entries of the vectors $\{\bm v^0(t), \bm v^1(t), \v^2(t), \v^3(t), \v^4(t)\} $. 
    \item The Gaussian sources $\{ u^1(t), u^2(t), u^3(t), u^4(t) \}$ which appear in the large system size limit. 
\end{enumerate}
In particular, the first implies that the model outputs $f(\x)$ will generally depend on random variations across datasets or model initializations. This means that we can consider drawing multiple realizations of, for example, projection matrices $\{ \bm A_{e} \}_{e=1}^E$ and then training $E$ separate models using each of them. Averaging these vectors gives us
\begin{align}
    \bar{\v}^0(t) = \frac{1}{E} \sum_{e=1}^E \v^0_e(t)
\end{align}
This operation will intuitively "average out" noise from the random projection matrices $\bm A_e$ and in the limit of infinite ensembling $E \to \infty$ will completely eliminate it. 

\subsection{Definition of Bias and Variance}\label{app:bias_var_defn}

We adopt the language of the fine-grained bias-variance decomposition in \cite{adlam2020understanding}. There, a given learned function generally depends on both the dataset $\mathcal D$ and initialization seed $\theta_0$. We write this as $f_{\mathcal D, \mathcal \theta_0}$. The role of random initialization is played by the $\bm A$ matrix in our setting. For a given function, its variance over datasets and its \textit{variance} over initializations are respectively given by
\begin{align}
    \mathrm{Var}_{\mathcal D} f &\equiv \mathbb E_{\mathcal D} (f_{\mathcal D, \mathcal \theta_0} - \mathbb E_{\mathcal D} [f_{\mathcal D, \mathcal \theta_0}] )^2 \\
    \mathrm{Var}_{\theta_0} f &\equiv \mathbb E_{\theta_0} (f_{\mathcal D, \mathcal \theta_0}- \mathbb E_{\theta_0} [f_{\mathcal D, \mathcal \theta_0}] )^2
\end{align}
Here $ \mathbb E_{\mathcal D} [f_{\mathcal D, \mathcal \theta_0}] $ and $ \mathbb E_{\theta_0} [f_{\mathcal D, \mathcal \theta_0}] $ can be viewed as \textit{infinitely bagged} or \textit{infinitely ensembled} predictors respectively. The \textit{bias} of a function over datasets or initializations is given by the test error of $ \mathbb E_{\mathcal D} [f_{\mathcal D, \mathcal \theta_0}] $, $ \mathbb E_{\theta_0} [f_{\mathcal D, \mathcal \theta_0}] $  respectively. The irreducible bias is given by $ \mathbb E_{\mathcal D, \theta_0} [f_{\mathcal D, \mathcal \theta_0}] $. 

\subsection{Derivation}

In this section, we consider the effect of ensembling over $E$ random initial conditions and bagging over $B$ random datasets. We let $\bm v^0_{e,b}(t)$ represent the weight discrepancy for model $e$ on dataset $b$. Here $e$ runs from $1$ to $E$ and $b$ runs from $1$ to $B$. The $(e, b)$th vector has dynamics:
\begin{align}
    \frac{d}{dt} \v^0_{e,b}(t) = - \left( \frac{1}{N} \bm A_e^\top \bm A_e \right) \left( \frac{1}{N} \bm\Psi_b^\top \bm\Psi_b \right) \v^0_{e,b}(t).
\end{align}
Ensembling and bagging would correspond to averaging these $\v^0$s over these $EB$ systems
\begin{align}
    \bar{\bm v}^0(t) = \frac{1}{E B} \sum_{e=1}^E \sum_{b=1}^B \v^0_{e,b}(t) .
\end{align}
The key vectors to track for this computation are
\begin{align}
    \v^1_{e,b}(t) &= \frac{1}{\sqrt M} \bm\Psi_b \v^0_{e,b}(t) + \sigma \bm\epsilon_b \ , \ \v^2_{e,b}(t) = \frac{1}{\alpha \sqrt M} \bm\Psi_b^\top \bm v^1_{e,b}(t) \nonumber
    \\
    \bm v^3_{e,b}(t) &= \frac{1}{\sqrt M} \bm A_e \bm v^2_{e,b}(t) \ , \ \v^4_{e,b}(t) = \frac{1}{ \nu  \sqrt{M}} \bm A_e^\top \v^3_{e,b}(t).
\end{align}

We can further show that the $ \v^0_{e,b}$ and $\v^0_{e',b'}$ have response functions that decouple across $e, b$. Intuitively, giving the dynamical system $e, b$ a kick should not alter the trajectory of the separate $e', b'$ dynamical system, even if they share disorder $\{ \bm\Psi, \bm A \}$. The DMFT description of the proportional limit yields the following integral equations for the $v$ fields:
\begin{align}
    &\partial_t v_{e,b,k}^0(t) = - v^4_{e,b,k}(t) \nonumber
    \\
    &v^1_{e,b}(t) = u^1_{e,b}(t) + \frac{1}{\alpha} \int ds R_{0,2}(t,s) v^1_{e,b}(s) \nonumber
    \\
    &v^2_{e,b,k}(t) = u^2_{e,b,k}(t) + \lambda_k \int ds R_1(t,s) v^0_{e,b,k}(s) \nonumber
    \\
    &v^3_{e,b}(t) = u^3_{e,b}(t) + \frac{1}{\nu} \int ds R_{2,4}(t,s) v^1_{e,b}(s) \nonumber
    \\
    &v^4_{e,b,k}(t) = u^4_{e,b,k}(t) + \int ds R_3(t,s) v^2_{e,b,k}(s).
\end{align}
Here, the response functions $R$ are to be computed within a single system. In what follows, we will use $\langle \cdot \rangle$ to denote averages over the disorder, and explicitly write out any averages over the ensemble members and datasets.


The Gaussian variables in the DMFT have the following covariance
\begin{align}
    &\left< u^1_{e,b}(t) u^1_{e',b'}(s) \right> = \delta_{b,b'} C^{0}_{e,e'}(t,s) \nonumber
    \\
    &\left< u^2_{e,b,k}(t) u^2_{e',b',k}(s) \right> = \delta_{b,b'} \frac{\lambda_k}{\alpha} C_{1,e,e'}(t,s) \nonumber
    \\
    &\left< u^3_{e,b}(t) u^3_{e',b'}(s) \right> = \delta_{e,e'}  C_{2,b,b'}(t,s) \nonumber
    \\
    &\left< u^4_{e,b,k}(t) u^4_{e',b',k}(s) \right> = \delta_{e,e'} \frac{1}{\nu} C_{3,b,b'}(t,s)
\end{align}
The covariances above $C_{0,e,e'}, C_{1,e,e'} , C_{2,b,b'}, C_{3,b,b'}$ allow for different ensemble or dataset index but not both. We will use $C_0, C_1, C_2, C_3$ etc to represent the correlation functions \textit{within a single system}. For instance, $C^0_{e,e'}(t,s) = \frac{1}{M} \sum_k \lambda_k \left< v^0_{e, b}(t) v^0_{e',b}(s) \right>$ while $C^0 = \frac{1}{M} \sum_k \lambda_k \left< v^0_{e, b}(t) v^0_{e,b}(s) \right>$. The correlation function of interest is thus
\begin{align}
    \mathcal{C}_{0,k,e,e'}(\omega,\omega') &= \mathcal{H}_k(\omega)  \mathcal{H}_k(\omega') \left[ (w_k^\star)^2 + \frac{1}{\nu} \delta_{e,e'} \mathcal C_{3}(\omega,\omega') + \frac{\lambda_k}{\alpha}  \mathcal C_{1 , e , e'}(\omega,\omega') \right]  \nonumber
    \\
    \mathcal{C}_{1,e,e'}(\omega,\omega') &= \mathcal{R}_{1}(\omega) \mathcal{R}_1(\omega') \mathcal C^0_{e,e'}(\omega,\omega') \nonumber
    \\
    \mathcal{C}_{2,b,b',k}(\omega,\omega') &= (i\omega)(i\omega') \mathcal{H}_k(\omega) \mathcal{H}_k(\omega') \frac{\lambda_k}{\alpha} \delta_{b,b'}  \mathcal C_{1}(\omega, \omega') + \lambda_k^2 \mathcal{R}_1(\omega)\mathcal{R}_1(\omega') \mathcal H_k(\omega) \mathcal{H}_k(\omega') \left[ (w_k^\star)^2 + \frac{1}{\nu} \mathcal C_{3,b,b'}(\omega,\omega') \right] \nonumber
    \\
    \mathcal{C}_{3,b,b'}(\omega,\omega') &= \mathcal{R}_3(\omega)\mathcal{R}_3(\omega) \mathcal C_{2,b,b'}(\omega,\omega') 
\end{align}
We can combine the first two equations and the second two equations to identify the structure of the cross-ensemble and cross-dataset (across-system) correlations in terms of the marginal (within-system) correlation statistics
\begin{align}
    \mathcal{C}_{0,e,e'} (\omega,\omega') &= \frac{1}{1-\gamma_0(\omega,\omega')} \frac{1}{M} \sum_k \lambda_k \mathcal{H}_k(\omega)  \mathcal{H}_k(\omega') \left[ (w_k^\star)^2 + \frac{1}{\nu} \delta_{e,e'} \ \mathcal C_{3}(\omega,\omega') \right] \nonumber
    \\
    \gamma_0(\omega,\omega') &= \frac{1}{\alpha M} \sum_k \lambda_k^2 \mathcal{H}_k(\omega)  \mathcal{H}_k(\omega') \mathcal{R}_1(\omega) \mathcal{R}_1(\omega') \mathcal R_3(\omega) \mathcal R_3(\omega') \nonumber
    \\
    \mathcal{C}_{2,b,b'} (\omega,\omega') &= \frac{1}{1-\gamma_2(\omega,\omega')} \frac{1}{M} \sum_k \lambda_k \mathcal H_k(\omega) \mathcal H_k(\omega')  \left[ \mathcal{R}_1(\omega)\mathcal{R}_1(\omega') \lambda_k (w_k^\star)^2 + \frac{1}{\alpha} (i\omega) (i\omega') \delta_{b,b'} \ 
    \mathcal C_{1}(\omega,\omega') \right]   \nonumber
    \\
    \gamma_2(\omega,\omega') &= \frac{1}{ \nu M} \sum_k \lambda_k^2 \mathcal{H}_k(\omega)  \mathcal{H}_k(\omega') \mathcal{R}_1(\omega) \mathcal{R}_1(\omega') \mathcal R_3(\omega) \mathcal R_3(\omega') 
\end{align}

These equations give the necessary cross-ensemble and cross-dataset correlations. Now we can consider the effect of ensembling and bagging on the dynamics. To do so, consider the Fourier transform of the bagged-ensembled error $\bar{v}^0_k(t) = \frac{1}{EB} \sum_{eb} v^0_{k,e,b}(t)$, which has the Fourier transform 
\begin{align}
    {\bar v}^0_{k}(\omega) = \mathcal{H}_k(\omega) \left[ w^\star_k - \frac{1}{E B} \sum_{e, b} \left( u^4_{e,b,k}(\omega) + \mathcal{R}_3(\omega)  u^2_{e,b,k}(\omega) \right) \right]
\end{align}
Computing the correlation function for this bagged-ensembled field random variable, we find
\begin{align}
    \left< {\bar v}^0_{k}(\omega) {\bar v}^0_{k}(\omega') \right> &= \mathcal{H}_k(\omega) \mathcal{H}_k(\omega') \left[ (w^\star_k)^2 + \frac{1}{\nu E^2 B^2} \sum_{e,e',b,b'} \delta_{e,e'} \mathcal C_{3,b,b'}(\omega,\omega') +  \frac{\lambda_k \mathcal R_3(\omega)\mathcal R_3(\omega')}{\alpha E^2 B^2}  \sum_{e e' b b'} \delta_{b,b'} \mathcal C_{1,e,e'}(\omega,\omega')  \right] \nonumber
    \\
    &= \mathcal{H}_k(\omega) \mathcal{H}_k(\omega') \left[ (w^\star_k)^2 + \frac{1}{\nu E B^2} \sum_{b,b'} \mathcal C_{3,b,b'}(\omega,\omega') +  \frac{\lambda_k \mathcal R_3(\omega)\mathcal R_3(\omega')}{\alpha E^2 B^2}  \sum_{e e' } \mathcal C_{1,e,e'}(\omega,\omega')  \right] \nonumber
    \\
    &= \mathcal{H}_k(\omega) \mathcal{H}_k(\omega') (w^\star_k)^2 \nonumber
    \\
    &+ \frac{1}{\nu E} \frac{\mathcal{H}_k(\omega) \mathcal{H}_k(\omega') \mathcal{R}_3(\omega)\mathcal{R}_3(\omega')\mathcal{R}_1(\omega)\mathcal{R}_1(\omega')}{1-\gamma_2(\omega,\omega')} \   \left[ \frac{1}{M} \sum_{\ell} \lambda_\ell^2 (w_\ell^\star)^2 \mathcal H_{\ell}(\omega) \mathcal H_\ell(\omega') \right]  \nonumber
    \\
    &+ \frac{1}{\nu \alpha E B} \frac{\mathcal{H}_k(\omega) \mathcal{H}_k(\omega') \mathcal R_3(\omega) \mathcal R_3(\omega') \mathcal{C}_1(\omega,\omega') (i\omega)(i\omega')}{1-\gamma_2(\omega,\omega')} \left[ \frac{1}{M}  \sum_\ell \lambda_\ell \mathcal{H}_\ell(\omega)\mathcal{H}_\ell(\omega') \right] \nonumber 
    \\
    &+  \frac{\lambda_k}{\alpha B} \frac{\mathcal{H}_k(\omega) \mathcal{H}_k(\omega') \mathcal{R}_1(\omega)\mathcal{R}_1(\omega') \mathcal R_3(\omega) \mathcal R_3(\omega')}{1-\gamma_0(\omega,\omega')} \left[ \frac{1}{M} \sum_\ell \lambda_\ell  (w^\star_\ell)^2 \mathcal{H}_\ell(\omega)\mathcal{H}_\ell(\omega') \right] \nonumber
    \\
    &+ \frac{\lambda_k}{\alpha \nu E B} \frac{\mathcal{H}_k(\omega) \mathcal{H}_k(\omega') \mathcal R_1(\omega) \mathcal R_1(\omega') \mathcal R_3(\omega) \mathcal R_3(\omega') 
\mathcal{C}_3(\omega,\omega')}{1-\gamma_0(\omega,\omega')} \left[ \frac{1}{M} \sum_{\ell} \lambda_\ell \mathcal{H}_\ell(\omega) \mathcal{H}_\ell(\omega')  \right]
\end{align}
The first term is the irreducible bias for mode $k$ which is the loss for mode $k$ when the learned function is averaged over all possible datasets and all possible projections. We see that the second term scales as $\frac{1}{\nu E}$ which will persist even if $B \alpha \to \infty$. Similarly, there is a term that is order $\frac{1}{\alpha B}$ which will persist even if $\nu E \to \infty$. Lastly, there are two terms which depend on both $B,E$. This is similar to the variance that is explained by the interaction of the dataset and the random projection \cite{adlam2020understanding}. The test loss is then a Fourier transform of the above function
\begin{align}
    \bar{\mathcal L}(t) = \frac{1}{M} \sum_k \lambda_k \left< \bar{v}^0_k(t)^2 \right> . 
\end{align}
If $E,B \to \infty$, then we obtain the stated \textit{irreducible bias} of the main paper
\begin{align}
    \lim_{E,B \to \infty} \bar{\mathcal L}(t) = \frac{1}{M} \sum_k \lambda_k (w^\star_k)^2 H_k(t)^2 .
\end{align}
This is the error of the mean output function over all possible datasets and random projections of a certain size. 

\begin{figure}[h]
    \centering
    \subfigure[$\nu = 0.25$]{\includegraphics[width=0.45\linewidth]{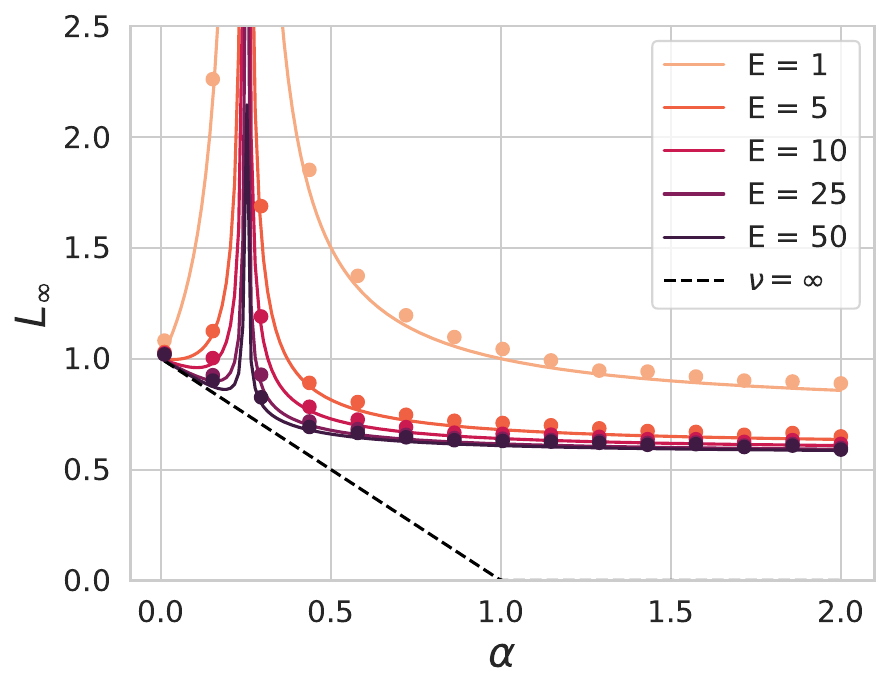}}
    \subfigure[$\nu = 0.75$]{\includegraphics[width=0.45\linewidth]{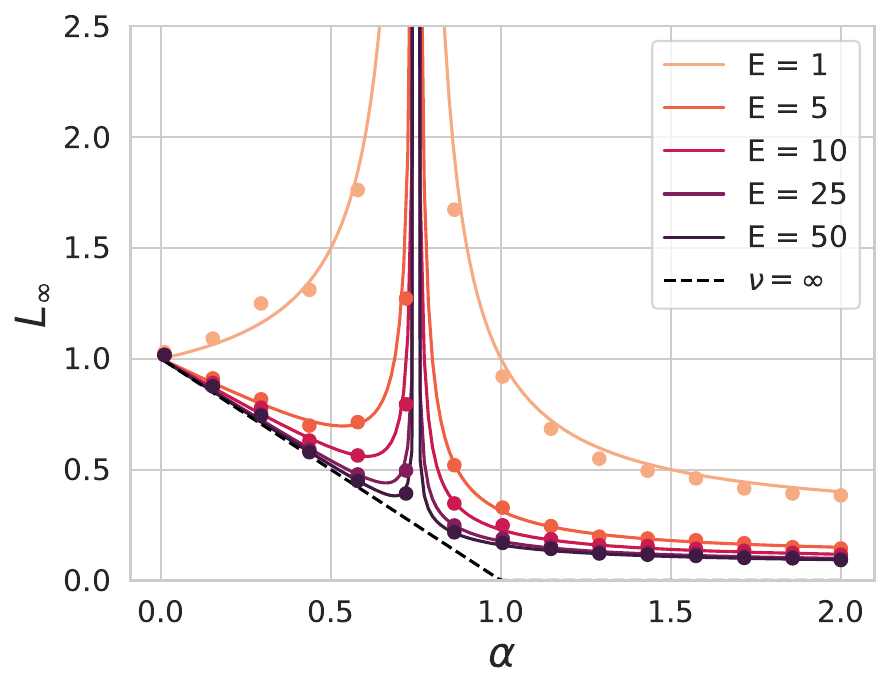}}
    \caption{The infinite time limit of the loss when ensembling with isotropic features $\lambda_k = 1$ recovers prior results on ensembling and double descent \cite{d2020double, adlam2020understanding}. There is an overfitting peak (double descent) at $\alpha = \nu$. In the \textit{overparameterized regime} where  $\alpha < \nu$, the infinite ensembled model matches the performance of the $\nu \to \infty$ limit. This is because the bias is limited by dataset size rather than model size. In the \textit{underparameterized regime} $\alpha > \nu$, the infinite ensembled model \textit{does not} achieve the loss of the infinite model due to a bias limited by $\nu$.  }
    \label{fig:enter-label}
\end{figure}

\subsection{Ensembling is Not Always Compute Optimal}

For a compute budget $C = N E t$, we find that ensembling does not provide as much benefit as increasing the size of the model. From the results in the last section, we note that ensembling reduces the variance. For this section, we consider the $P \to \infty$ limit. We let $\mathcal{B}(N,t)$ represent the bias and $\mathcal V(N,t)$ represent the variance within a single ensemble. The loss at fixed compute then takes the form
\begin{align}
    \mathcal{L}(\nu,C,t) = \mathcal{B}(\nu,t) + \frac{1}{\nu E} \mathcal{V}(\nu,t).
\end{align}
For any $\nu$ which satisfies the condition that
\begin{align}
    \frac{\partial}{\partial \nu} \mathcal B(\nu, t) \leq 0 \ , \ \frac{\partial}{\partial \nu} \mathcal V(\nu, t) \leq 0
\end{align}
we have that ensembling is strictly dominated by increasing $\nu$. 





\section{White Bandlimited Model}\label{app:white}

\begin{figure}[h]
    \centering
    \includegraphics[width=0.6\linewidth]{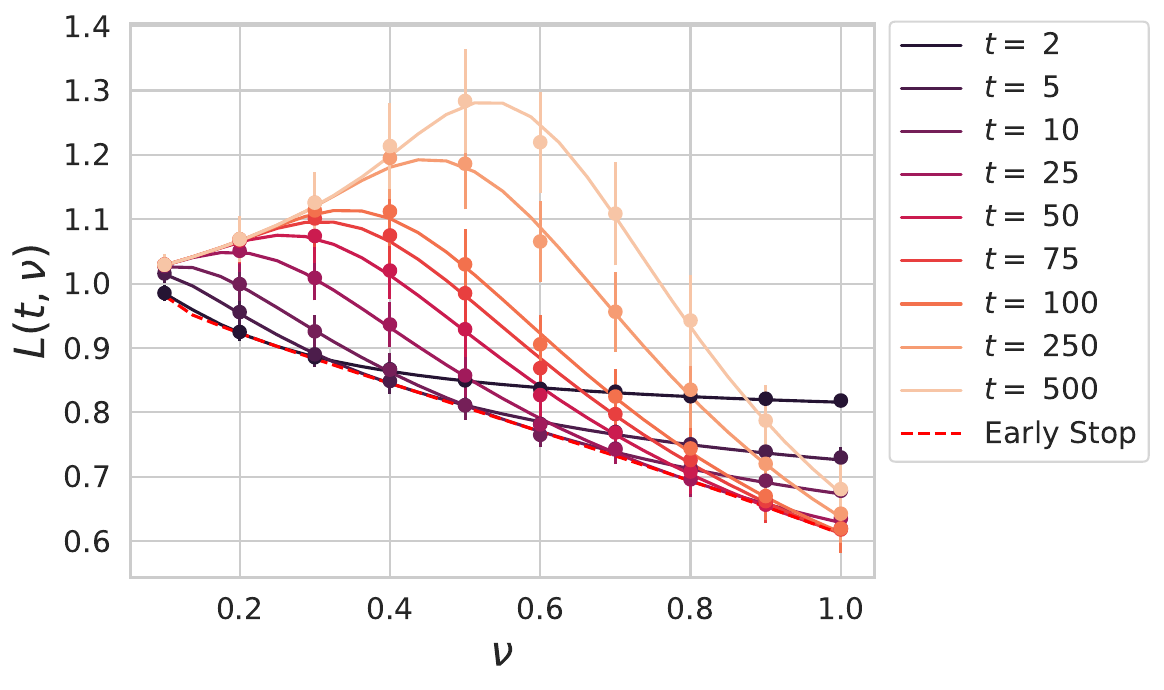}
    \caption{The white bandlimited model ($\lambda_k = 1$) with $\alpha = 0.8$ and varying model size $\nu$ with no explicit noise $\sigma=0$ exhibits double descent at late time. Optimal early stopping, like optimal regularization, recovers monotonic scaling with $\nu$. }
    \label{fig:double_descent}
\end{figure}

To gain intuition for the model, we can first analyze the case where $\lambda_k =1$, which has a simpler DMFT description since each of the $M$ features are statistically identical. We illustrate the dependence of the loss on model size $\nu$ and training time $t$ for $\alpha < 1$ in Figure \ref{fig:double_descent}. We note that the loss can be non-monotonic in $\nu$ at late training times, but that monotonicity is maintained for optimal early stopping, similar to results on optimal regularization in linear models \cite{advani2020high} and random feature models \cite{mei2022generalization, simon2023more}.

\subsection{Derivation}

In the case of all $\lambda_k = 1$ we have the following definitions
\begin{equation}
\begin{aligned}
    \mathcal R_1 (\omega) &= 1 - \frac1\alpha \frac{\mathcal R_1(\omega) \mathcal R_3(\omega)}{i \omega + \mathcal R_1(\omega) \mathcal R_3(\omega)}\\
    \mathcal R_3 (\omega) &= 1 - \frac1\nu \frac{\mathcal R_1(\omega) \mathcal R_3(\omega)}{i \omega + \mathcal R_1(\omega) \mathcal R_3(\omega)}
\end{aligned}
\end{equation}
Writing $\mathcal R_1 = 1 - \frac{\nu}{\alpha} (\mathcal R_3-1)$ allows us to solve for $\mathcal R_3$ exactly:
\begin{equation}
\begin{aligned}
    \mathcal R_3 \left( i \omega + \mathcal R_3 (1 + \frac{\nu}{\alpha} (\mathcal R_3- 1)) \right) =  i \omega + \mathcal R_3 (1 + \frac{\nu}{\alpha} (\mathcal R_3- 1)) - \frac{1}{\nu} \left( \mathcal R_3 (1 + \frac{\nu}{\alpha} ( \mathcal R_3 - 1)) \right).
\end{aligned}
\end{equation}
This is a cubic equation that can be solved for $\mathcal R_3$ as a function of $\omega$. In the limit of $\alpha \to \infty$ this simplifies to:

\begin{equation}
\begin{aligned}
    \mathcal R_3 i \omega + \mathcal R_3^2 &= i \omega + \left(1 - \frac{1}{\nu} \right) \mathcal R_3   \\
    \Rightarrow \mathcal R_3 &= \frac{1}{2} [ (1- \nu^{-1} - i\omega) + \sqrt{ (1- \nu^{-1} - i\omega)^2 + 4 i \omega } ].
\end{aligned}
\end{equation}

\subsection{Timescale Corrections in The Small $\nu$ Regime }

By expanding the above in the limit of small $\nu$ we get that $\mathcal R_3$ goes as
\begin{align}
    \mathcal{R}_3 \sim \frac{i\omega}{\nu^{-1} - 1 + i\omega} \ , \ \nu \to 0
\end{align}
From this approximate response function, we find that the transfer function takes the form
\begin{align}
    H(\tau) &= \int \frac{d\omega}{2\pi} \frac{e^{i\omega \tau}}{i\omega + \frac{i\omega}{\nu^{-1} - 1 + i\omega}} = \int \frac{d\omega}{2\pi} \frac{(\nu^{-1} - 1 + i\omega)e^{i\omega \tau}}{i\omega \left[ \nu^{-1} + i\omega   \right]} \nonumber
    \\
    &= (1-\nu) + \nu e^{-\tau / \nu},
\end{align}
where in the last line, we used the residue theorem. We note that in this perturbative approximation that this transfer function is always greater than the transfer function at $\nu \to \infty$ which is $e^{-\tau}$. Thus finite $\nu$ leads to higher bias in this regime.  We define bias and variance precisely in Appendix \ref{app:bias_var_defn}. 

\subsection{Timescale corrections in fully expressive regime $\nu > 1$}

\begin{figure}[H]
    \centering
    \includegraphics[width=0.7\linewidth]{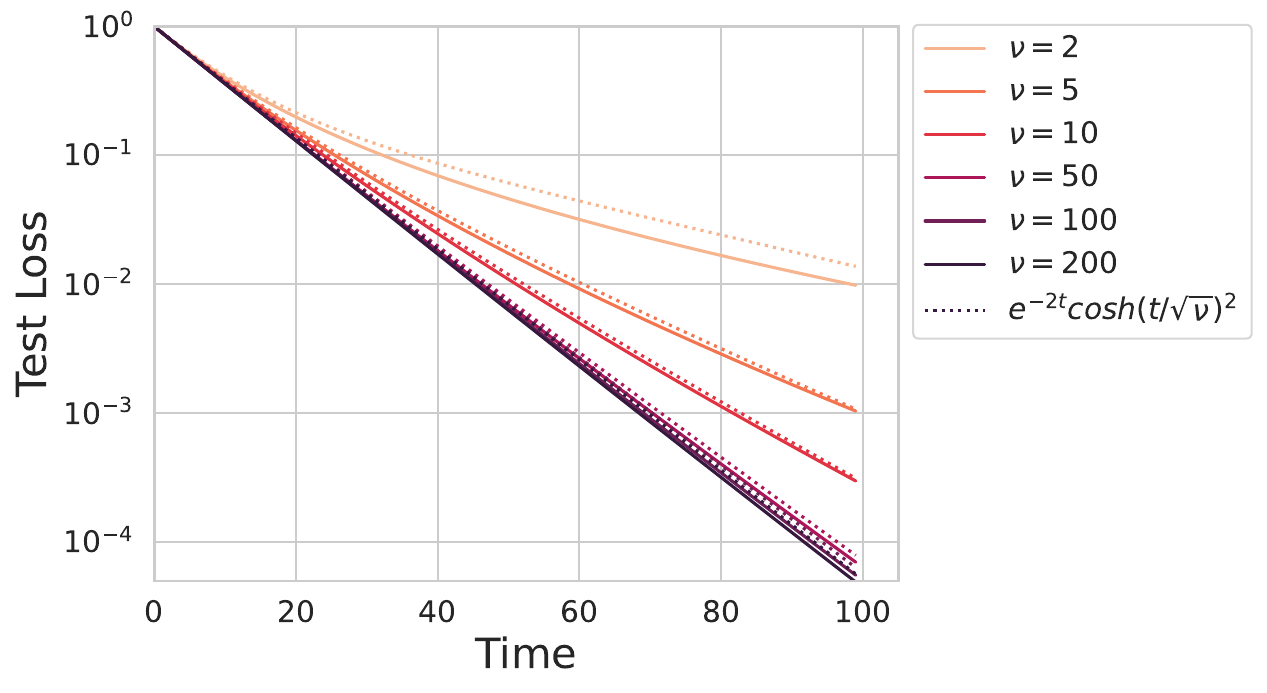}
    \caption{Slower timescales in the $\nu > 1$ regime for white bandlimited features. }
    \label{fig:slower_timescales}
\end{figure}

For $\nu \gg 1$, we can approximate $R_3(\omega) \sim 1- \nu^{-1} (1+ i\omega)^{-1}$, we have
\begin{align}
    H(\tau) &\sim \int \frac{d\omega}{2\pi} \frac{e^{i\omega \tau}}{i\omega + 1 - \nu^{-1} (i\omega + 1)^{-1}} =\int \frac{d\omega}{2\pi} \frac{e^{i\omega \tau} (1 + i\omega)}{(i\omega + 1 - \nu^{-1/2}) (i\omega + 1 + \nu^{-1/2})}  \nonumber
    \\
    &= \frac{1}{2} e^{-\tau(1+\sqrt\nu)} + \frac{1}{2} e^{-\tau(1-\sqrt \nu)}  = e^{-\tau} \cosh\left( \tau /\sqrt{\nu} \right)
\end{align}
where we used the residue theorem after closing the contour in the upper half-plane. In Figure \ref{fig:slower_timescales}, we show that this perturbative approximation does capture a slowdown in the dynamics for large but finite $\nu$. 

\section{Power-Law Bottleneck Scalings}\label{app:power_law_bottleneck}

In this section we calculate the scaling of the loss with the various limiting resources (time, model size, and data) when using power law features. Since the power-law features give a trace class kernel (\textit{i.e.} $\sum_{k=1}^\infty \lambda_k < \infty$), we use the non-proportional limit formalism in Appendix \ref{app:non_prop_limit}, which gives an expression for $\mathcal{L}(t,N,P)$ with $M$ already considered infinite. While the resulting expressions are not a formal proportional thermodynamic limit and finite $N,P$ corrections exist in the form of fluctuations from one random realization of the system to another. These corrections decay rapidly enough at finite $N,P$ for this mean field theory to be accurate and descriptive in realistic systems \cite{bordelon2020spectrum, simon2023more, cheng2022dimension}. We plot this variability of random finite size experiments as highlighted standard deviations in the main text figures. 

\subsection{Time Bottleneck}
The time bottleneck is defined as the limiting dynamics in the absence of any model or data finite size effects. To eliminate those effects, we simply study the $\alpha,\nu \to \infty$ limit
\begin{align}
    \mathcal{L}_\infty(t) = \lim_{P, N \to \infty} \mathcal{L}(t,P,N) .
\end{align}
In this limit, the response functions simplify to $\mathcal{R}_1(\omega) \mathcal{R}_3(\omega) \to 1$ so that
\begin{align}
    \mathcal{H}_k(\omega) = \frac{1}{i \omega + \lambda_k} \implies H_k(\tau)= e^{-\lambda_k \tau} \Theta(\tau).
\end{align}
Further, in this limit, we have that $C_0(t,s) = \frac{1}{M} \sum_k \lambda_k H_k(t) H_k(s) (w^\star_k)^2$ since all the variance terms (which depend on $\nu^{-1}, \alpha^{-1}$) drop out. Thus we have the following loss at time $t$,
\begin{align}
    \mathcal{L}(t) = \sum_k \lambda_k (w_k^\star)^2 e^{-2 \lambda_k t} \sim \int_1^\infty dk k^{-a} \exp\left( - 2  k^{-b} t \right) \sim t^{-(a-1)/b} .
\end{align}
where the final scaling with time can be obtained through either change of variables or steepest descent methods \cite{bordelon2022learning}. 

\subsection{Model Bottleneck}

In this section we take $\alpha, t \to \infty$. This leaves us with the following equation for $r \equiv \lim_{\omega \to 0} (i\omega)^{-1} \mathcal{R}_3(\omega)$. 
\begin{align}
    N &=  \sum_k \frac{\lambda_k r}{\lambda_k r + 1} \approx \int_1^\infty \frac{dk}{k^b/r + 1} \approx  r^{1/b}  \implies r \approx N^b.
\end{align}
Now, the large time limit of the transfer functions $H_k(\tau)$ can be obtained from the final-value theorem
\begin{align}
    \lim_{t \to \infty } H_k(\tau) = \lim_{\omega \to 0} \frac{i\omega}{i\omega + \lambda_k r i \omega} = \frac{1}{1+\lambda_k r}.
\end{align}
Now, integrating over the eigenvalue density to get the total loss (and disregarding prefactors)
\begin{align}
    \mathcal{L}(t) &\sim \int_1^\infty dk \ \frac{k^{-a}}{(1 + k^{-b} r)^2} \nonumber
    \\
    &\approx \frac{1}{r^2} \int_1^{N} dk \ k^{2b-a} + \int_{N}^\infty k^{-a}  \nonumber
    \\
    &= \frac{1}{2b+1 -a} [ N^{-(a-1)} - N^{-2b} ] +  \frac{1}{a-1} N^{-(a-1)}  \nonumber
    \\
    &\sim N^{-\min\{a-1,2b\}}
\end{align}
For difficult tasks where $a-1 < 2b$, we thus expect a powerlaw scaling of the form $\mathcal{L} \sim N^{-(a-1)}$ in this regime.

\subsection{Data Bottleneck}

In this section we take $\nu, t \to \infty$. This leaves us with the following equation for $r \equiv \lim_{\omega \to 0} (i\omega)^{-1} \mathcal{R}_1(\omega)$. 
\begin{align}
    P &=  \sum_k \frac{\lambda_k r}{\lambda_k r + 1} \approx \int_1^\infty \frac{dk}{k^b/r + 1} \approx  r^{1/b}  \implies r \approx P^b.
\end{align}
Now, the large time limit of the transfer functions $H_k(\tau)$ can again be obtained from the final-value theorem
\begin{align}
    \lim_{t \to \infty } H_k(\tau) = \lim_{\omega \to 0} \frac{i\omega}{i\omega + \lambda_k r i \omega} = \frac{1}{1+\lambda_k r} 
\end{align}
Now, integrating over the eigenvalue density to get the total loss gives
\begin{align}
    \mathcal{L}(t) &\sim \int_1^\infty dk \ \frac{k^{-a}}{(1 + k^{-b} r)^2} \nonumber
    \\
    &\approx \frac{1}{r^2} \int_1^{P} dk \ k^{2b-a} + \int_{P}^\infty k^{-a}  \nonumber
    \\
    &= \frac{1}{2b+1 -a} [ P^{-(a-1)} - P^{-2b} ] +  \frac{1}{a-1} P^{-(a-1)}  \nonumber
    \\
    &\sim P^{-\min\{a-1,2b\}}
\end{align}
For difficult tasks with $a < 2b + 1$, the loss will therefore scale as $P^{-(a-1)}$ in this data-bottleneck regime. 

\section{Optimization Extensions}\label{app:opt_exts}

\subsection{Discrete Time}

In this section, we point out that DMFT can also completely describe discrete time training as well. In this section we consider discrete time gradient descent with learning rate $\eta$ 
\begin{align}
    &\v^0(t+1) = \v^0(t) - \eta \v^4(t) \nonumber
    \\
    &\v^4(t) = \frac{1}{\nu \sqrt M} \bm A^\top \v^3(t) \ , \ \v^3(t) = \frac{1}{\sqrt M} \A \v^2(t)  \nonumber
    \\
    &\v^2(t) = \frac{1}{\alpha \sqrt M} \bm\Psi^\top \v^1(t) \ , \ \v^1(t) = \frac{1}{\sqrt M} \bm\Psi \v^0(t)
\end{align}
Following either the MSR or cavity derivation, we obtain an analgous set of limiting DMFT equations defined for integer times $t \in \mathbb{Z}$,
\begin{align}
    &v^0_k(t+1) = v^0_k(t) - \eta v^4_k(t) + \delta(t+1) w^\star_k \nonumber
    \\
    &v^1(t) = u^1(t) + \alpha^{-1} \sum_s R_{0,2}(t,s) v^1(s) \nonumber
    \\
    &v^2_k(t) = u^2_k(t)+ \sum_s R_1(t,s) v^0_k(s)\nonumber
    \\
    &v^3(t) = u^3(t) + \sum_s R_{2,4}(t,s) v^3(s) \nonumber
    \\
    &v^4_k(t) = u^4_k(t) + \sum_s R_3(t,s) v^2_k(s)
\end{align}
The delta function in this context is defined as 
\begin{align}
    \delta(t+1) = \begin{cases}
    1 & t = -1
    \\
    0 & \text{else}
\end{cases}
\end{align}
ensures that the initial condition $v_k^0(0) = w^\star_k$ is satisfied. These iteration equations can be closed for the response functions and correlation functions and solved over $T \times T$ matrices. 

Alternatively, we can also solve this problem in an analogous frequency space. Analogous to the Fourier transform method, the equations in discrete time can be closed in terms of the $Z$-transform
\begin{align}
    v(z) = \sum_{t=-\infty}^\infty z^{-t} v(t) 
\end{align}
Applying this transform gives us the following expression for the $v^0_k$ fields. 
\begin{align}
    v^0_k(z) = \frac{z w^\star_k - \eta u^4_k(z) - \eta \mathcal R_3(z) u^2_k(z)}{z-1 + \eta \lambda_k \mathcal R_1(z) \mathcal R_3(z) } \equiv \mathcal{H}_k(z) \left[ z w^\star_k - \eta u^4_k(z) - \eta \mathcal R_3(z) u^2_k(z) \right]
\end{align}
Similar to the Fourier case, the final losses can be extracted as the $z \to 1$ limit of these objects. 

\subsection{Momentum}

As mentioned in appendix \ref{app:field_th_deriv}, it is straightforward to extend the DMFT treatment beyond just gradient descent dynamics to include a momentum term with momentum $\beta$. 

We first consider this replacement in continuous time. This requires applying the following replacement:
\begin{equation}
    \partial_t v^0_k(t) = - v^4_k(t) \to (\beta \partial_t^2 + \partial_t) v^0_k(t) = - v^4_k(t).
\end{equation}
This slightly modifies the expressions for the response functions. For example, in Fourier space the response functions become:
\begin{equation}
  \begin{aligned}
    &\mathcal{R}_{0,2}(\omega) = - \frac{1}{M} \sum_k \frac{ \lambda_k }{\epsilon + i\omega + \beta(i\omega)^2 + \lambda_k \mathcal{R}_{1}(\omega) \mathcal{R}_{3}(\omega)} \mathcal{R}_{3}(\omega) 
    \\
    &\mathcal{R}_{2,4}(\omega) = - \frac{1}{M}\sum_k \frac{\lambda_k }{\epsilon + i\omega + \beta(i\omega)^2 + \lambda_k\mathcal{R}_1(\omega)\mathcal{R}_3(\omega) }\mathcal{R}_1(\omega).
\end{aligned}  
\end{equation}
In discrete time, momentum updates can be expressed as 
\begin{align}
    \v^0(t+1) = \v^0(t) - \eta \bm b(t) \nonumber
    \\
    \bm b(t) = \v^4(t) + \mu \bm b(t-1)
\end{align}
where $\bm b(t)$ is the filtered version of the loss gradient (the $\bm v^4(t)$ field) with momentum coefficient $\mu$ and $\eta$ is the learning rate. The dependence on the $\bm b(t)$ field can be eliminated by turning this into a second order difference equation
\begin{align}
    \v^0(t+1) - \v^0(t) - \mu\left( \v^0(t) - \v^0(t-1) \right) = - \eta \v^4(t) .
\end{align}
Again, the final result can be expressed in terms of the $Z$-transformed transfer functions $\mathcal H_k(z)$ which have the form
\begin{align}
    \mathcal{H}_k(z) = \frac{1}{z - 1 -\mu + \mu z^{-1}  + \eta \mathcal{R}_1(z) \mathcal R_3(z) } .
\end{align}

\subsection{One Pass SGD}\label{app:one_pass_sgd}

In this section we derive online SGD with projected features. At each step a random batch of $B$ samples are collected (independent of previous samples), giving a matrix $\bm\Psi(t) \in \mathbb{R}^{B \times M}$ of sampled features. The update at step $t$ is 
\begin{align}
    \v^0(t+1) = \v^0(t) + \eta \left( \frac{1}{N} \A^\top \A \right) \left( \frac{1}{B} \bm\Psi(t)^\top \bm\Psi(t) \right) \v^0(t).
\end{align}
The DMFT limit gives the following statistical description of the fields, which decouple over time for the $v^1(t),v^2_k(t)$ but remain coupled across time for $v^3(t), v^4_k(t)$
\begin{equation}
\begin{aligned}
    v^1(t) &= u^1(t)  \ , \ u^1(t) \sim \mathcal{N}(0, C_0(t,t) \delta(t-s) ) ,
    \\ 
    v^2_k(t) &= u^2_k(t) + \lambda_k v^0_k(t) \ , \ u^2_k(t) \sim \mathcal{N}\left(0, \frac{1}{B} \lambda_k C_1(t,t) \delta(t-s) \right) 
    \\
    v^3(t) &= u^3(t) + \frac{1}{N} \sum_{s} R_{2,4}(t,s) v^3(s) \ , \ u^3(t) \sim \mathcal{N}(0, C_2(t,s) ) 
    \\
    v^4_k(t) &= u^4_k(t) + \sum_{s} R_3(t,s) v^2_k(s) \ , \ u^4_k(t) \sim \mathcal{N}\left(0, \frac{1}{N} C_3(t,s) \right)
    \\
    v^0_k(t+1) &= v^0_k(t) - \eta v^4_k(t) .
\end{aligned}    
\end{equation}
This system cannot exhibit overfitting effects as we have the statistical equivalence between the covariance of $v^1$ and the test loss: 
\begin{align}
    \hat{\mathcal L}(t) = \left< v^1(t)^2 \right> = \left< u^1(t)^2 \right> = C_0(t,t) = \mathcal L(t)
\end{align}
We note that this is very different than the case where data is reused at every step, which led to a growing gap between train and test loss as we derive in Appendix \ref{app:buildup_overfitting}.

We visualize some example results for one-pass SGD with power law features in Figure \ref{fig:online_sgd_dmft_version}. While we see that the same scaling laws with $t$ and $N$ hold, the dependence on batchsize $B$ is much weaker: the model never reaches an asymptote that scales with $B$ but rather experiences SGD noise that scales with $\eta/B$ for learning rate $\eta$. 

We summarize the key similarities and differences between the one-pass SGD and multi-pass batch GD settings

\begin{enumerate}
    \item If the learning rate is small and a continuous time limit of the dynamics is taken, then the SGD dynamics will agree with the $P \to \infty$ limit of our full batch gradient flow theory. This is a setting where finite data and SGD noise are negligible.
    \item If learning rate is non-negligible and batch size is finite, then SGD noise cannot be neglected and the SGD dynamics will be different than full pass GD. The SGD dynamics will be described by a discrete time DMFT given above.
    \item In general, the multi-pass version of the theory can have a train loss and test loss gap while the SGD theory never has a gap between training and test loss. 
    \item The SGD test loss can be limited by $t, N$, but the effect of finite batch size is basically some additive variance in the model outputs. Finite dataset size in the full batch GD can lead to a bottleneck scaling law ( like $L \sim P^{-(a-1)}$). 
\end{enumerate}

\section{Kernel Analysis of Feature Learning Networks}\label{app:feature_learning}

\begin{figure*}[h]
    \centering
    \subfigure[Final NTK Spectra for Varying Widths]{\includegraphics[width=0.32\linewidth]{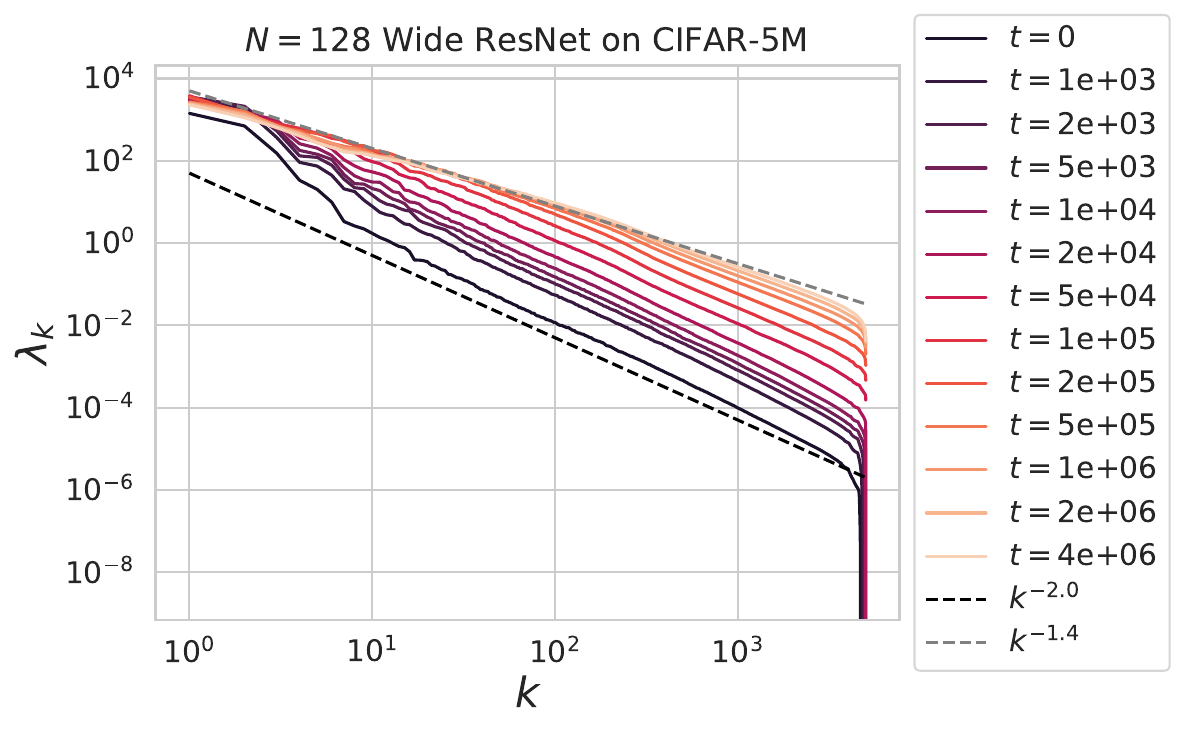}}
    \subfigure[Final NTK Task-Power Decay]{\includegraphics[width=0.32\linewidth]{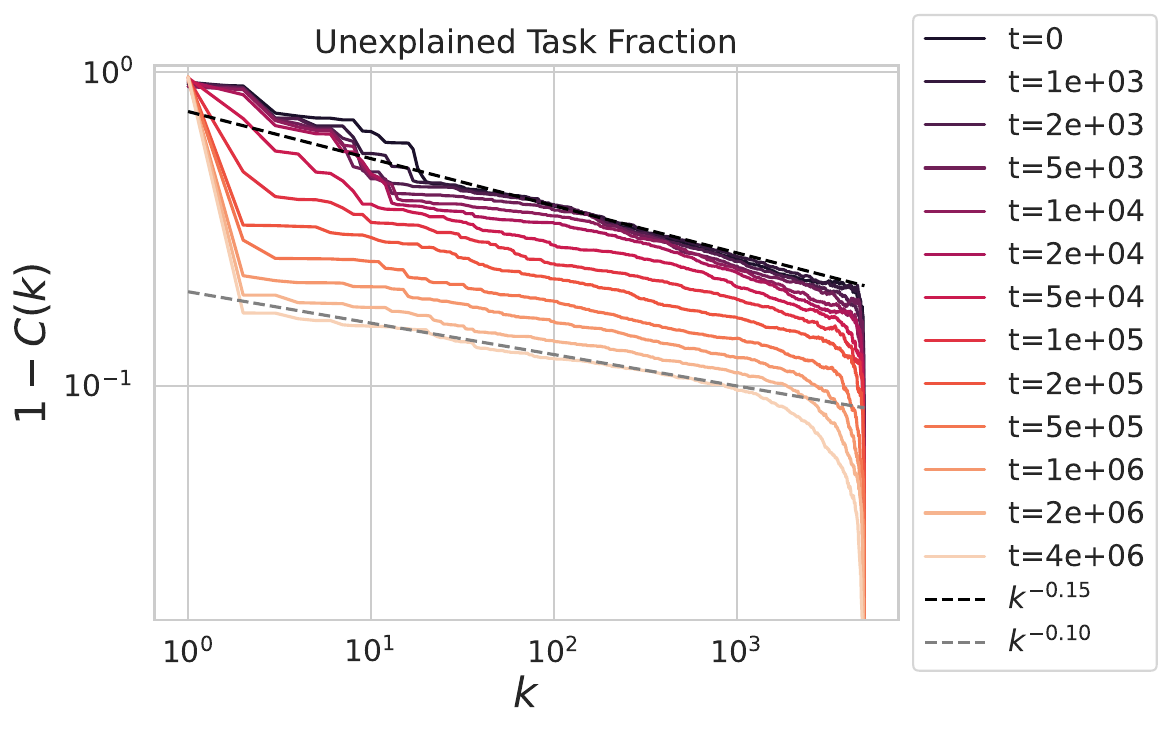}}
    \subfigure[Losses for Feature Learning Networks]{\includegraphics[width=0.32\linewidth]{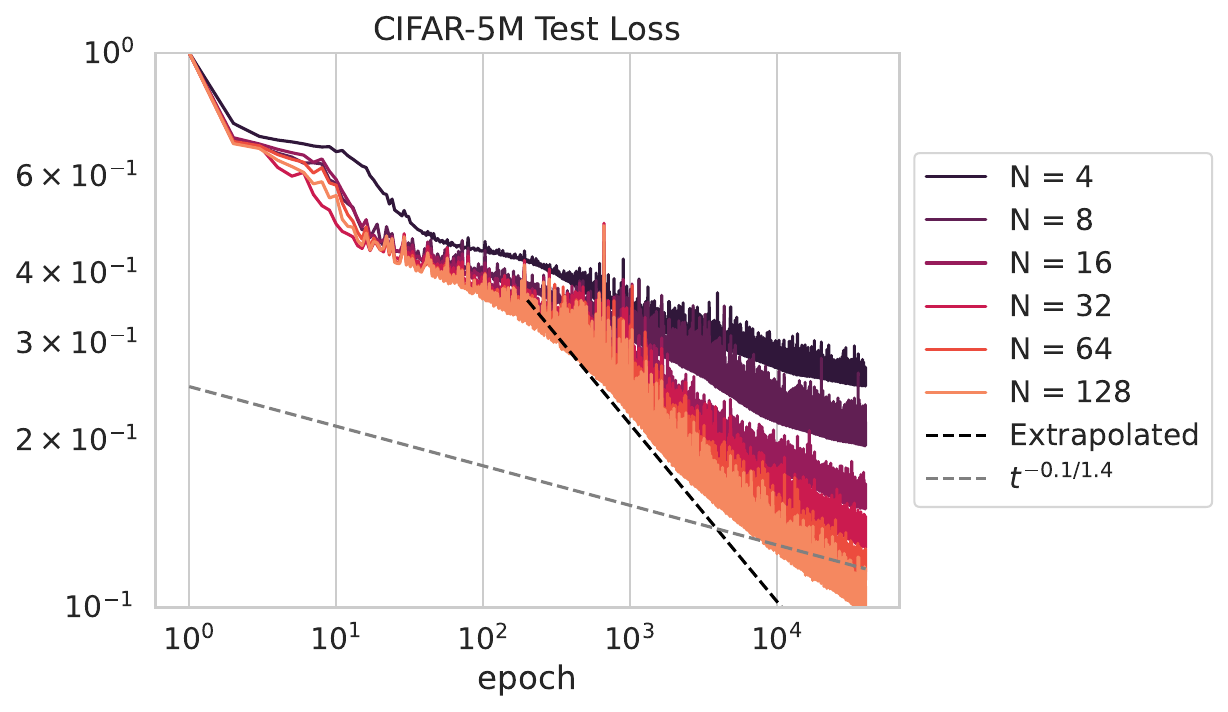}}\\
    \subfigure[]{\includegraphics[width=0.32\linewidth]{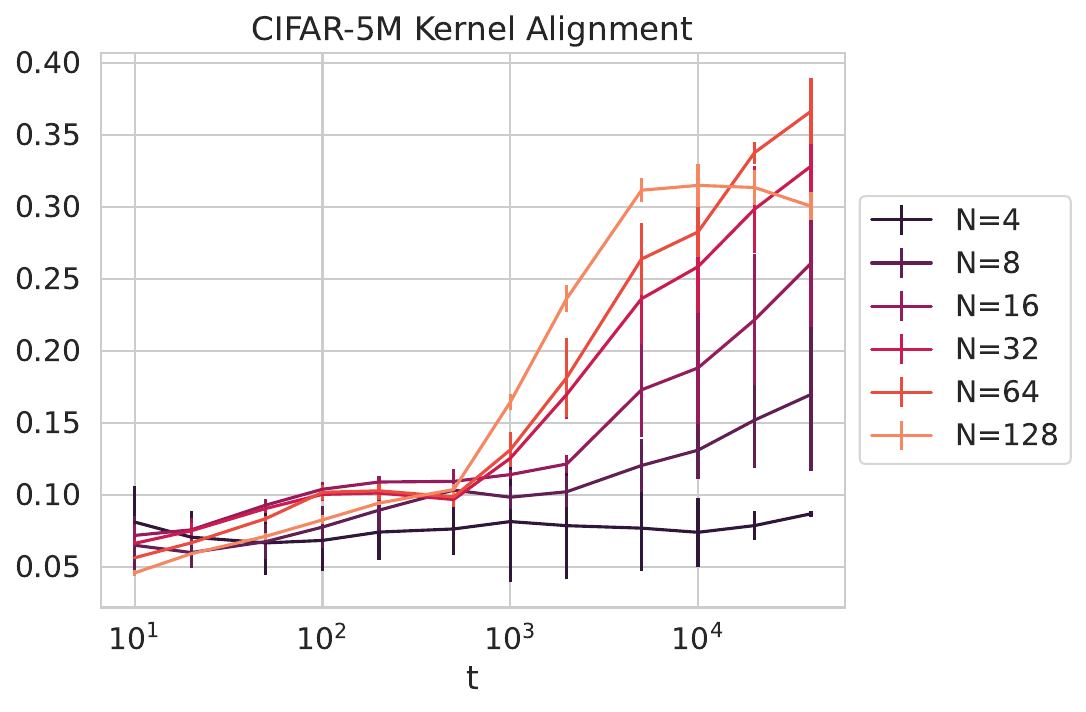}}
    \subfigure[]{\includegraphics[width=0.32\linewidth]{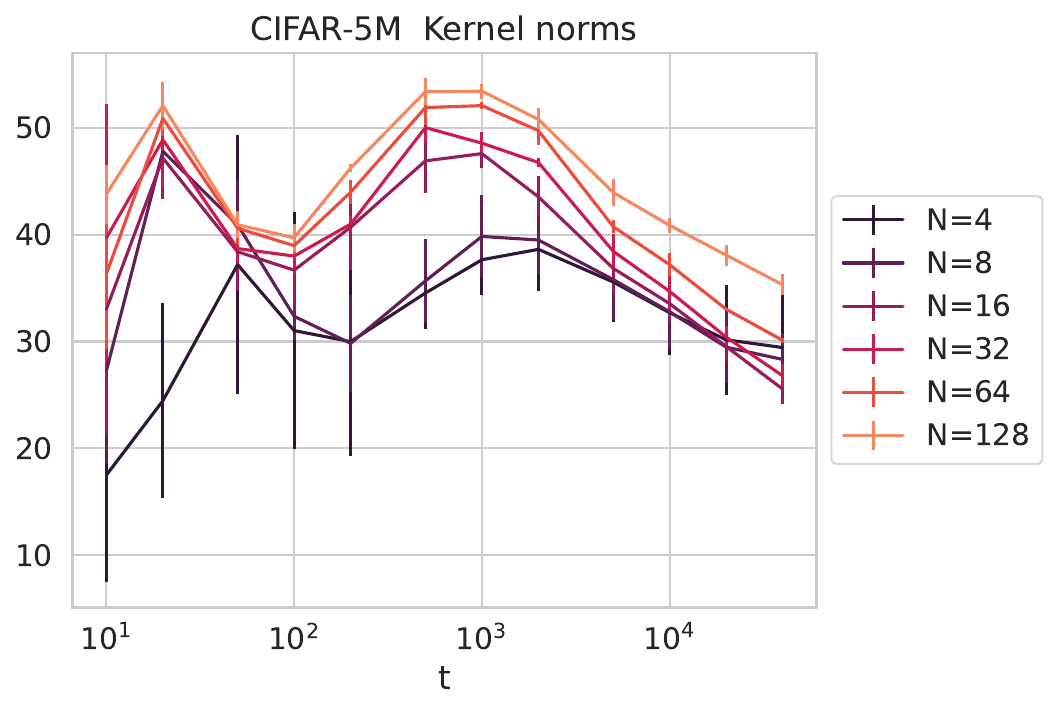}}
    \caption{a) The observed power law spectrum on a held out test set of the after-kernel for a width $N=128$ ResNet trained on CIFAR-5m. Early on in training, the spectrum flattens quite rapidly. At later times, the spectral decay remains relatively constant. b) The fraction of the task unexplained, as defined in Equation \ref{eq:Ck_defn}. Throughout training, the top eigenmode of the after-kernel explains more and more of the task. c) The test loss of the network. We see that the observed scaling of this quantity is faster than that predicted from analyzing the after-kernel. d) The kernel-target alignment of the after kernel improves throughout training time. The error bars here denote different ensemble members. Their relatively small size implies that the kernel trajectory is relatively deterministic over different initialization seeds. e) The norm of the after-kernel throughout training is relatively constant for this task.}
    \label{fig:cifar_5m_feature}
\end{figure*}

In Section \ref{sec:feature_learning}, we observed that feature learning networks can achieve better loss and compute-optimal scaling. In such settings, it may be useful to observe the \textit{after kernel}, namely the NTK at the end of training. This object can often shed insight into the structure of the learned network function \cite{atanasov2022neural, long2021properties} and its generalization. In some cases, it has been observed that the final kernel stabilizes during the course of training \cite{FortDPK0G20}, potentially allowing one to potentially deduce scaling laws from the spectrum and task-model alignment of this after-kernel, though other papers have observed contrary results \cite{vyas2022limitations}. 

Motivated by this, we study the NTKs of the finite-width networks trained for 64 epochs with  the animate-inanimate CIFAR-5m discrimination task. We observe in Figure \ref{fig:cifar_5m_feature} a) that the spectrum becomes flatter, with a decay exponent of close to $1.4$ down from $2.0$ for the initial kernel. 

The fraction of the task power unexplained is also observed to have a lower exponent in Figure \ref{fig:cifar_5m_feature} b), however there is also the presence of a low rank spike indicative of the kernel aligning to this discrimination tasks. 

From these scalings we can obtain the $a$ and $b$ exponents and get a prediction for the scaling of the test loss. We plot this in grey in Figure \ref{fig:cifar_5m_feature} c). The observed scaling (in black) is much better than that predicted by the after-kernel. This is an indication the the after kernel continues evolving in this task, improving the scaling exponent of the test loss. 

The kernel-target alignment \cite{cortes2012algorithms}, as measured by 
\begin{equation}
A = \frac{\bm y^\top \bm K \bm y}{\bm y^\top \bm y |\bm K|_F},    
\end{equation}
is plotted in \ref{fig:cifar_5m_feature} d). Here $\bm y$ is the target labels on a held-out test set, and $\bm K$ is the gram matrix of the after-kernel on this test set. We indeed observe a consistent increase in this quantity across time. This gives an indication that understanding the evolution of the after-kernel will be useful

\section{Numerical Recipes}

\subsection{Iteration of DMFT Equations on Time $\times$ Time matrices}

The simplest way to solve the DMFT equations is to iterate them from a reasonable initial condition \cite{mignacco2020dynamical, bordelon2022self}. We solve in discrete time for $T \times T$ matrices $\{ \bm R_{0,2}, \bm R_{1} , \bm R_{2,4} , \bm R_3, \bm C_0, \bm C_1, \bm C_2, \bm C_3 \}$ which have entries $[\bm R]_{t,s} = R(t,s), [\bm C]_{t,s} = C(t,s)$, etc. We let $\bm\Theta(t,s) = \eta \Theta(t-s)$ where $\eta$ is the learning rate. 
\begin{enumerate}
    \item Solve for the response functions by updating the closed equations as matrices by iterating the equations. 
    \begin{equation}
    \begin{aligned}
        &\bm R_{0,2,k} \leftarrow -  \left[ \bm \Theta^{-1} + \lambda_k \bm R_3 \bm R_1 \right]^{-1} \bm R_3 ,
        \\ 
        &\bm R_{0,2} \leftarrow \frac{1}{M} \sum_{k} \lambda_k \bm R_{0,2,k} ,
        \\
        &\bm R_1 \leftarrow \left[ \I - \alpha^{-1} \bm R_{0,2} \right]^{-1}  ,
        \\
        &\bm R_{2,4,k} \leftarrow - \lambda_k \left[ \I + \lambda_k \bm R_1 \bm\Theta \bm R_3 \right]^{-1} \bm R_1 \bm \Theta  \ , \ \bm R_{2,4} = \frac{1}{M} \sum_{k} \bm R_{2,4,k},
        \\
        &\bm R_3 \leftarrow \left[ \I - \nu^{-1} \bm R_{2,4} \right]^{-1}.
    \end{aligned}
    \end{equation}

    \item Once these response functions have converged, we can iterate the equations for the correlation functions
    \begin{equation}
     \begin{aligned}
        &\bm C_{0,k} \leftarrow \left[ \I +\lambda_k \bm \Theta \bm R_3 \bm R_1 \right]^{-1} \left[ (w^\star_k)^2 \bm 1 \bm 1^\top +  \bm \Theta \left( \nu^{-1} \bm C_3 + \frac{\lambda_k}{\alpha} 
 \bm R_3 \bm C_1 \bm R_3^\top \right) \bm\Theta^\top \right] \left[  \I +\lambda_k \bm \Theta \bm R_3 \bm R_1 \right]^{-1 \top} ,
 \\
 &\bm C_0 \leftarrow  \frac{1}{M} \sum_{k=1}^M \lambda_k  \bm C_{0,k}  ,
 \\
 &\bm C_1  \leftarrow \bm R_1 \bm C_0 \bm R_1^\top ,
 \\
 &\bm C_{2,k} \leftarrow \left[ \bm I + \lambda_k \bm R_1 \bm\Theta \bm R_3 \right]^{-1} \left( \frac{\lambda_k}{\alpha} \bm C_1 +  \bm R_1 \left[ (w^\star_k)^2 \lambda_k^2  \bm 1 \bm 1^\top  + \frac{\lambda_k^2}{\nu} \bm\Theta \bm C_3 \bm\Theta^\top  \right] \bm R_1^\top \right) \left[ \bm I + \lambda_k \bm R_1 \bm\Theta \bm R_3 \right]^{-1 \top} ,
 \\
 &\bm C_{2} \leftarrow \frac{1}{M} \sum_{k} \bm C_{2,k}.
    \end{aligned}
    \end{equation}
\end{enumerate}
After iterating these equations, one has the discrete time solution to the DMFT order parameters and any other observable can then be calculated. 

\subsection{Fourier Transform Method}

To accurately compute the Fourier transforms in the model/data bottleneck regime ($\alpha < 1$ or $\nu <1$) we have that $\mathcal R_1(\omega) \mathcal R_3(\omega) \sim i\omega r$ as $\omega \to 0$ so we must resort to analyzing the principal part and the delta-function contribution to the integral. Construct a shifted and non-divergent version of the function $\mathcal H(\omega)$. 
\begin{equation}
\begin{aligned}
    \mathcal{H}(\omega) &= \mathcal{\tilde H}(\omega) + \frac{1}{\epsilon + i \omega(1+r)} 
    \\
    \mathcal{\tilde H}(\omega) &= \frac{1}{\epsilon + i\omega + \mathcal R_1(\omega) \mathcal R_3(\omega)} - \frac{1}{\epsilon + i \omega (1 + r)} = \frac{i\omega r - \mathcal{R}_1(\omega) \mathcal{R}_3(\omega)}{(\epsilon + i\omega + \mathcal R_1(\omega) \mathcal R_3(\omega)) (\epsilon + i \omega (1 + r)) },
\end{aligned}    
\end{equation}
where $r = \lim_{\omega \to 0} \frac{1}{i\omega} \mathcal{R}_1(\omega) \mathcal{R}_3(\omega)$. We see that rather than diverging like $\mathcal H(\omega)$, this function $\mathcal{\tilde H}(\omega)$ vanishes as $\omega \to 0$. We therefore numerically perform Fourier integral against $\tilde{H}(\omega)$ and then add the singular component which can be computed separately. 
\begin{align}
    H(\tau) = \int \frac{d\omega}{2\pi} e^{i\omega\tau} \mathcal{\tilde H}(\omega) + \int \frac{d\omega}{2\pi} \frac{e^{i\omega\tau}}{\epsilon + i\omega(1 + r)} =  \int \frac{d\omega}{2\pi}  e^{i\omega\tau} \mathcal{\tilde H}(\omega) + \frac{1}{1+r}
\end{align}
where we used the fact that
\begin{align}
    \frac{1}{\epsilon + i\omega(1+r)} = \frac{\pi}{1+r} \delta(\omega) - \frac{i}{1+r} \mathcal P( \omega^{-1} ) \ , \ \epsilon \to 0
\end{align}
The Dirac mass is trivial to integrate over giving $\frac{1}{2(1+r)}$. Lastly, we must perform an integral of the type
\begin{align}
   -   \frac{i}{1+r}  \mathcal P \int_{-\infty}^\infty \frac{d\omega}{2\pi}  \frac{e^{i\omega \tau}}{\omega} =  \frac{1}{\pi(1+r)} \int_0^\infty d\omega \frac{\sin(\omega \tau)}{\omega} = \frac{1}{2(1+r)}  
\end{align}
Adding these two terms together, our transfer function has the form
\begin{align}
    H(\tau) = \frac{1}{1+ r} + \int_{-\infty}^\infty \frac{d\omega}{2\pi} e^{i\omega\tau} \mathcal{\tilde H}(\omega) .
\end{align}
The last integral can be performed numerically, giving a more stable result.

\section{Compute Optimal Scaling from Sum of Power-Laws}\label{app:C_scale}

We suppose that the loss scales as (neglecting irrelevant prefactors)
\begin{align}
    \mathcal L= t^{-r_t} + N^{-r_N} + P^{-r_P} + \mathcal{L}_\infty
\end{align}
Our goal is to minimize the above expression subject to the constraint that compute $C = Nt$ is fixed. Since $C$ is fixed we can reduce this to a one-dimensional optimization problem
\begin{align}
    \min_{N} \left[ C^{-r_t} N^{r_t} + N^{-r_N}  \right] 
\end{align}
The optimality condition $\partial_N L = 0$ is
\begin{align}
    &r_t C^{-r_t}  N^{r_t-1} - r_N N^{-r_N - 1} = 0 \nonumber
    \\
    &\implies N \propto C^{\frac{r_t}{r_t + r_N}} \implies t \propto C^{\frac{r_t}{r_t + r_N}} 
\end{align}
From this last expression one can evaluate the loss at the optimum
\begin{align}
    \mathcal{L}_\star(C) \propto C^{- \frac{r_t r_N}{r_t + r_N}} .
\end{align}

\end{document}